\documentclass[10pt,twocolumn,letterpaper]{article}

\usepackage{cvpr}              

\usepackage{multirow}
\usepackage{comment}
\usepackage{array}

\makeatletter
\renewcommand\thefootnote{\fnsymbol{footnote}}
\makeatother

\usepackage{bm}

\definecolor{cvprblue}{rgb}{0.21,0.49,0.74}
\usepackage[pagebackref,breaklinks,colorlinks,allcolors=cvprblue]{hyperref}

\def\paperID{*****} 
\def\confName{CVPR}
\def\confYear{2026}

\title{Adverse-to-the-eXtreme Panoptic Segmentation: URVIS 2026 Study and Benchmark}

\author{
Yiting Wang\footnotemark[1] \and
Nolwenn Peyratout\footnotemark[1] \and
Tim Brodermann\footnotemark[1] \and
Jiahui Wang \and
Yusi Cao \and
Michele Cazzola \and
Elie Tarassov \and
Takuya Kobayashi \and
Abderrahim Kasmi \and
Guillaume Allibert \and
Cédric Demonceaux \and
Valentina Donzella\and
Kurt Debattista \and
Radu Timofte \and
Zongwei Wu \and
Christos Sakaridis 
}

\begin{document}

\maketitle
\footnotetext[1]{Organizers of the URVIS 2026 Challenge : MUSES-AXPS — Adverse-to-the-eXtreme Panoptic Segmentation \\ Webpage of the MUSES-AXPS challenge: \url{https://urvis-workshop.github.io/challenge-Muses.html}}
\makeatletter
\renewcommand\thefootnote{\arabic{footnote}}
\makeatother
\begin{abstract}
This paper presents the report of the URVIS 2026 challenge on adverse-to-extreme panoptic segmentation. As the first challenge of its kind, it attracted 17 registered participants and 47 submissions, with 4 teams reaching the final phase. The challenge is based on the MUSES dataset, a multi-sensor benchmark for panoptic segmentation in adverse-to-extreme weather, including RGB frame camera, LiDAR, radar, and event camera data. Weighted Panoptic Quality (wPQ) is designed and adopted as the official ranking metric for fair evaluation across weather conditions. In this report, we summarise the challenge setting and benchmark results, analyse the performance of the submitted methods, and discuss current progress and remaining challenges for robust multimodal panoptic segmentation.
\end{abstract}

\section{Introduction}
Reliable situational awareness is a key requirement for the safe deployment of artificial intelligence in real-world systems, especially in robotics and autonomous driving. Among scene understanding tasks, panoptic segmentation is particularly attractive because it provides a unified pixel-level interpretation of both semantic regions and object instances, enabling the identification of traffic participants, potential hazards, and drivable areas in complex road scenes \cite{brodermann2024muses,wang2025robustness,kirillov2019panoptic,brodermann2025cafuser,wang2023effect}. However, real-world deployment remains challenging, as adverse weather introduces modality-specific degradations in multimodal sensor data owing to the different physical characteristics of each sensor, thereby undermining the robustness of panoptic segmentation.

RGB-based panoptic segmentation has advanced rapidly in recent years, spanning classical convolutional architectures, masked transformer-based designs, and more recently diffusion-based formulations \cite{cheng2021per,cheng2020panoptic,li2023mask,jain2023oneformer,xu2023open}. Nevertheless, most existing progress has been built around frame-based visual data alone, and its robustness is inherently limited by the failure modes of a single modality. 

\begin{figure}
    \centering
    \includegraphics[width=0.8\linewidth, trim=0 50 0 30, clip]{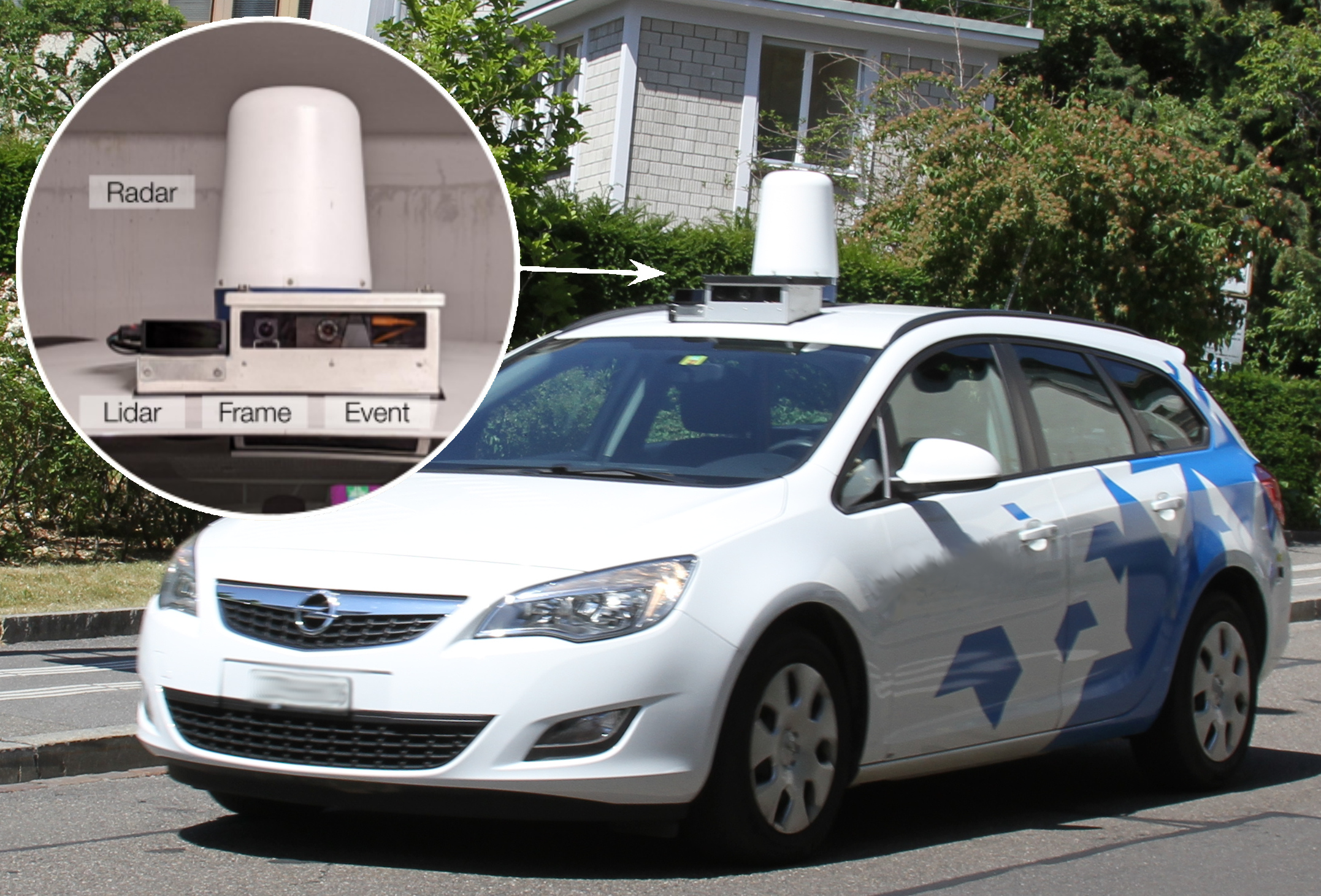}
    \caption{\textbf{Sensor configuration of MUSES.}}
    \label{fig:car_pic}
    \vspace{-8pt}
\end{figure}

This situation changed with MUSES, which introduced synchronized RGB frame camera, LiDAR, RADAR, and event camera data together with high-quality 2D panoptic annotations under diverse weather and illumination conditions, thereby making multimodal panoptic segmentation in adverse driving scenarios practically accessible \cite{brodermann2024muses}. More importantly, MUSES highlights that the value of multimodal perception lies not only in sensor complementarity, but also in the fact that different modalities degrade in fundamentally different ways. 

Under adverse weather, RGB cameras often suffer from reduced visibility, contrast attenuation, blur, and partial occlusion caused by atmospheric scattering, precipitation, and lens contamination. LiDAR, in contrast, is affected by attenuation and backscatter: adverse particles such as fog droplets, raindrops, or snowflakes can weaken valid returns from distant objects while also generating spurious echoes at incorrect ranges, leading to both missing far-away points and noisy false measurements \cite{wang2026aurora}. Event cameras offer high dynamic range and low latency, but since they respond to changes in brightness rather than absolute intensity, their signals can become unstable or noisy when low illumination, precipitation, and sensor noise perturb the temporal contrast triggering process. RADAR is generally the most weather-robust modality; however, provided as FMCW range--azimuth power measurements, whose representation differs substantially from other sensor data observations, so its contribution to dense pixel-level parsing is less direct and remains challenging to exploit effectively.

Recent methods have begun to address this gap. CAFuser \cite{brodermann2025cafuser} proposes condition-aware multimodal fusion, where a global condition token inferred from RGB guides the fusion process and modality-specific adapters align heterogeneous sensor inputs into a shared latent space. DGFusion \cite{brodermannn2026dgfusion} further extends this direction by introducing depth-guided fusion, using an auxiliary depth prediction branch and depth-aware local tokens to spatially modulate cross-modal interactions according to distance-dependent sensor reliability. These works represent important first steps toward robust multimodal semantic and panoptic perception under adverse conditions. Nevertheless, this remains a relatively new research area with substantial room for progress. In this context, our challenge aims to provide a common benchmark and stimulate broader exploration of robust multimodal panoptic segmentation under diverse real-world weather conditions. 

\section{MUSES Dataset}

We base the URVIS 2026 challenge on the MUSES~\cite{brodermann2024muses}, which is a real-world multimodal dataset for dense scene understanding under adverse environmental conditions. In contrast to other driving datasets, MUSES emphasizes both the diversity and severity of real-world environmental degradations, making it particularly suitable for evaluating robust perception systems.

MUSES comprises 2,500 synchronized driving scenes, each recorded with a sensor suite including an RGB frame camera, LiDAR, radar, an event camera, and IMU/GNSS signals as seen in Figure~\ref{fig:car_pic}. These modalities provide complementary information: while RGB images capture appearance, LiDAR and radar offer robustness to illumination and weather, and event cameras provide high dynamic range and temporal resolution.

\begin{figure*}[t]
\centering
\begin{tabular}{@{}c@{\hspace{0.03cm}}c@{\hspace{0.03cm}}c@{\hspace{0.03cm}}c@{\hspace{0.03cm}}c@{\hspace{0.03cm}}c@{}}
\subfloat{\small  RGB Image} &
\subfloat{\small  Lidar} &
\subfloat{\small  Events}&
\subfloat{\small  Radar}&
\subfloat{\small  Ref. Image}&
\subfloat{\small  Panoptic GT}  \\


\includegraphics[width=0.15\textwidth]{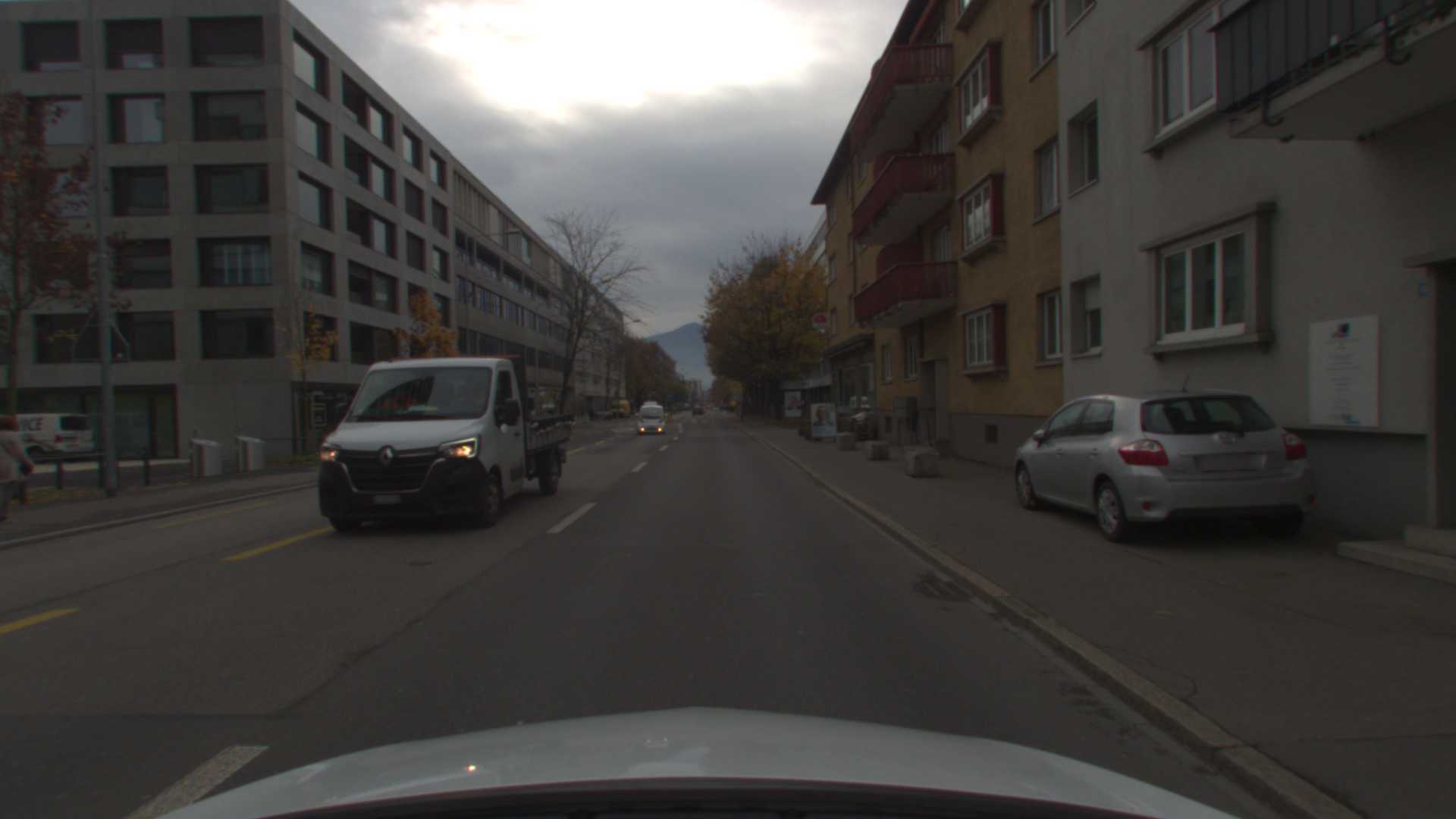} & 
\includegraphics[width=0.15\textwidth]{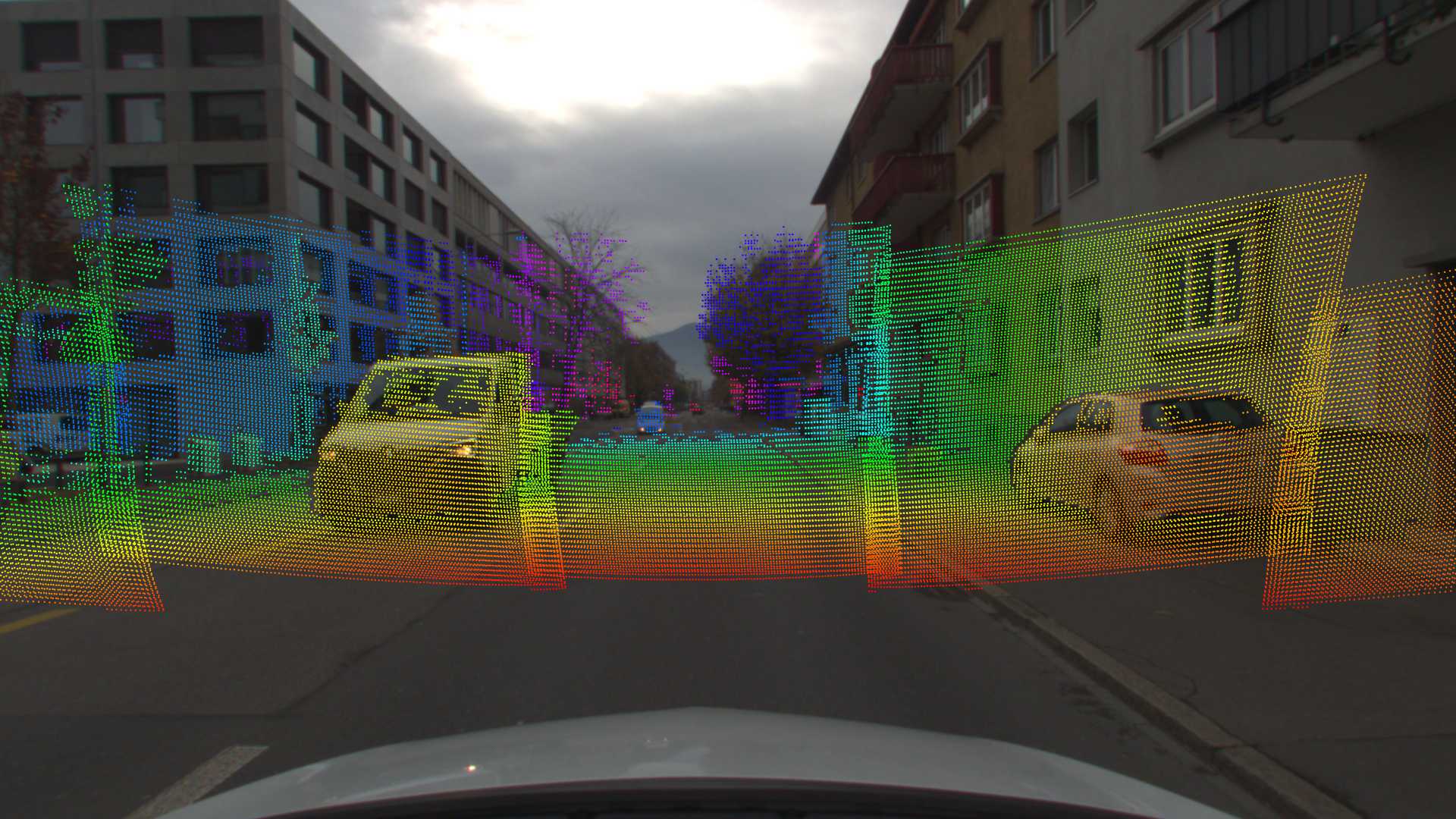} & 
\includegraphics[width=0.15\textwidth]{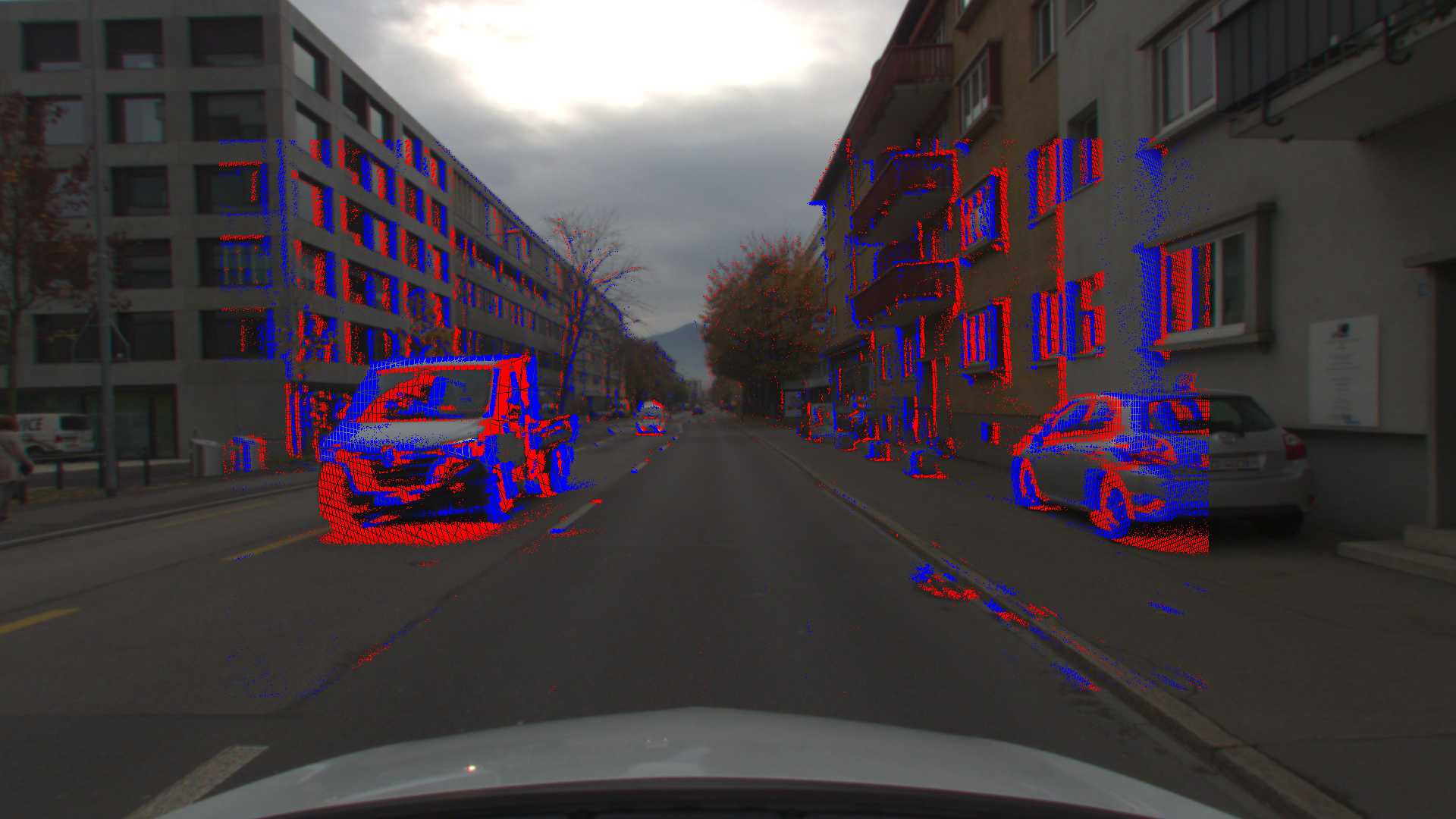} & 
\includegraphics[angle=90, trim=0 6835 0 0, clip,width=0.15\textwidth]{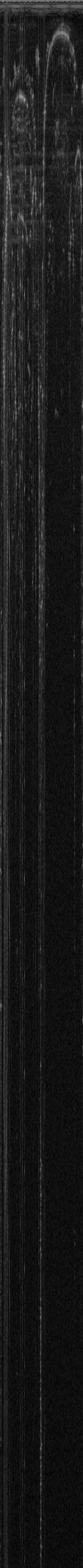} & 
\includegraphics[width=0.15\textwidth]{img/muses/REC0008_frame_080289/REC0008_frame_080289.jpg} & 
\includegraphics[width=0.15\textwidth]{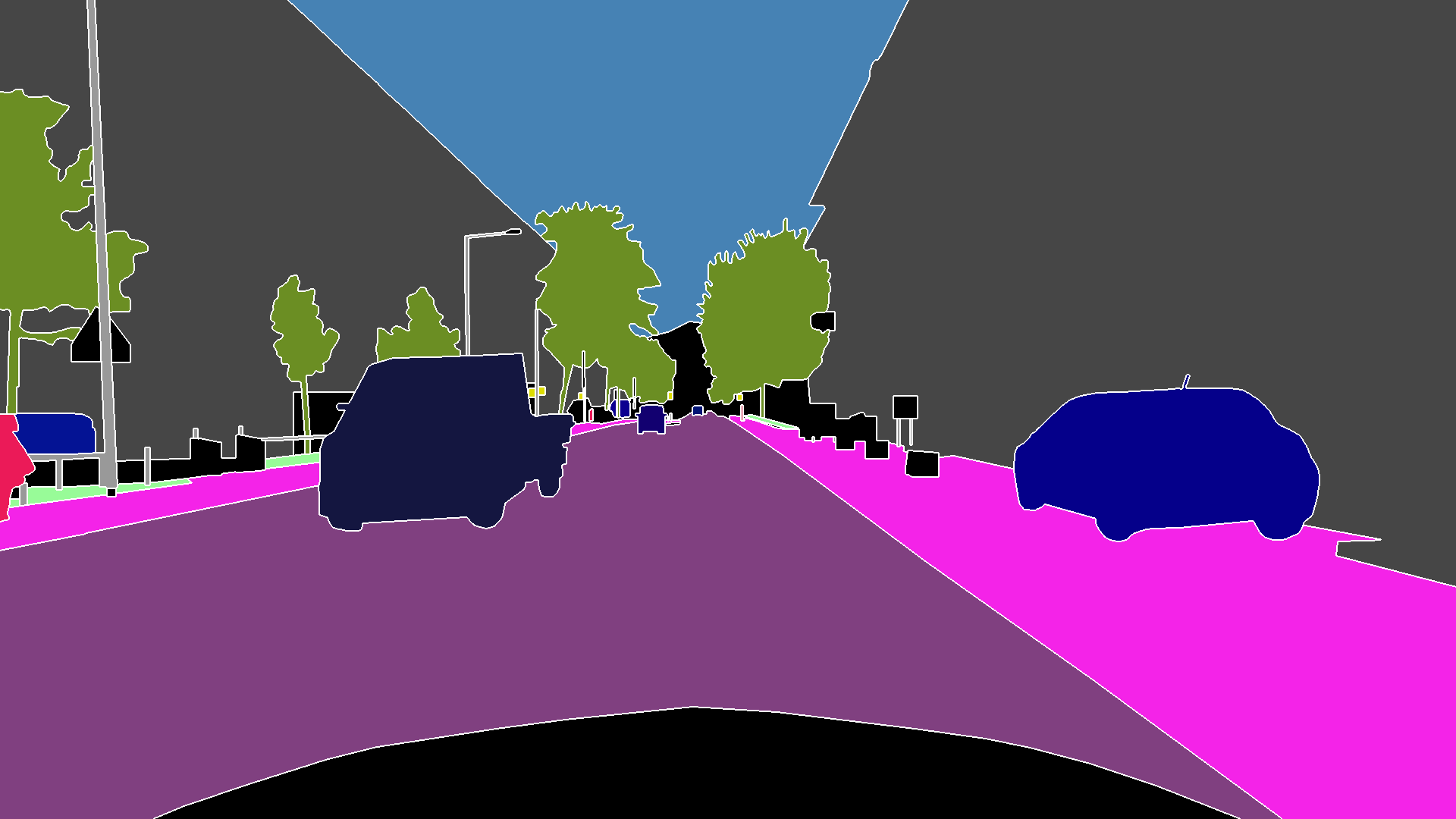} \\

\includegraphics[width=0.15\textwidth]{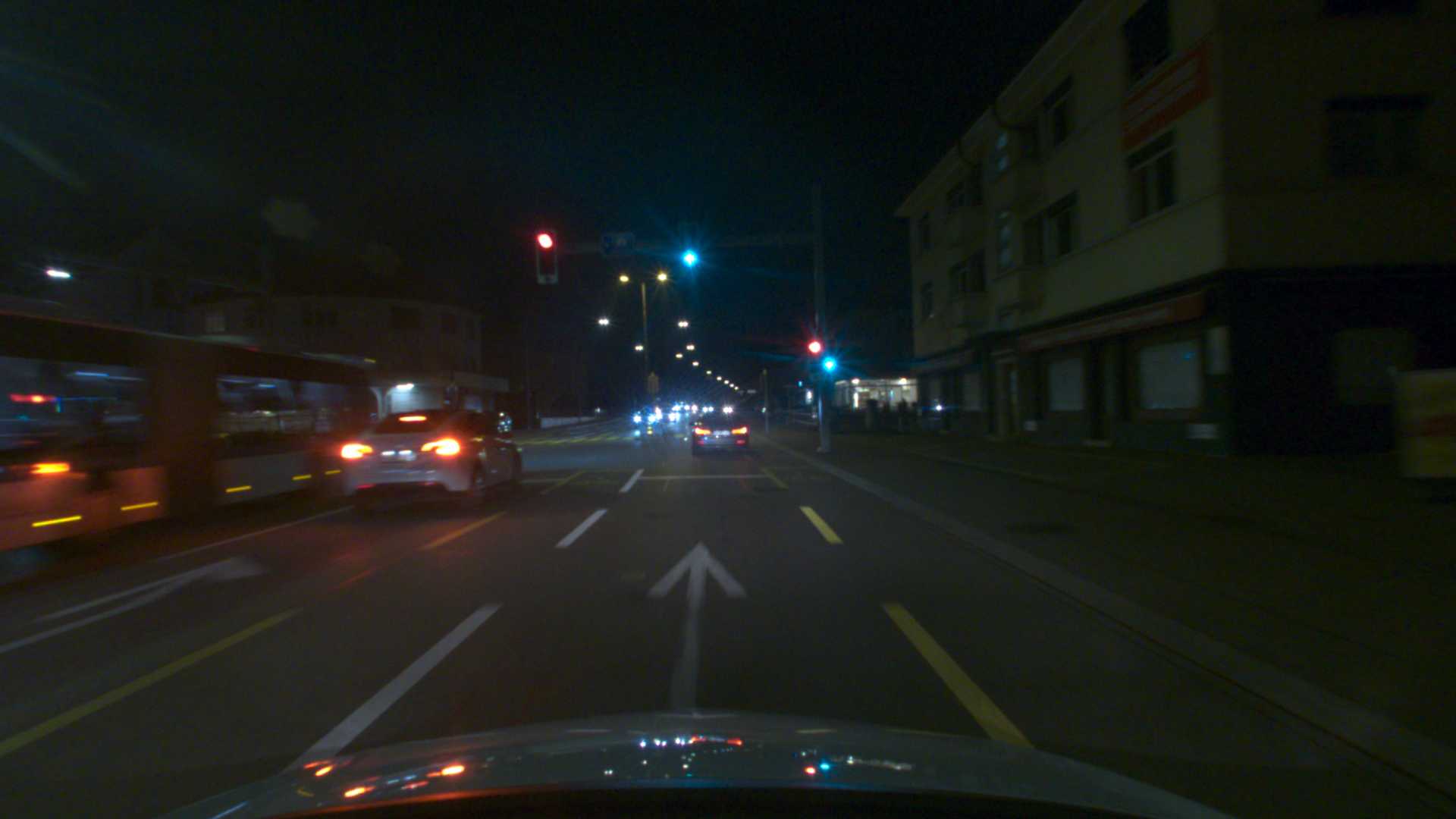} & 
\includegraphics[width=0.15\textwidth]{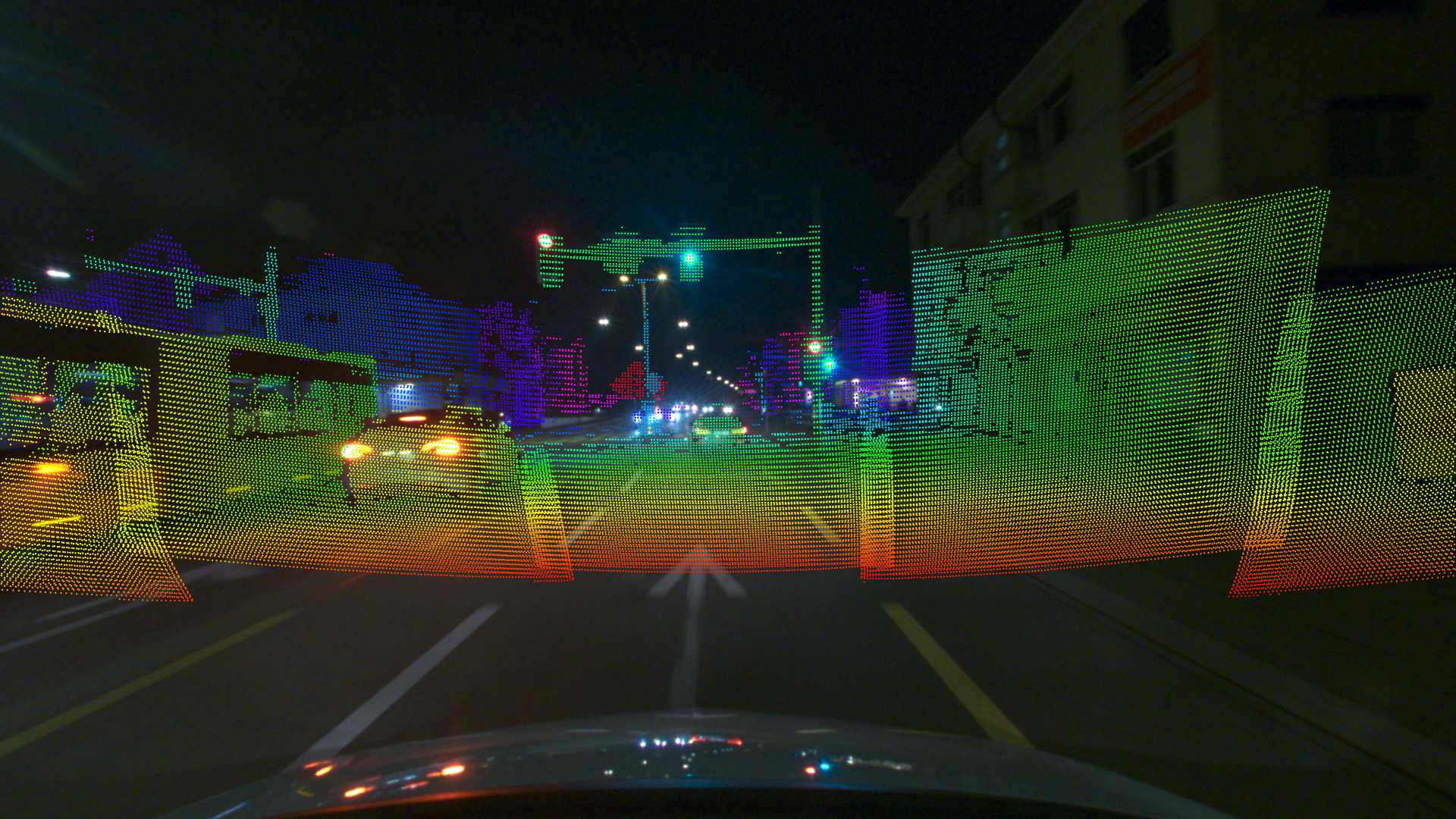} & 
\includegraphics[width=0.15\textwidth]{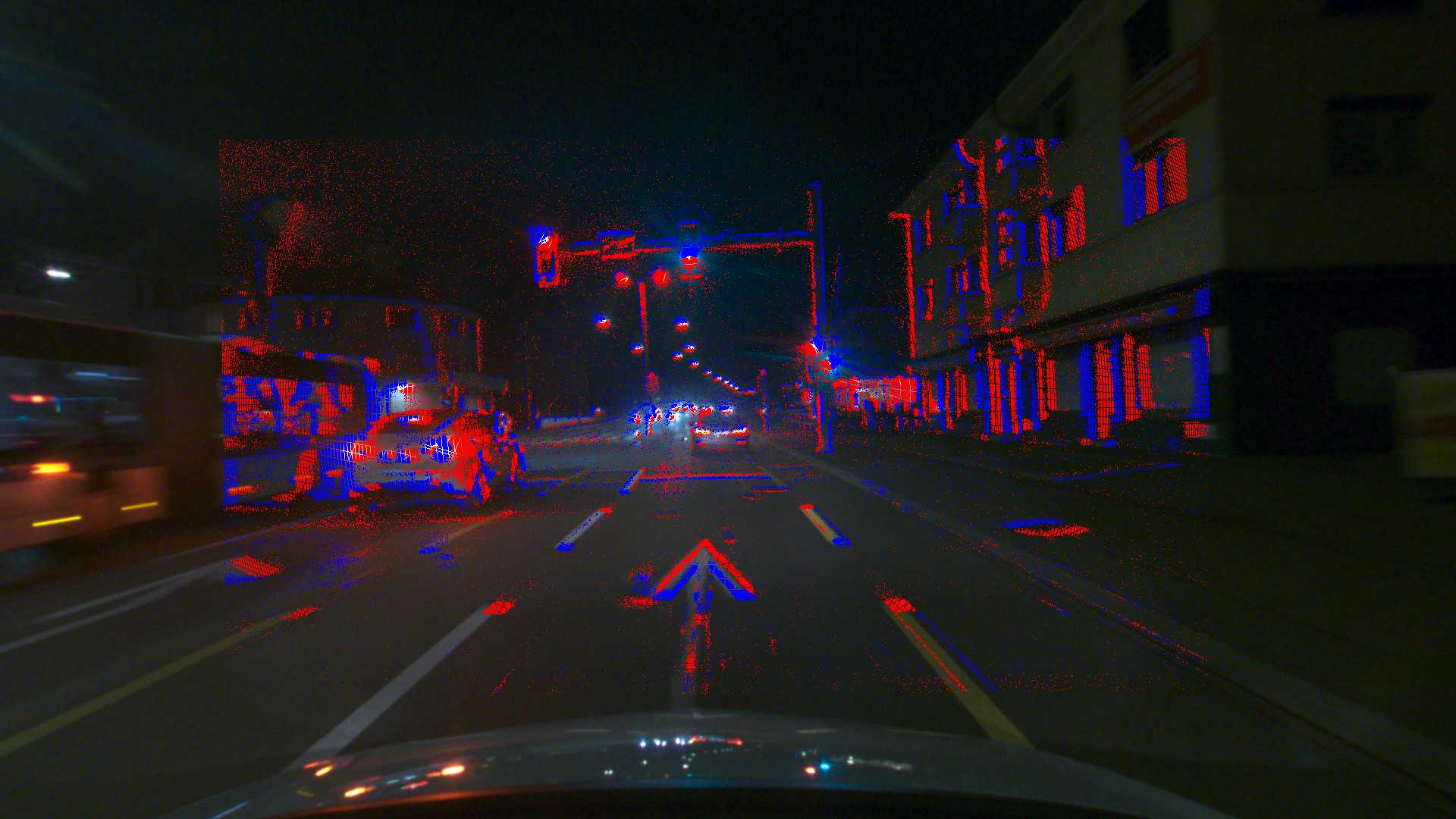} & 
\includegraphics[angle=90, trim=0 6835 0 0, clip,width=0.15\textwidth]{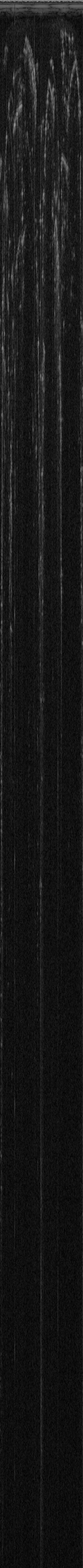} & 
\includegraphics[width=0.15\textwidth]{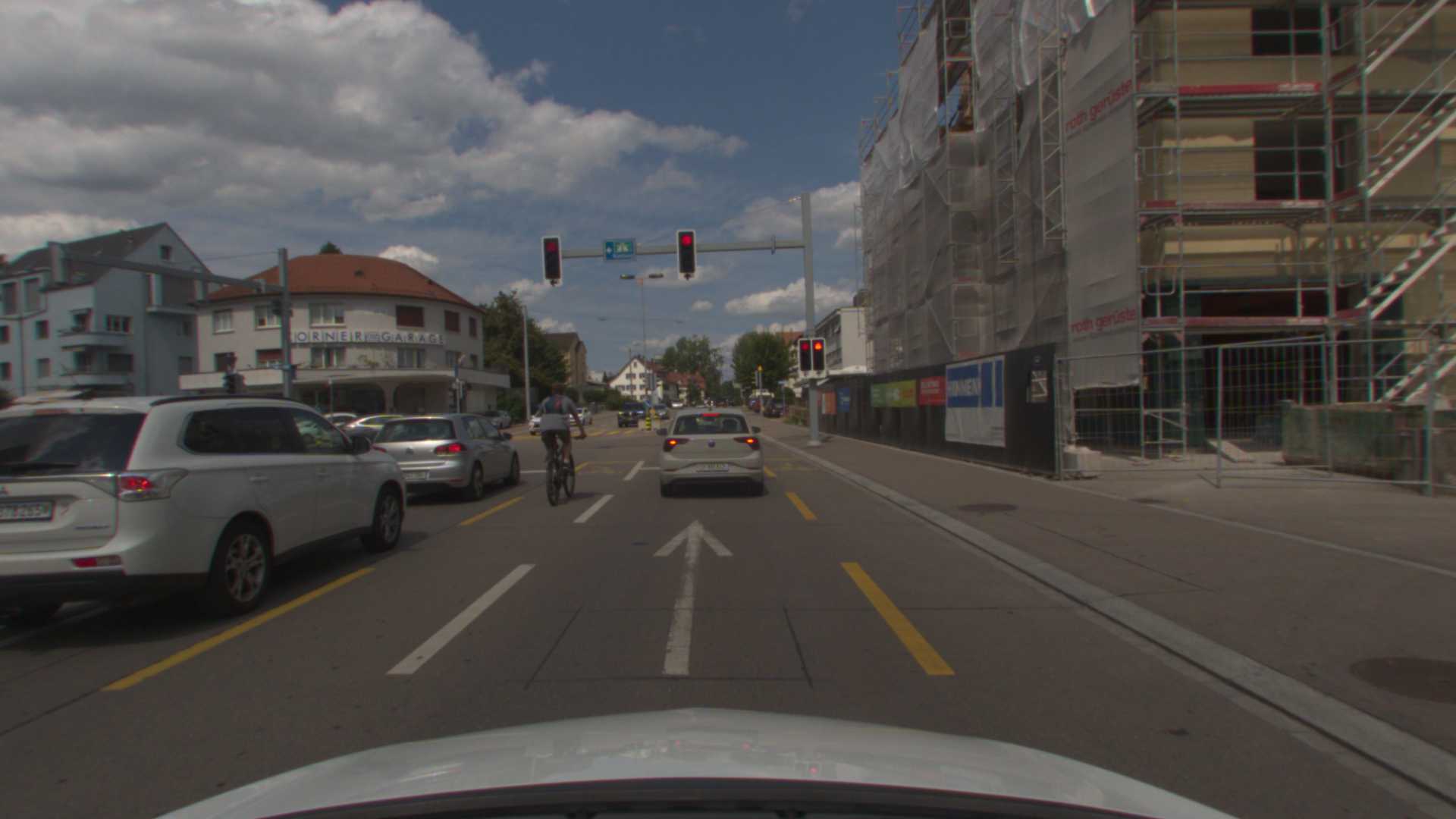} & 
\includegraphics[width=0.15\textwidth]{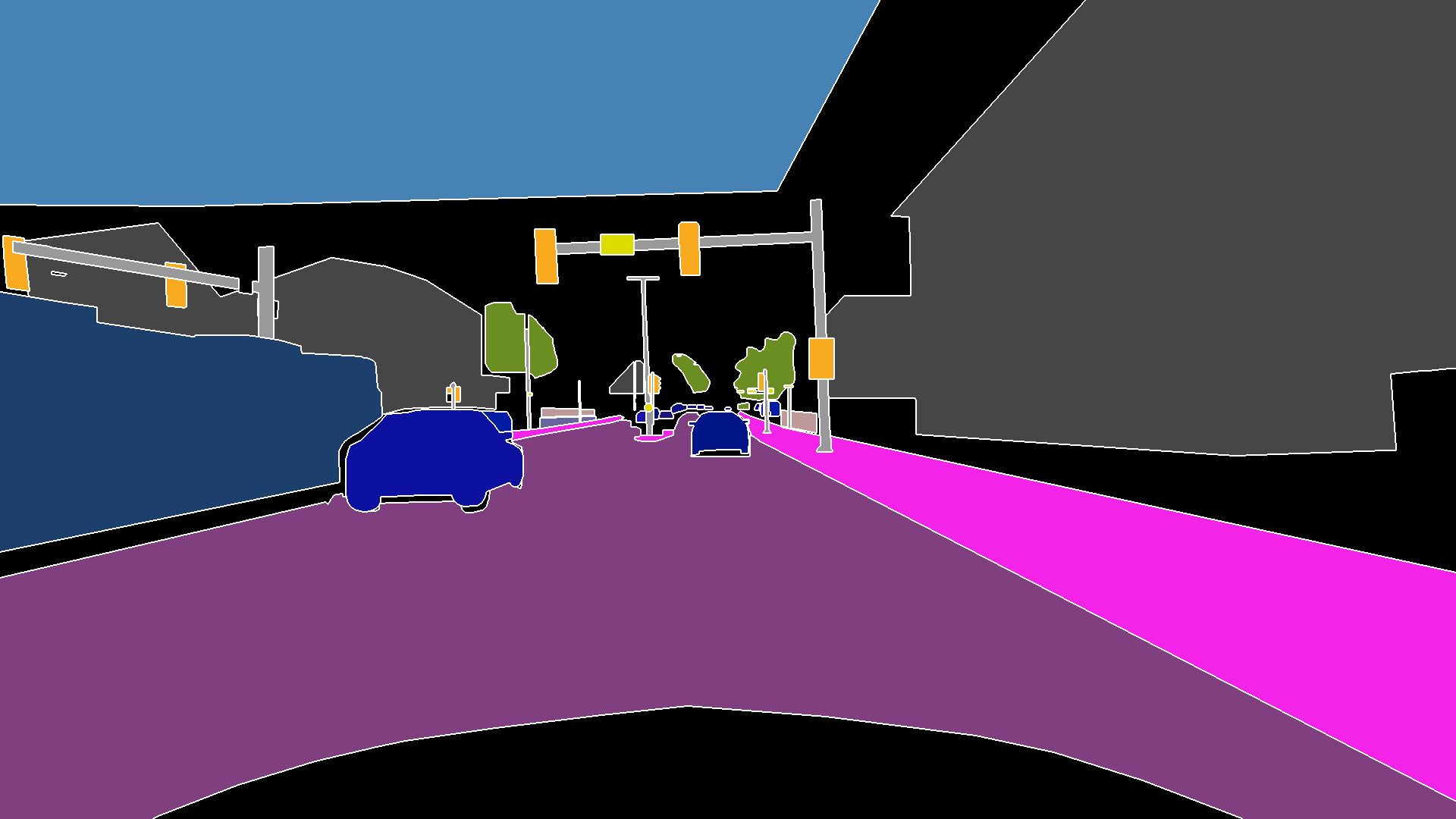} \\

\includegraphics[width=0.15\textwidth]{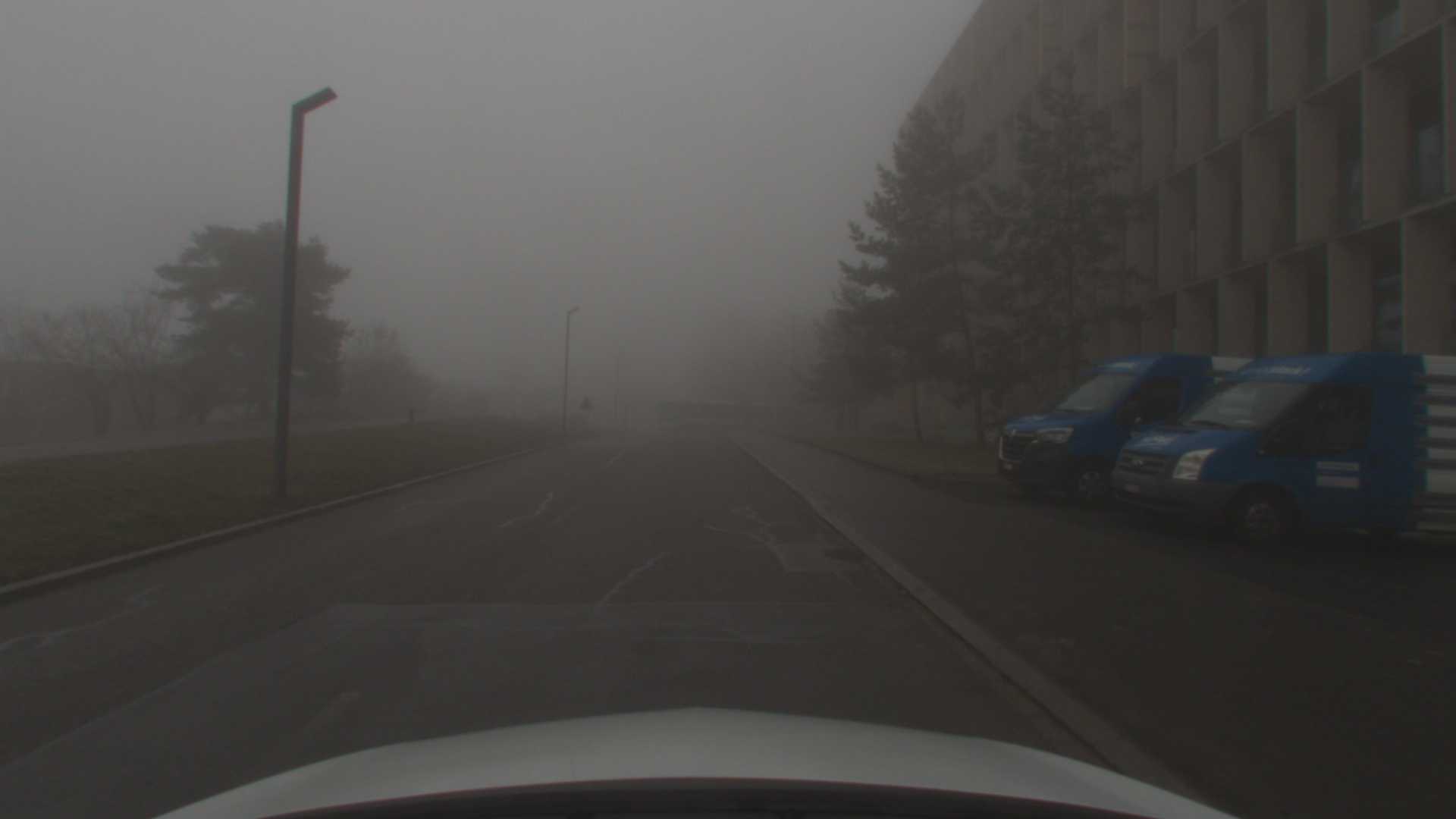} & 
\includegraphics[width=0.15\textwidth]{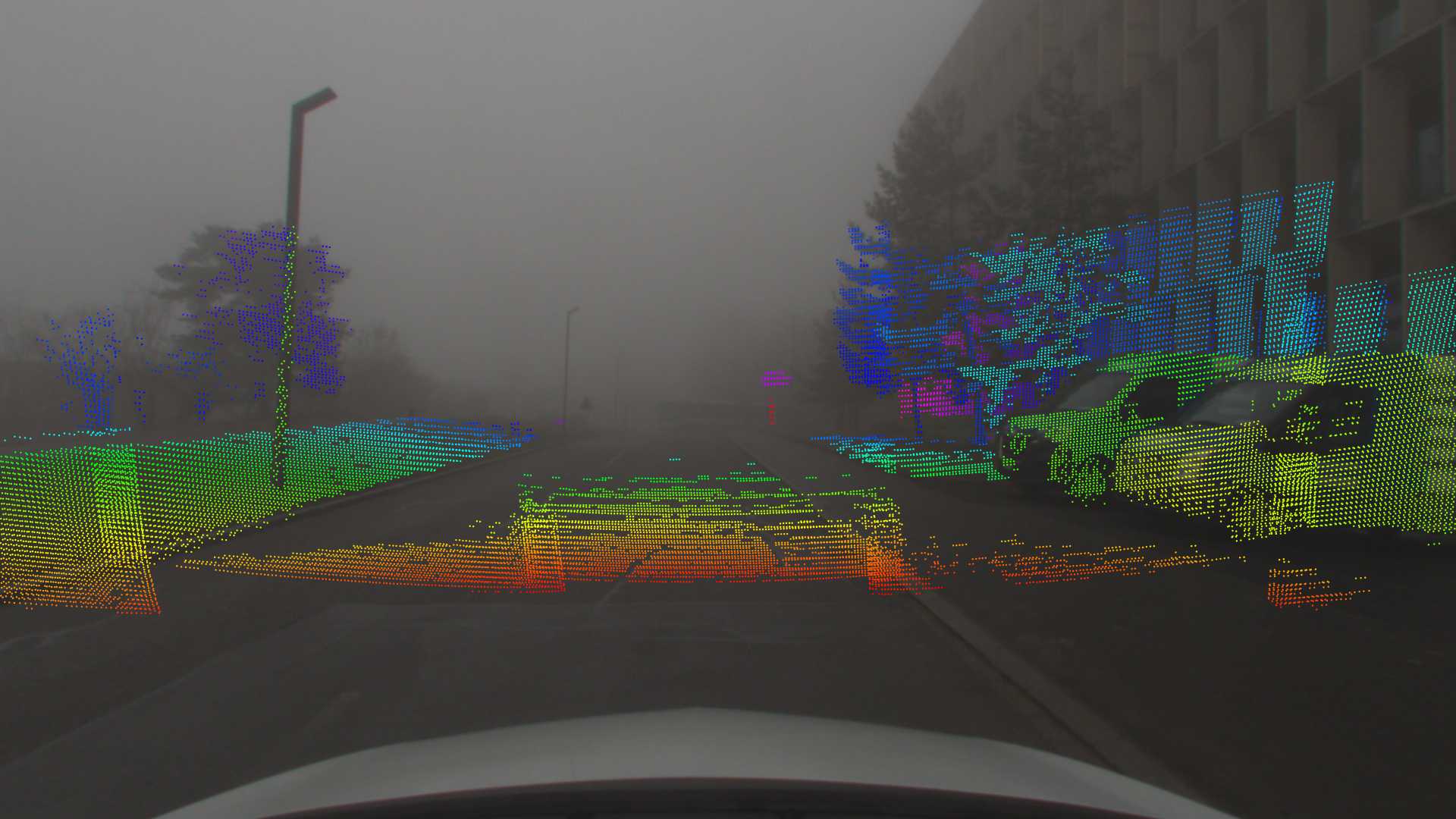} & 
\includegraphics[width=0.15\textwidth]{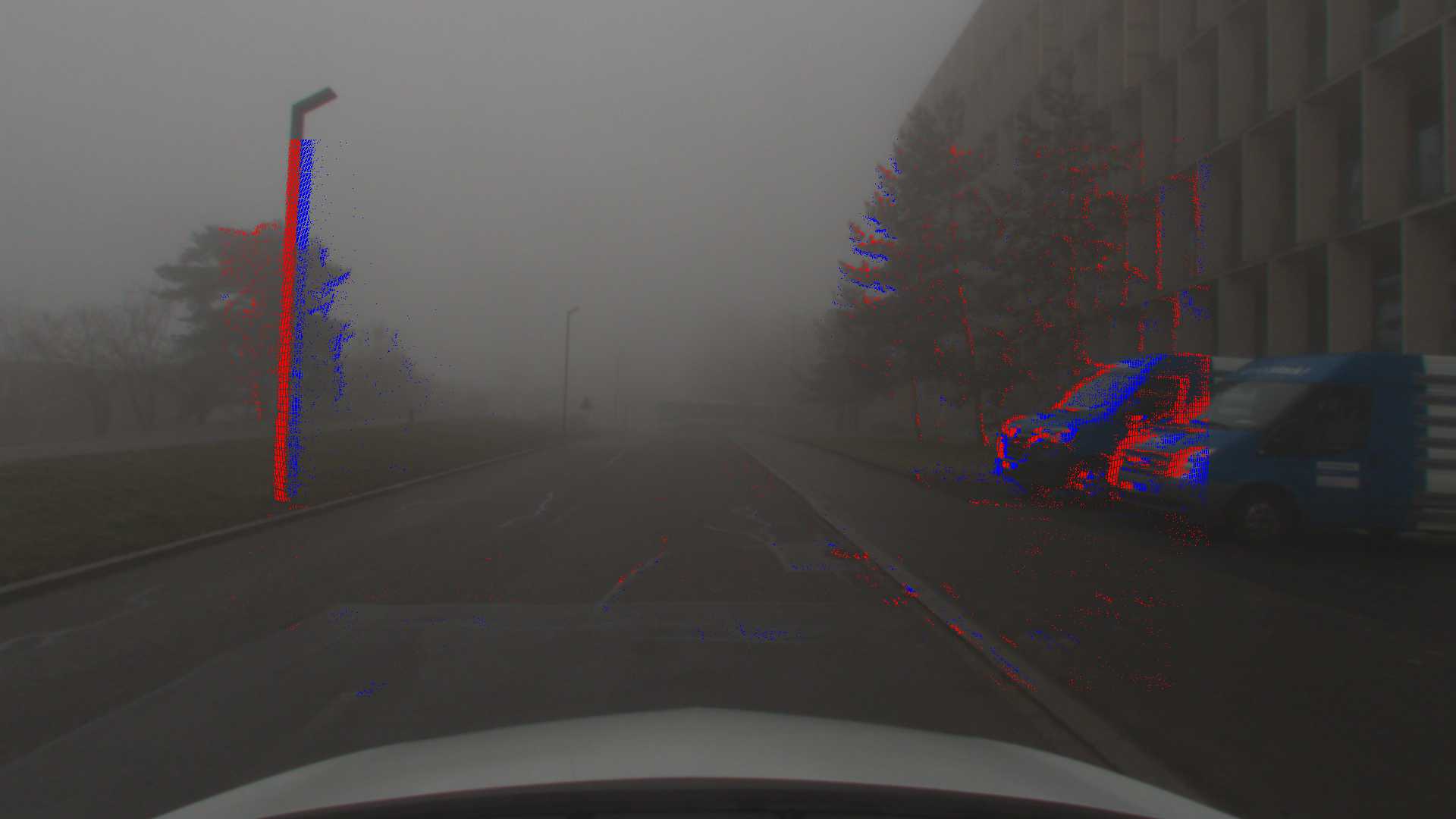} & 
\includegraphics[angle=90, trim=0 6835 0 0, clip,width=0.15\textwidth]{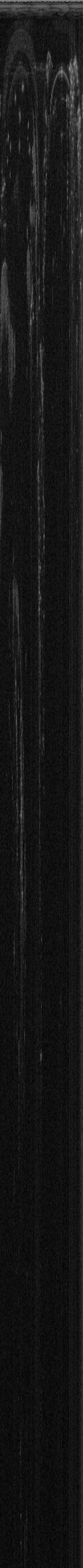} & 
\includegraphics[width=0.15\textwidth]{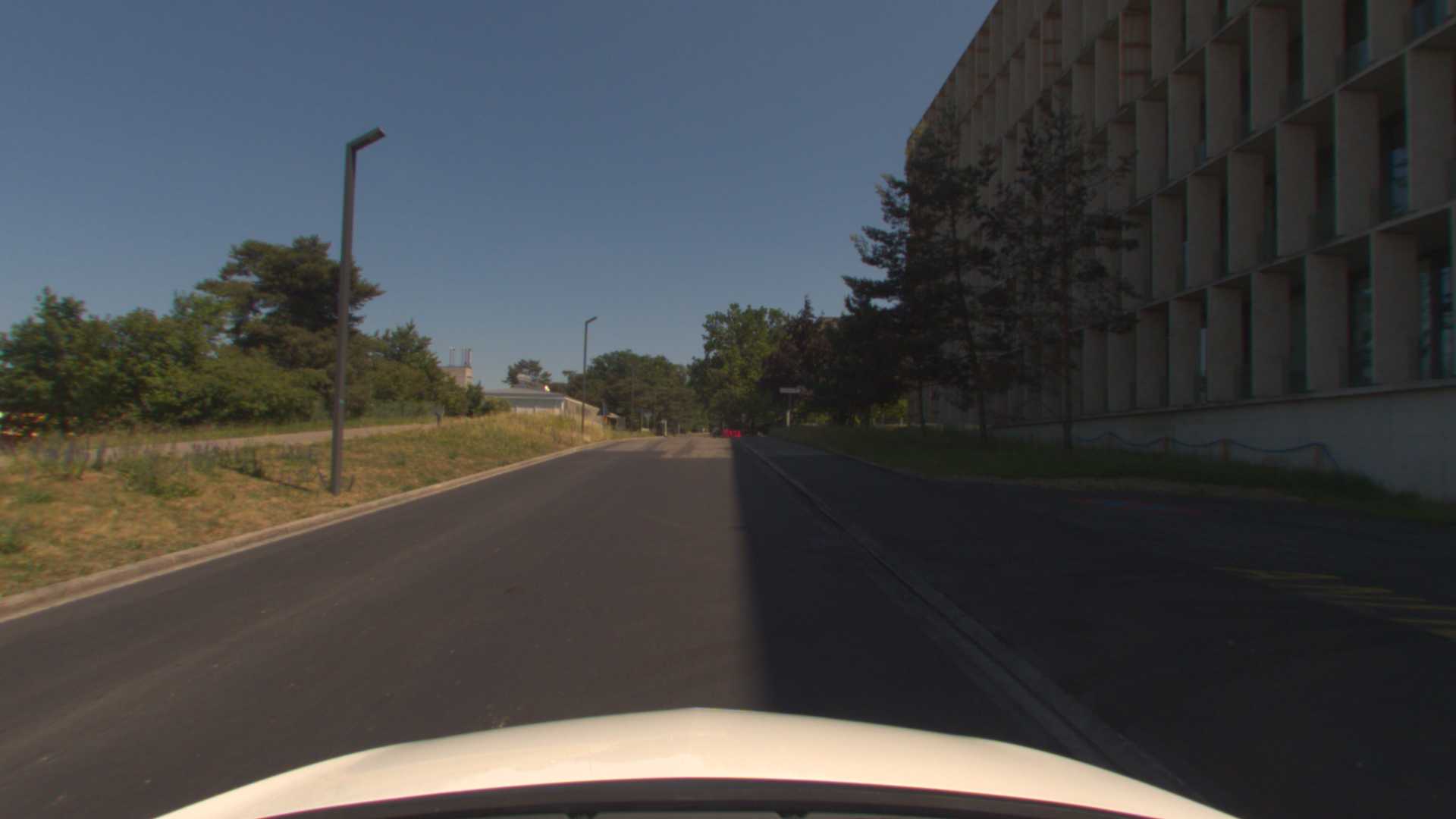} & 
\includegraphics[width=0.15\textwidth]{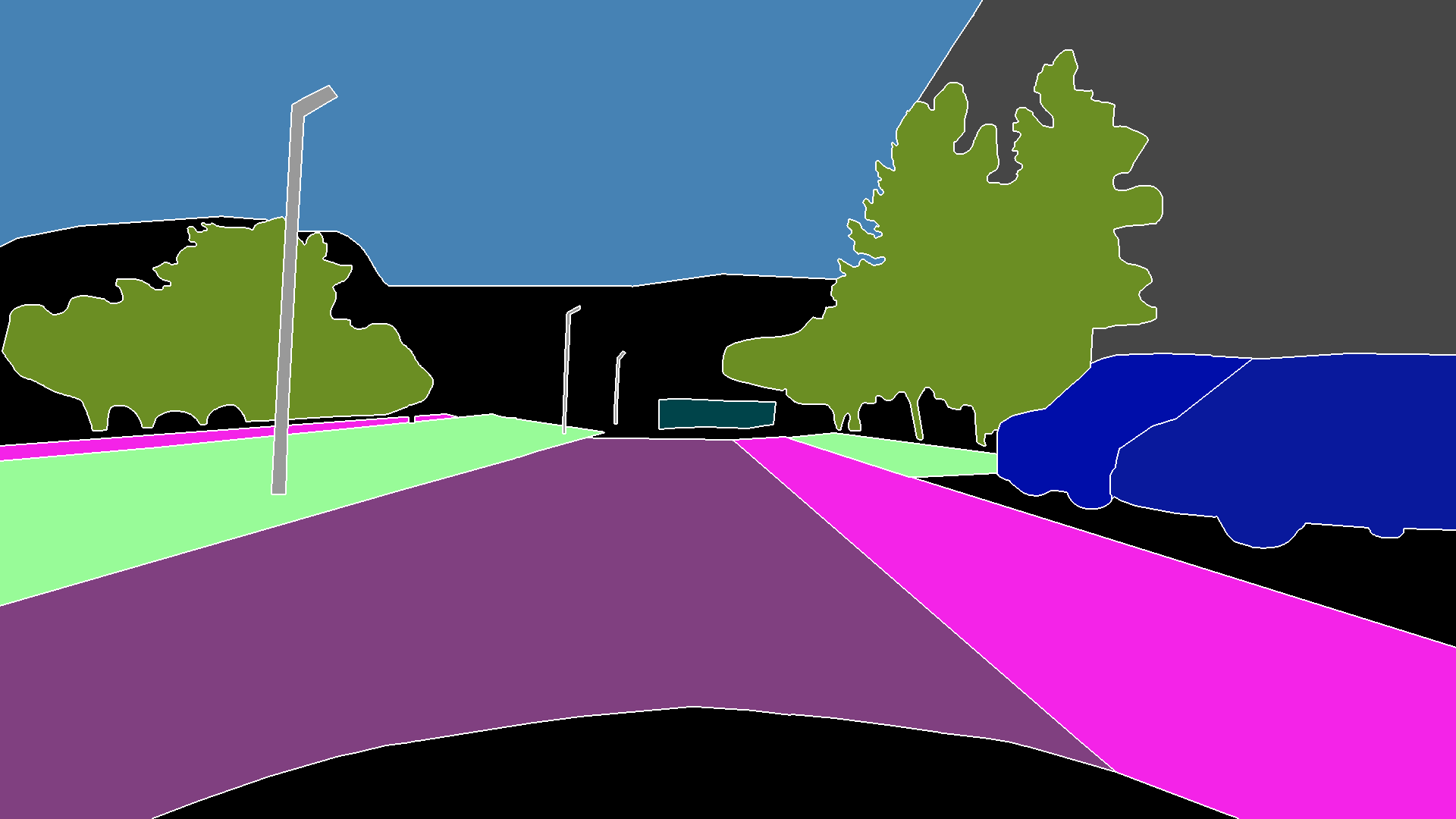} \\

\includegraphics[width=0.15\textwidth]{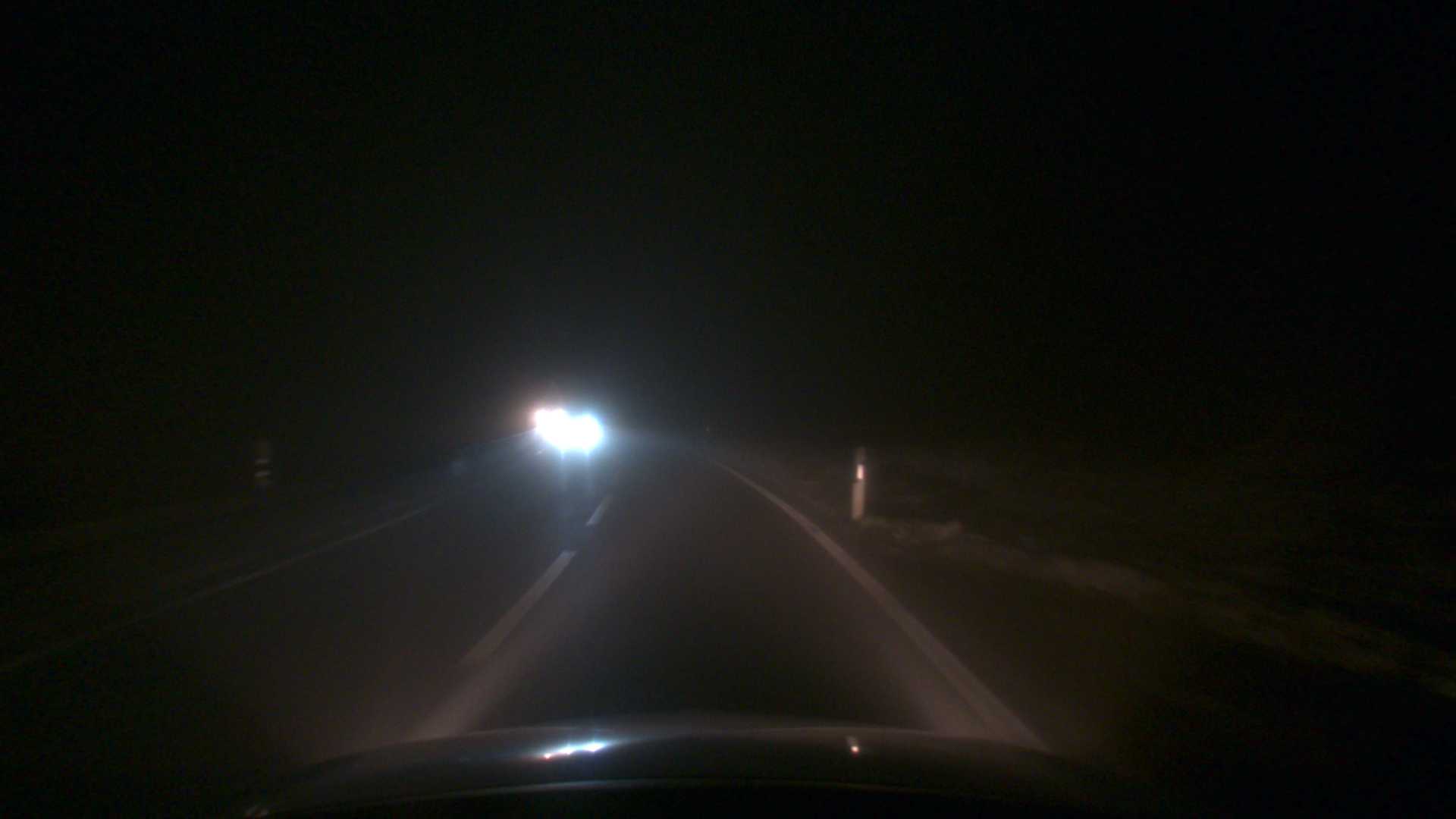} & 
\includegraphics[width=0.15\textwidth]{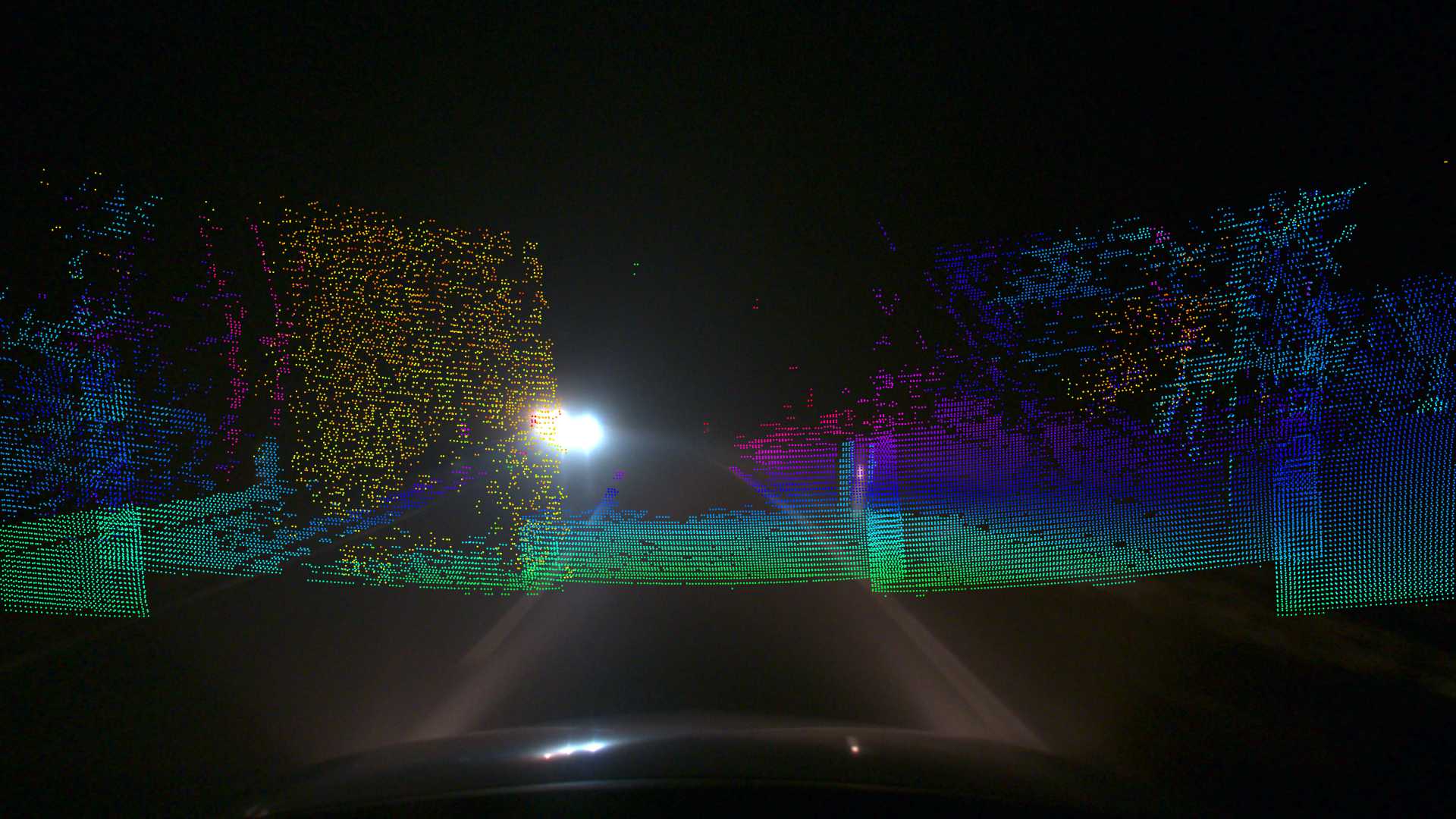} & 
\includegraphics[width=0.15\textwidth]{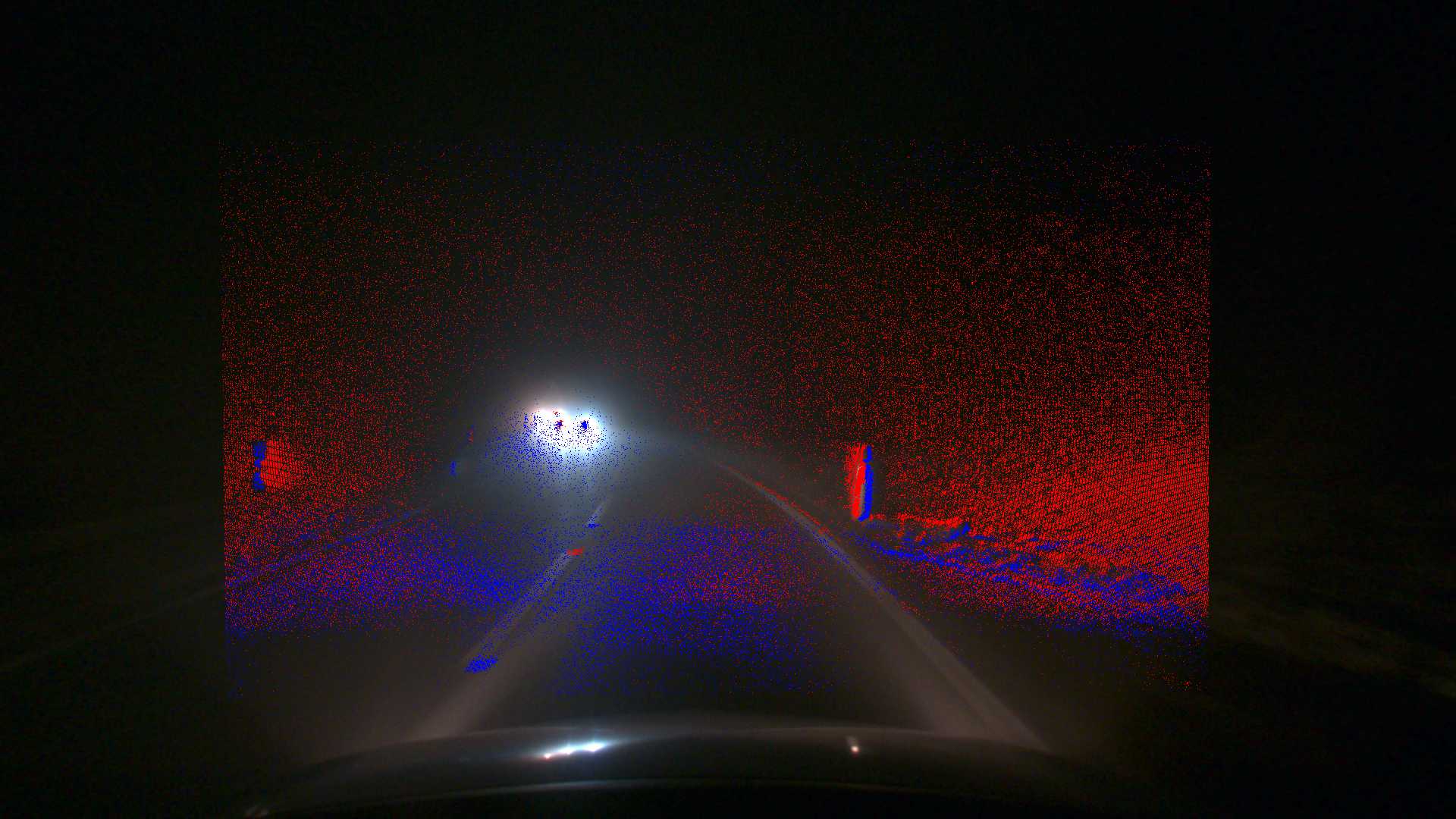} & 
\includegraphics[angle=90, trim=0 6835 0 0, clip,width=0.15\textwidth]{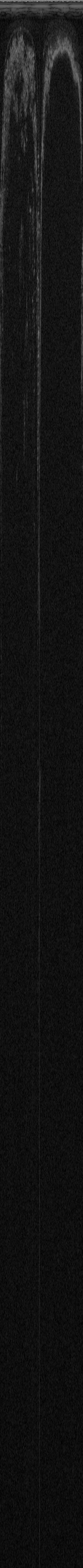} & 
\includegraphics[width=0.15\textwidth]{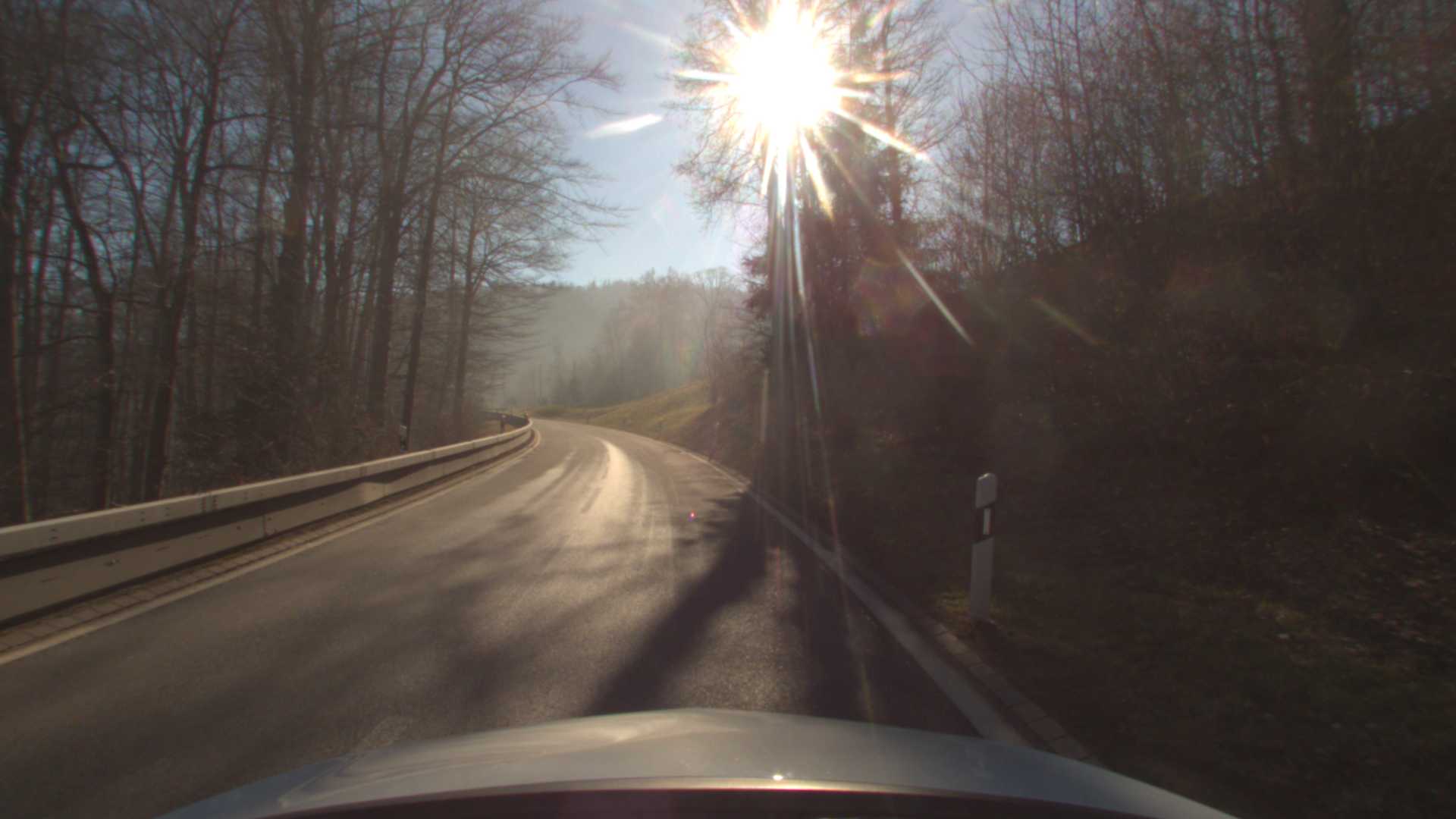} & 
\includegraphics[width=0.15\textwidth]{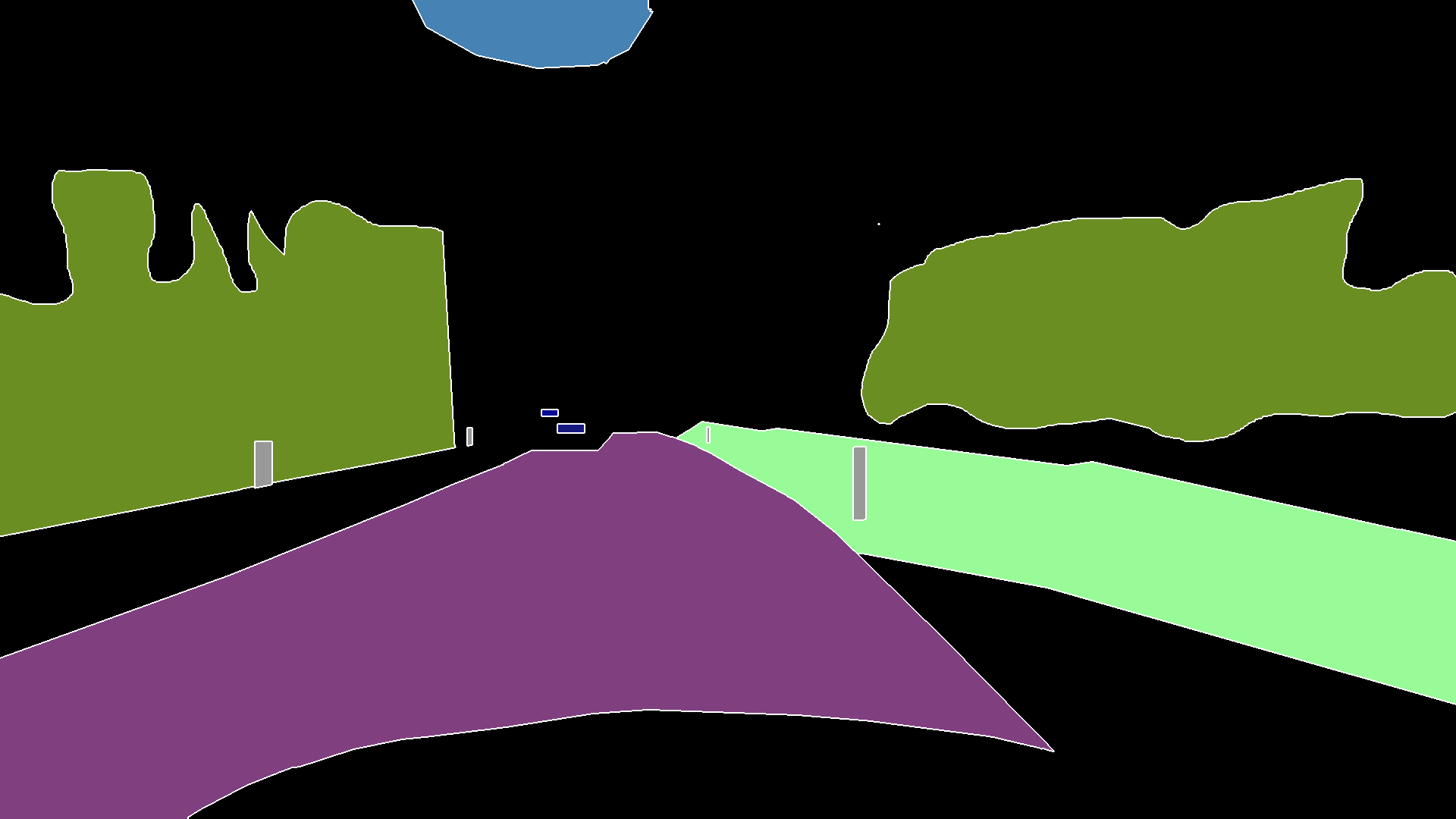} \\

\includegraphics[width=0.15\textwidth]{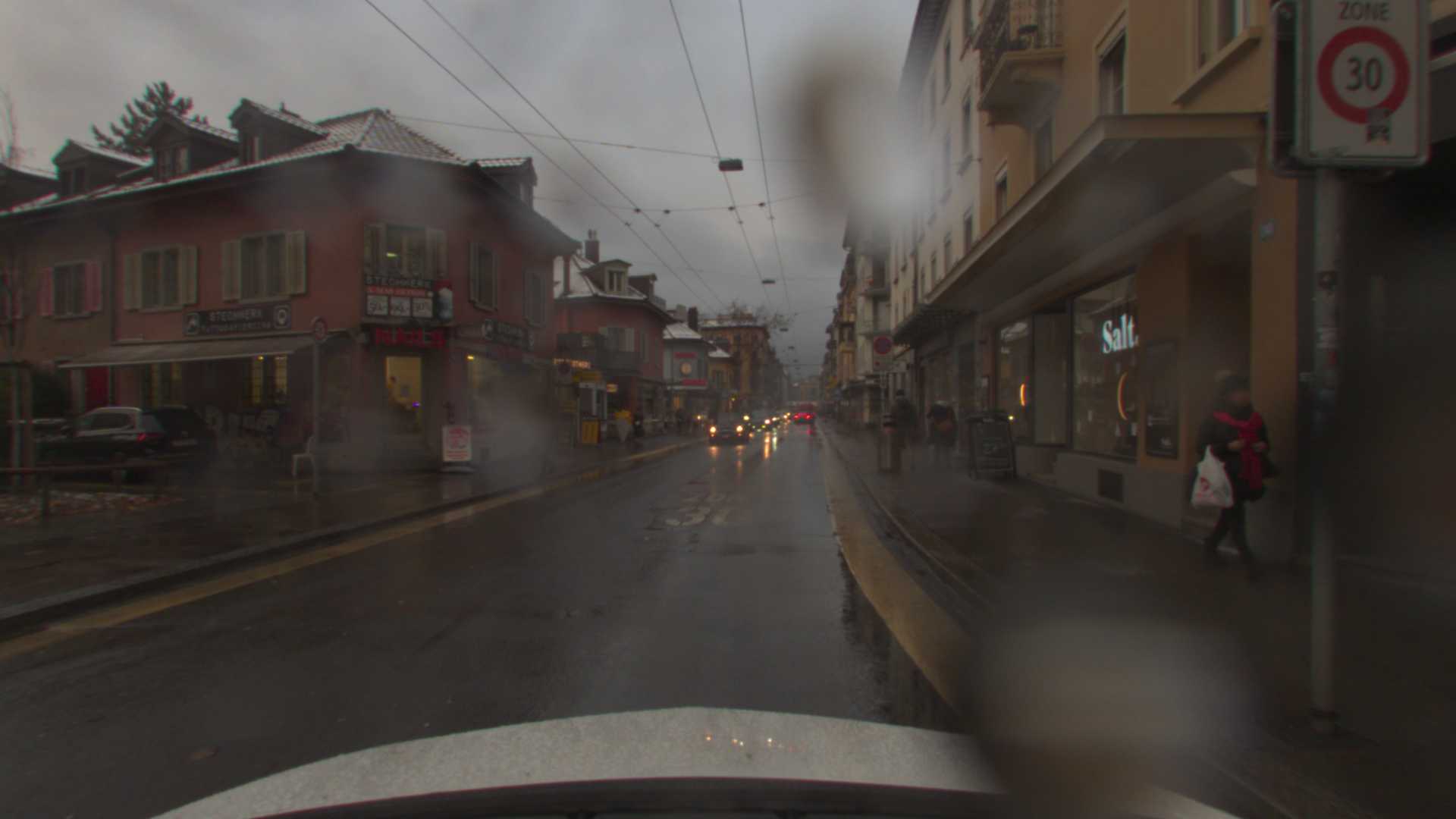} & 
\includegraphics[width=0.15\textwidth]{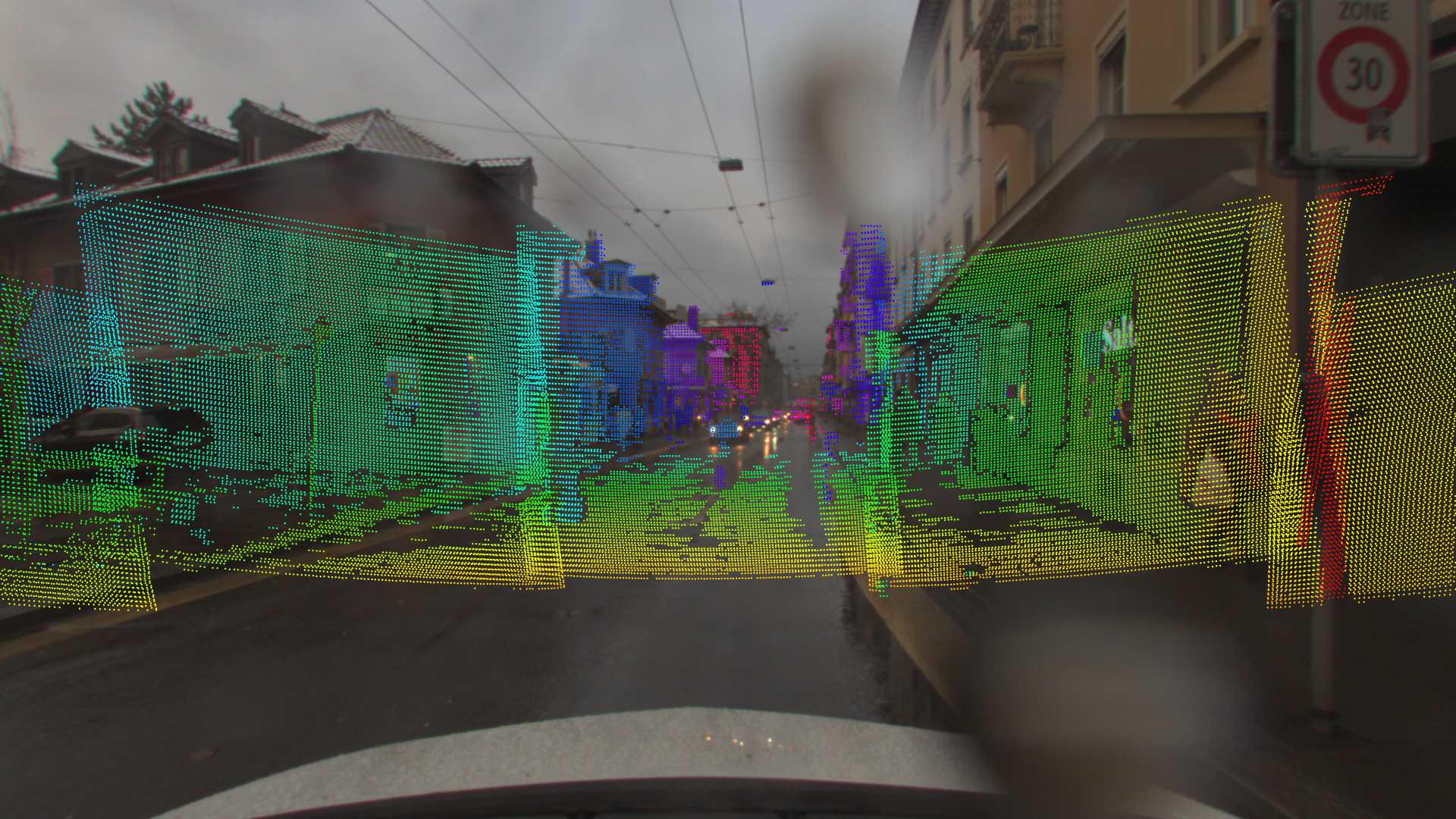} & 
\includegraphics[width=0.15\textwidth]{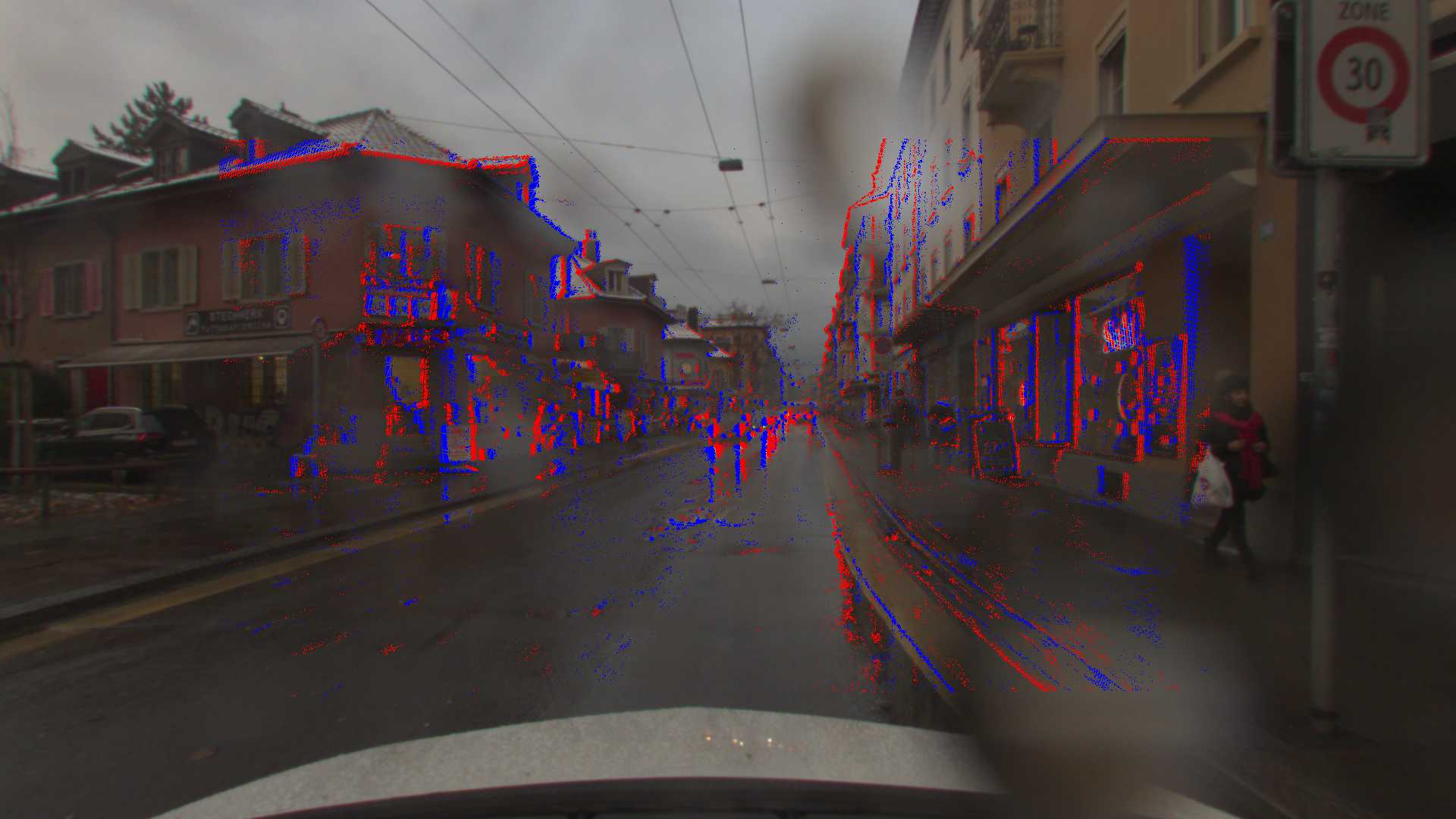} & 
\includegraphics[angle=90, trim=0 6835 0 0, clip,width=0.15\textwidth]{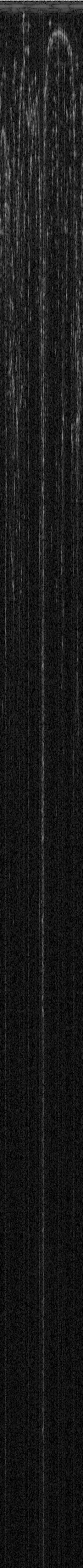} & 
\includegraphics[width=0.15\textwidth]{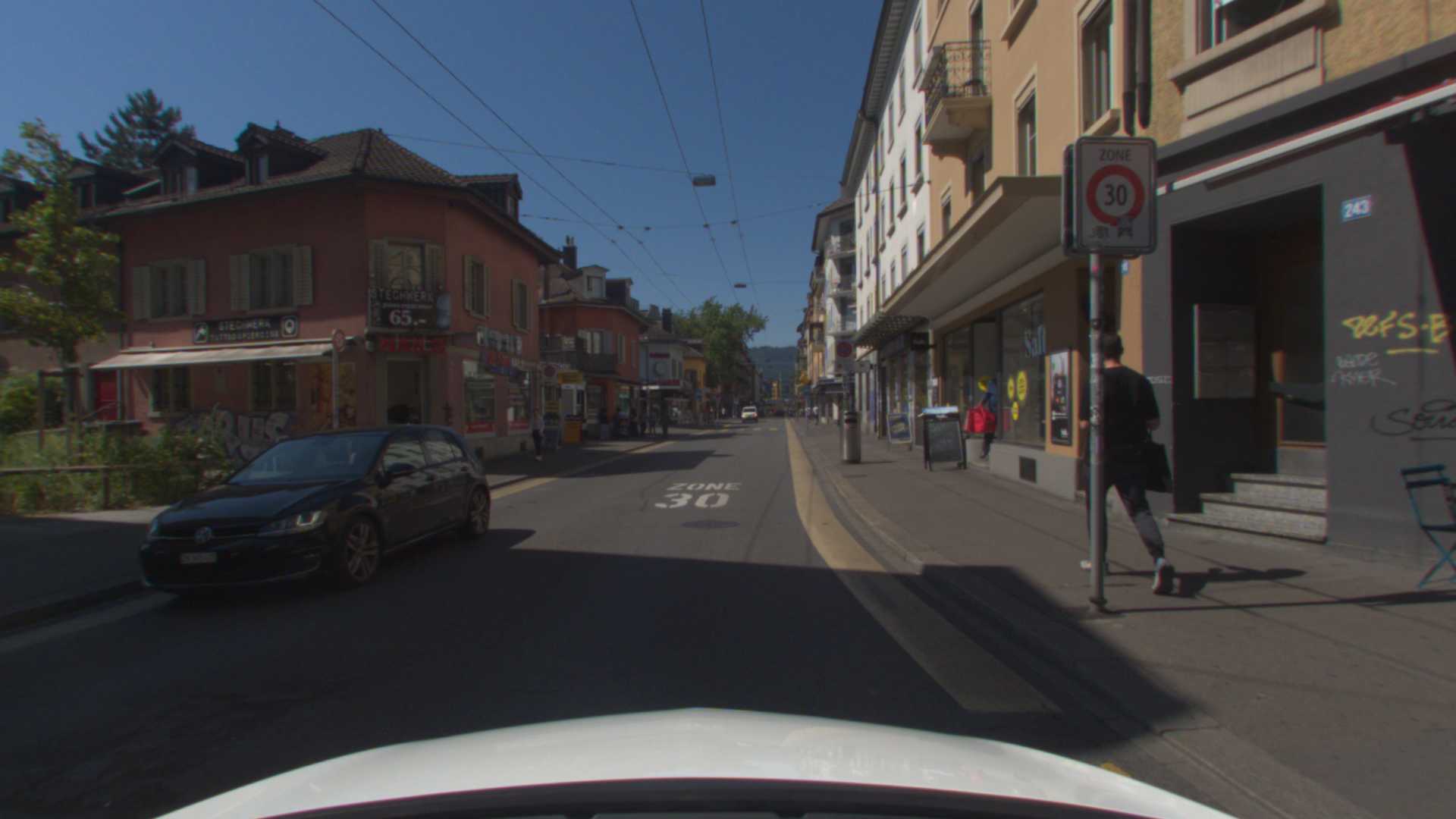} & 
\includegraphics[width=0.15\textwidth]{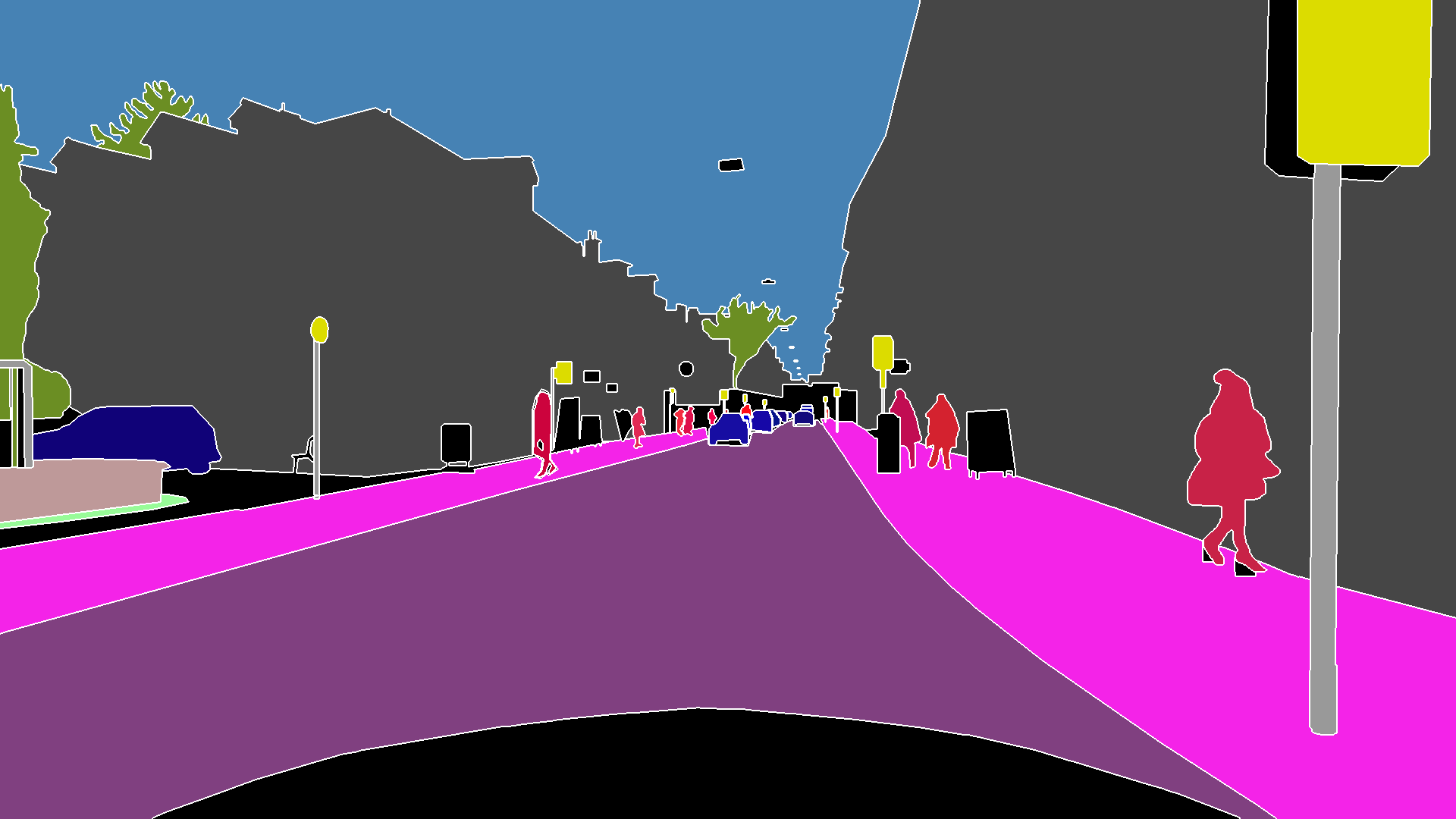} \\

\includegraphics[width=0.15\textwidth]{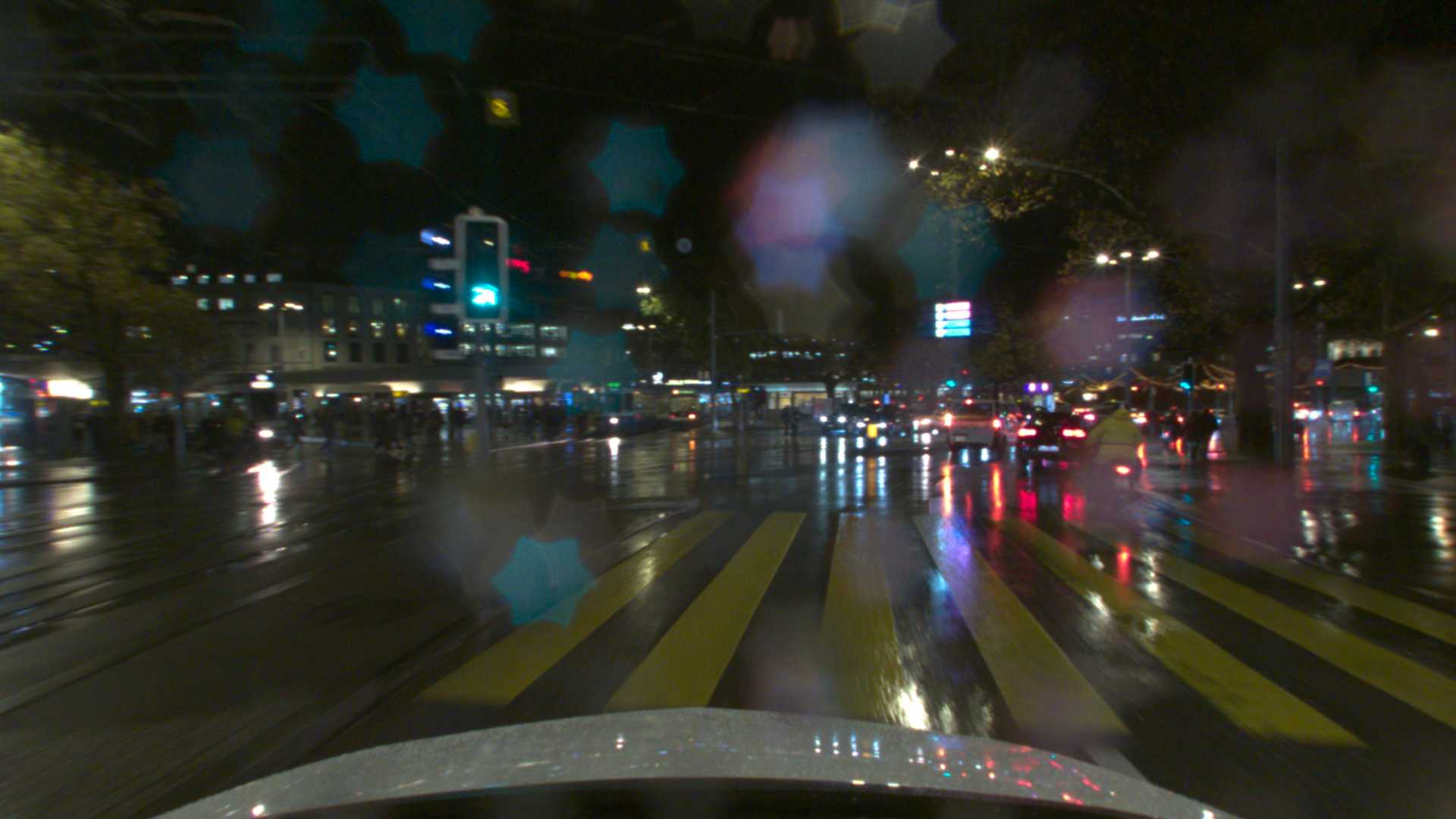} & 
\includegraphics[width=0.15\textwidth]{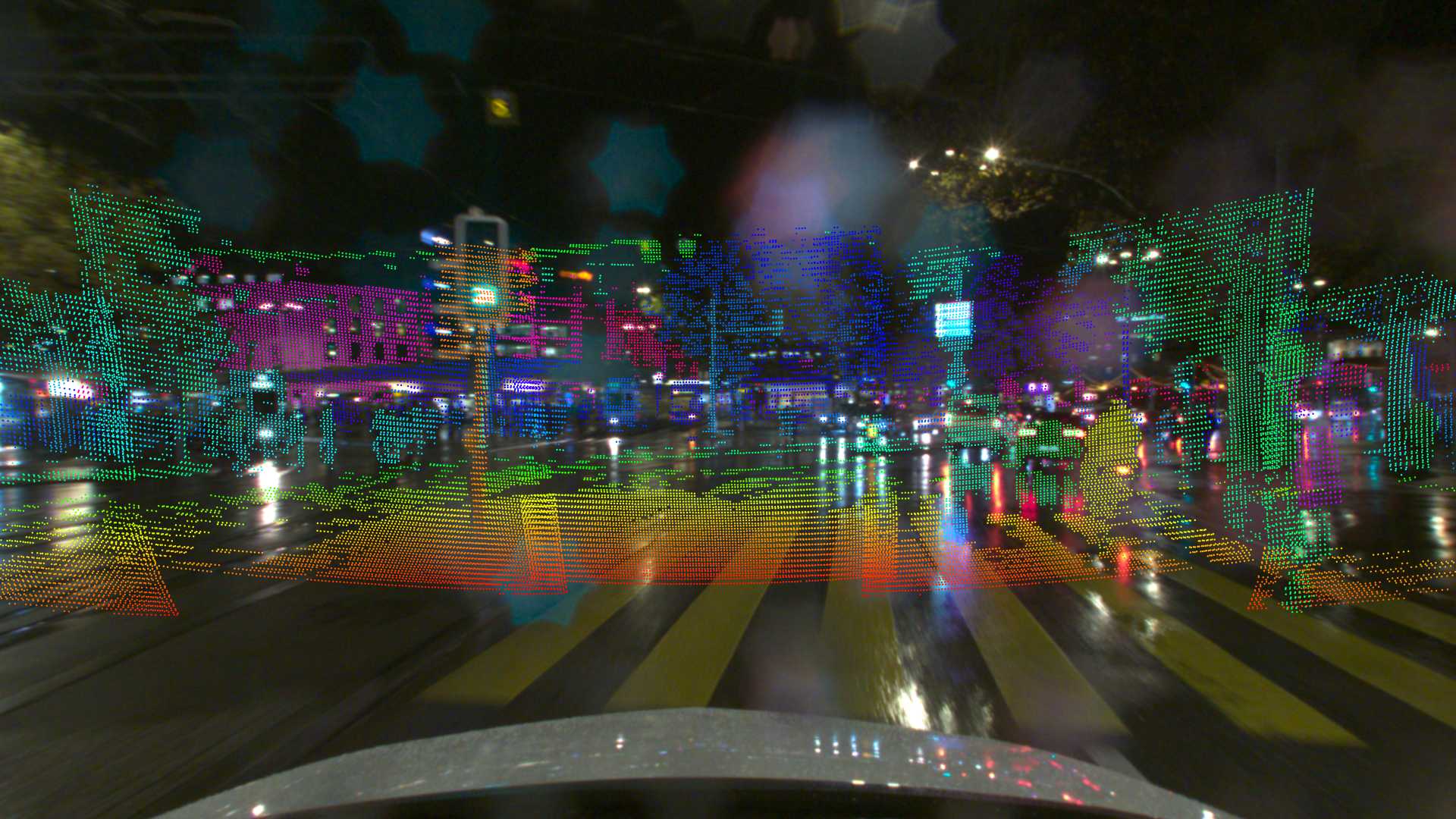} & 
\includegraphics[width=0.15\textwidth]{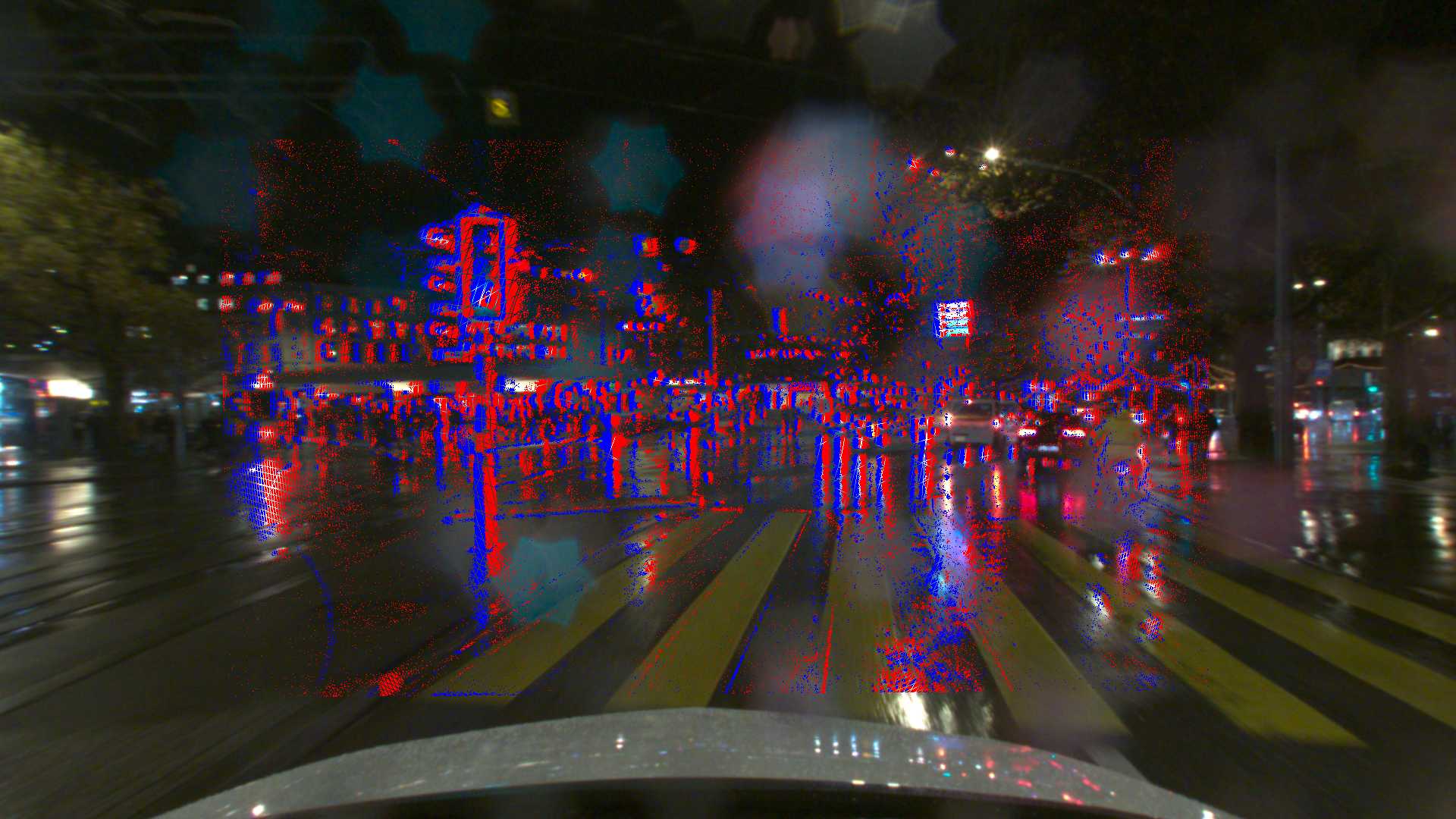} & 
\includegraphics[angle=90, trim=0 6835 0 0, clip,width=0.15\textwidth]{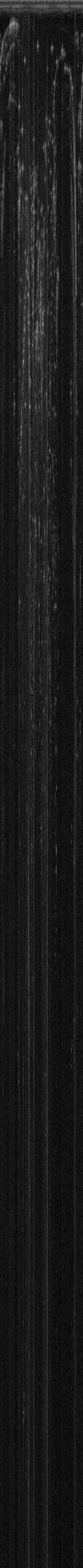} & 
\includegraphics[width=0.15\textwidth]{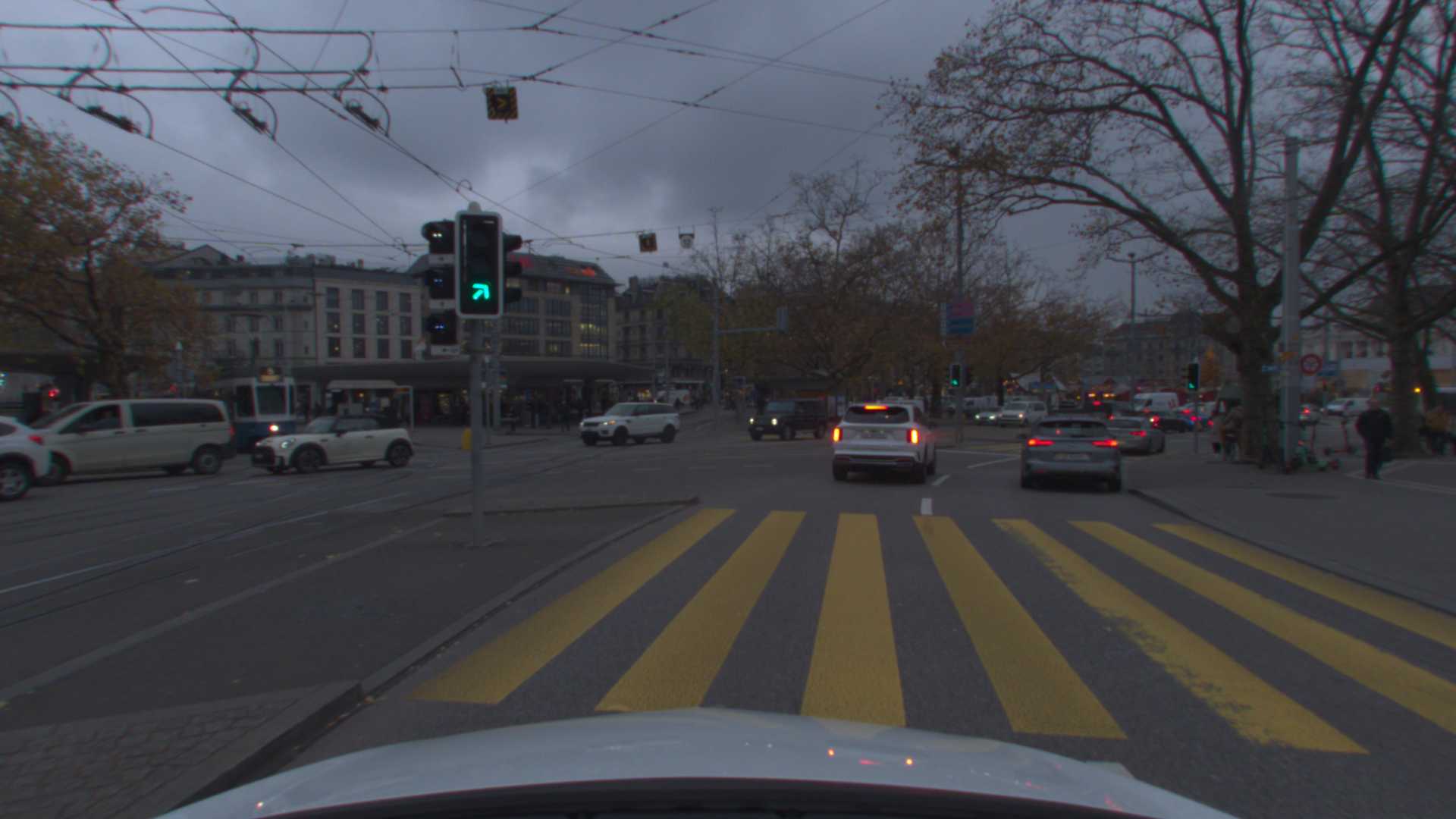} & 
\includegraphics[width=0.15\textwidth]{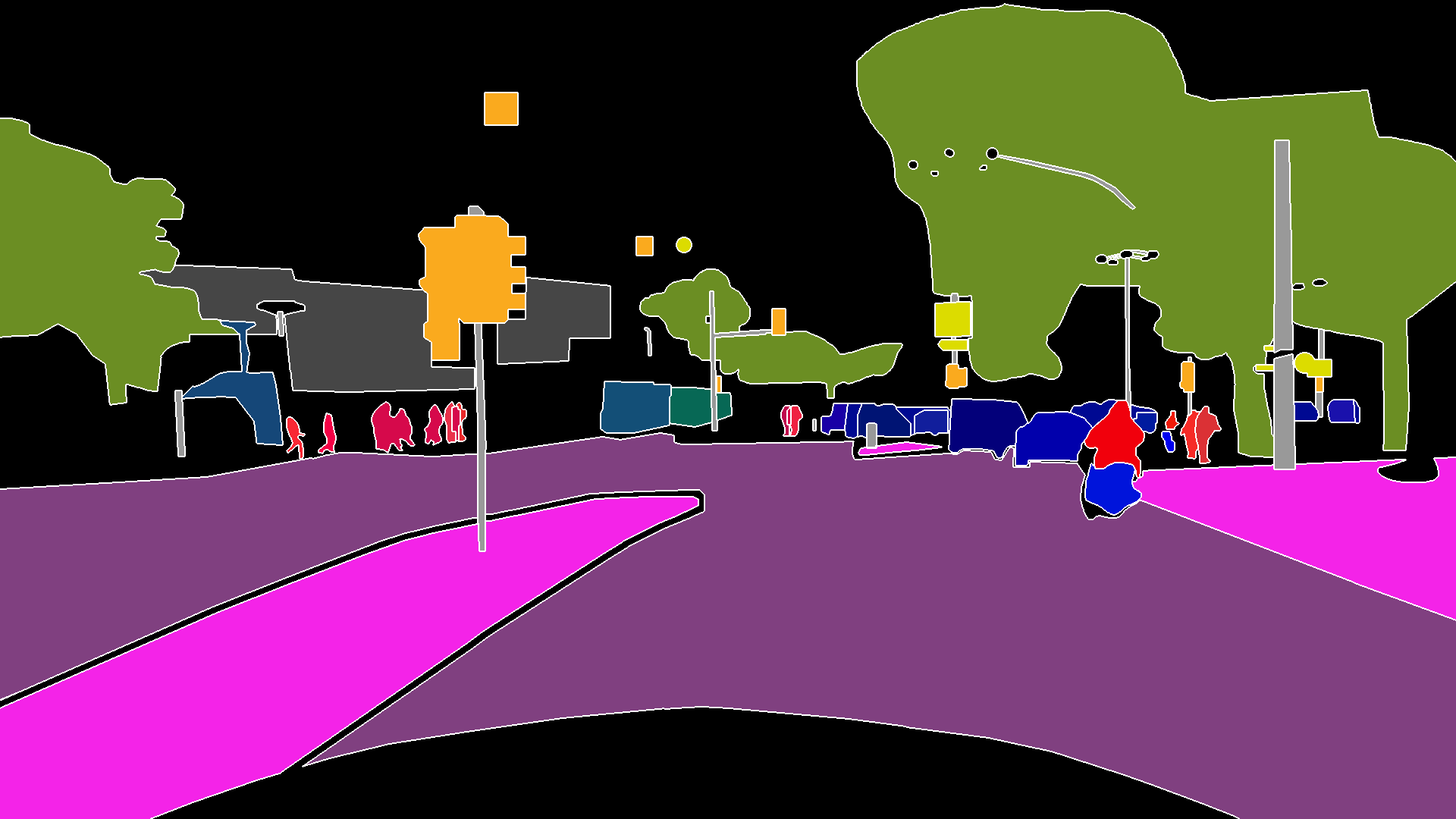} \\

\includegraphics[width=0.15\textwidth]{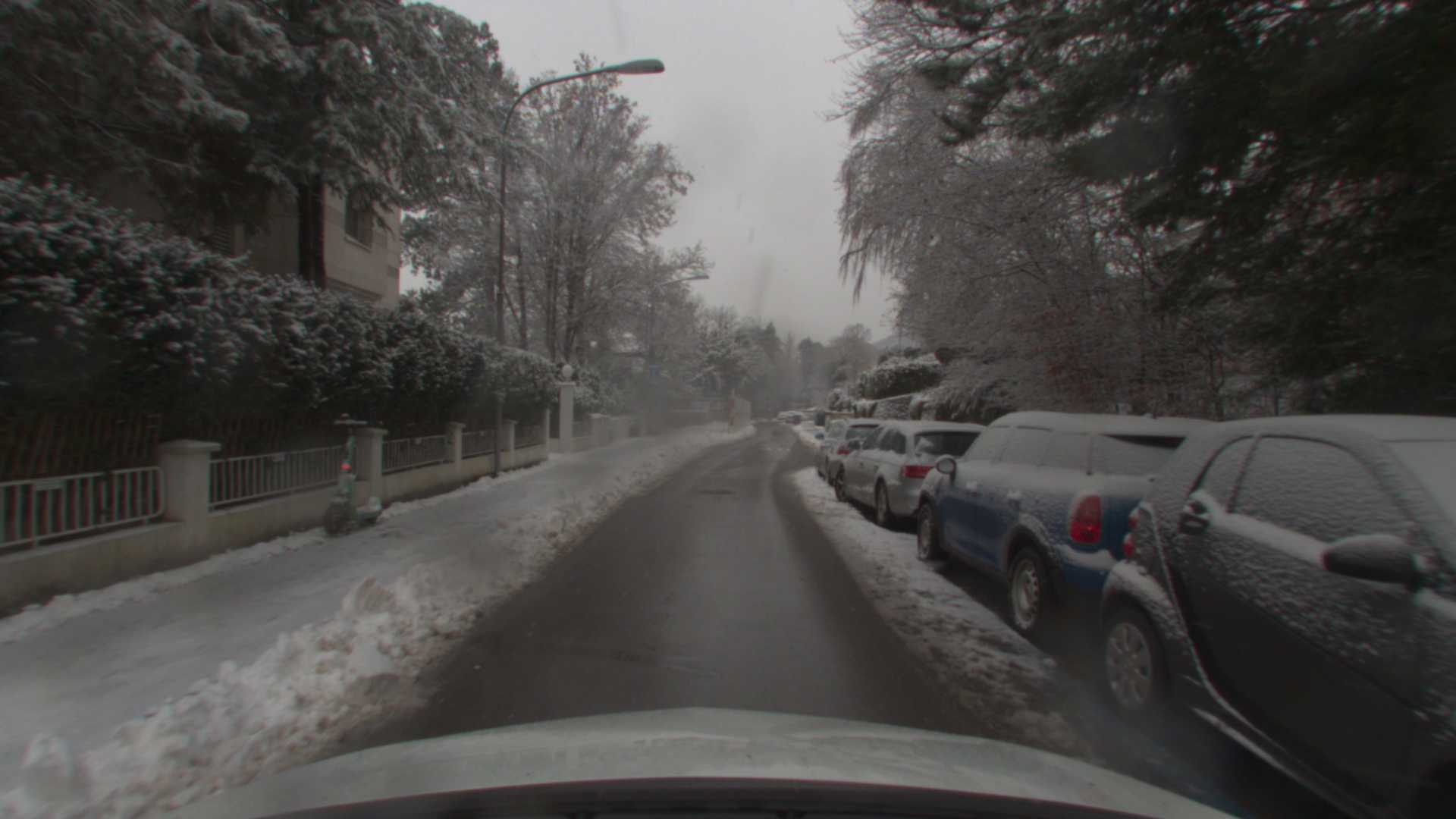} & 
\includegraphics[width=0.15\textwidth]{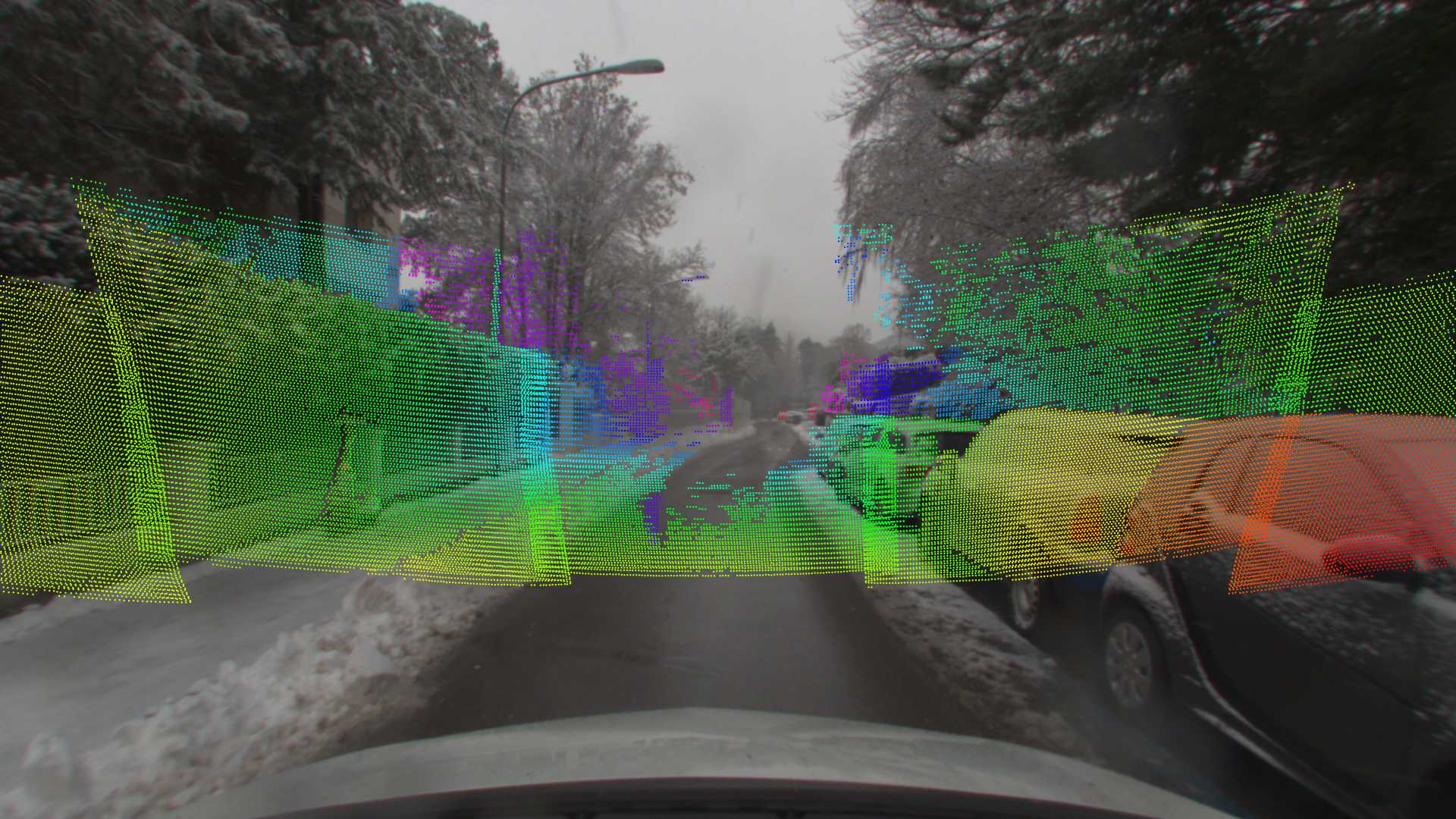} & 
\includegraphics[width=0.15\textwidth]{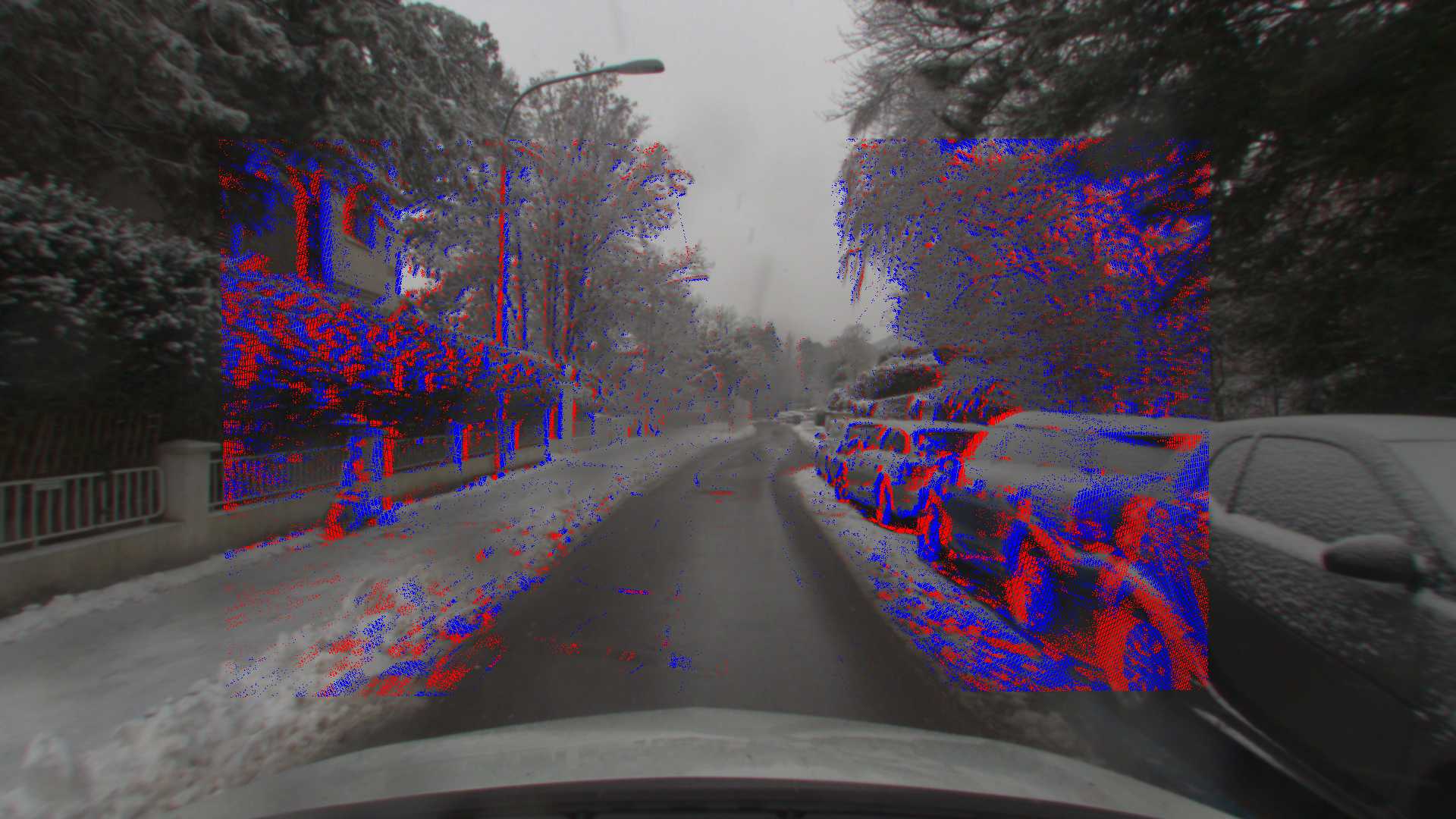} & 
\includegraphics[angle=90, trim=0 6835 0 0, clip,width=0.15\textwidth]{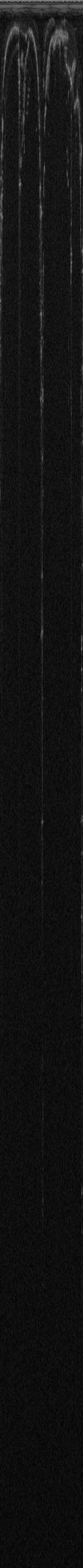} & 
\includegraphics[width=0.15\textwidth]{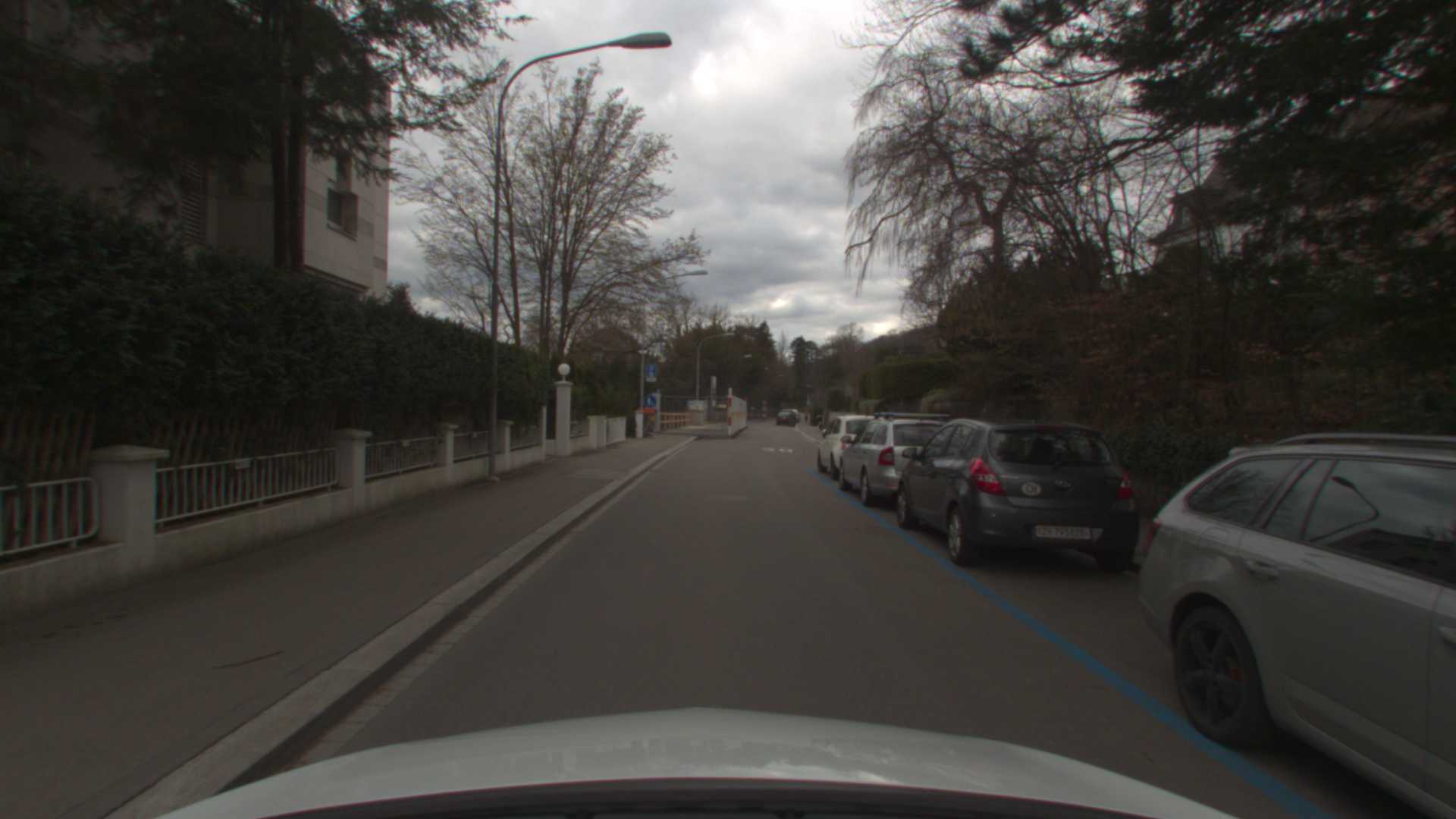} & 
\includegraphics[width=0.15\textwidth]{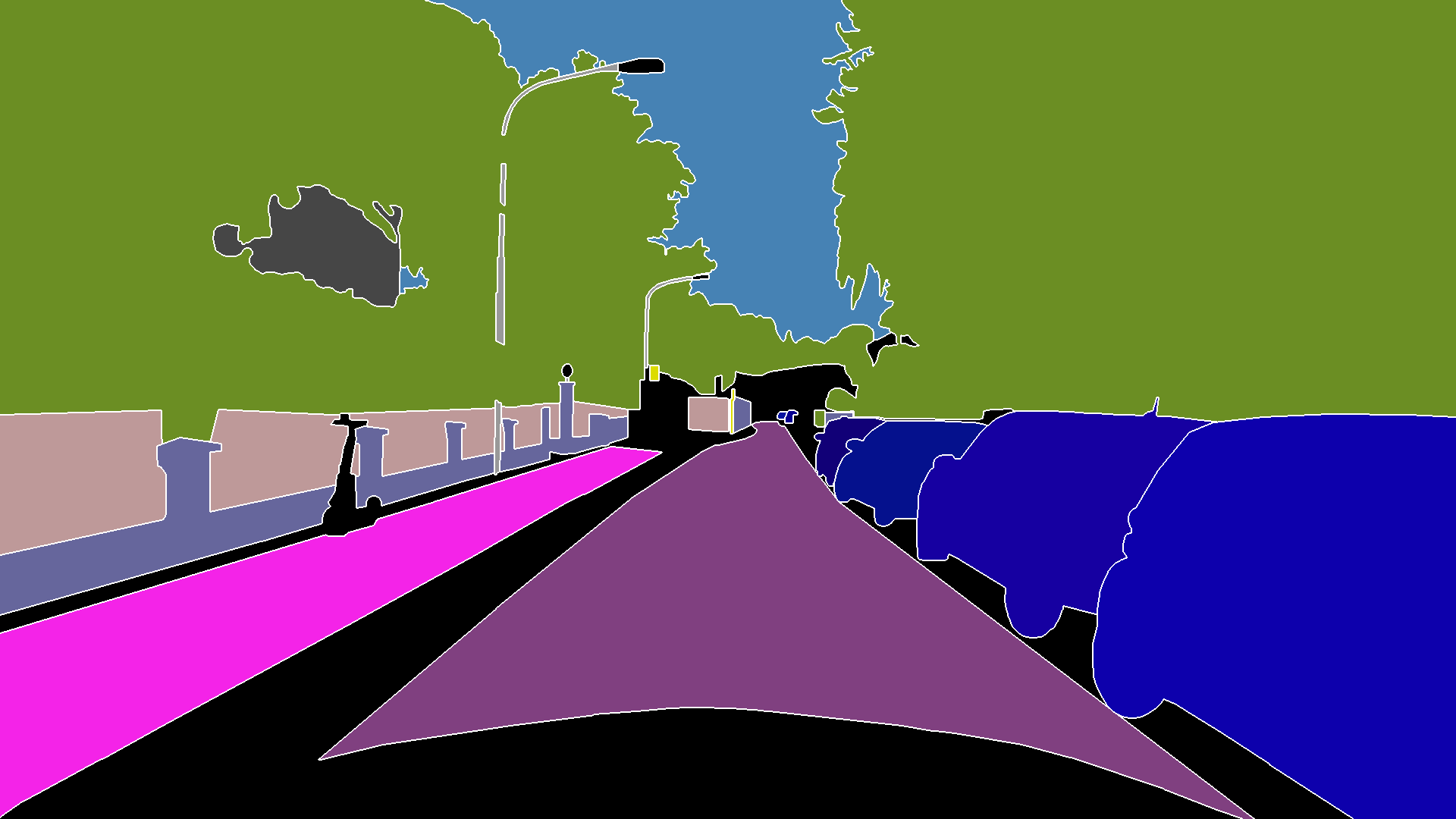} \\

\includegraphics[width=0.15\textwidth]{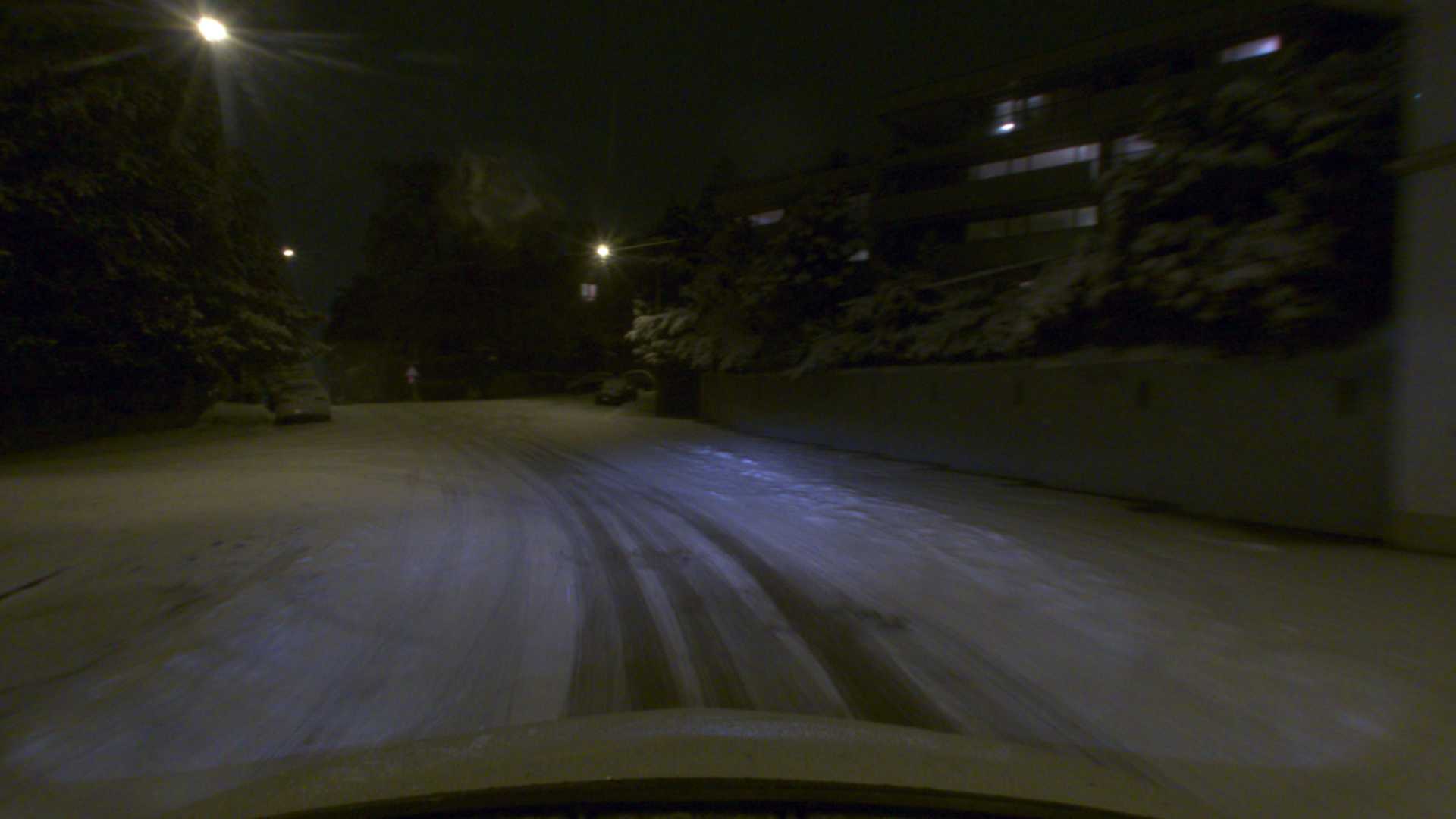} & 
\includegraphics[width=0.15\textwidth]{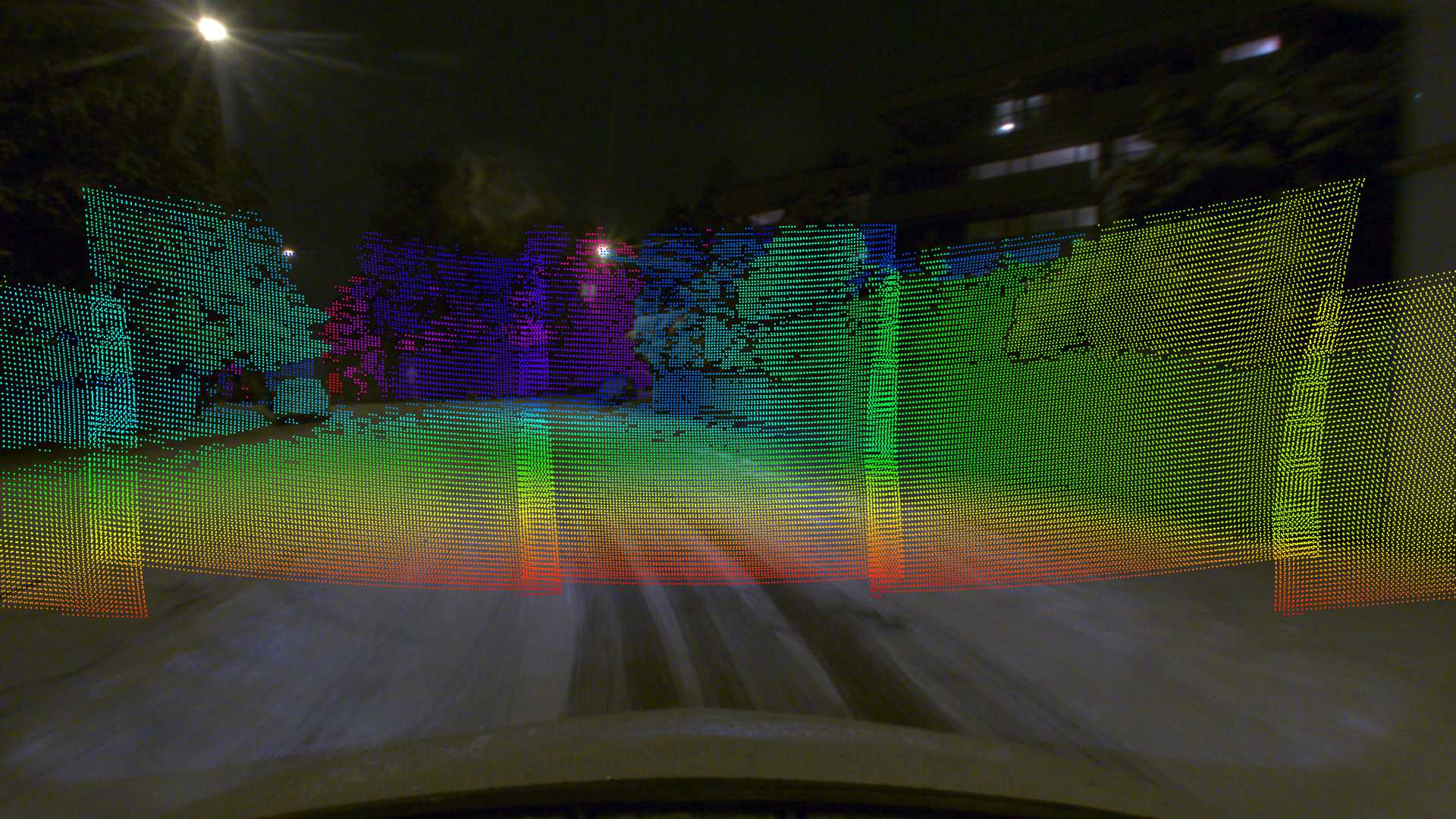} & 
\includegraphics[width=0.15\textwidth]{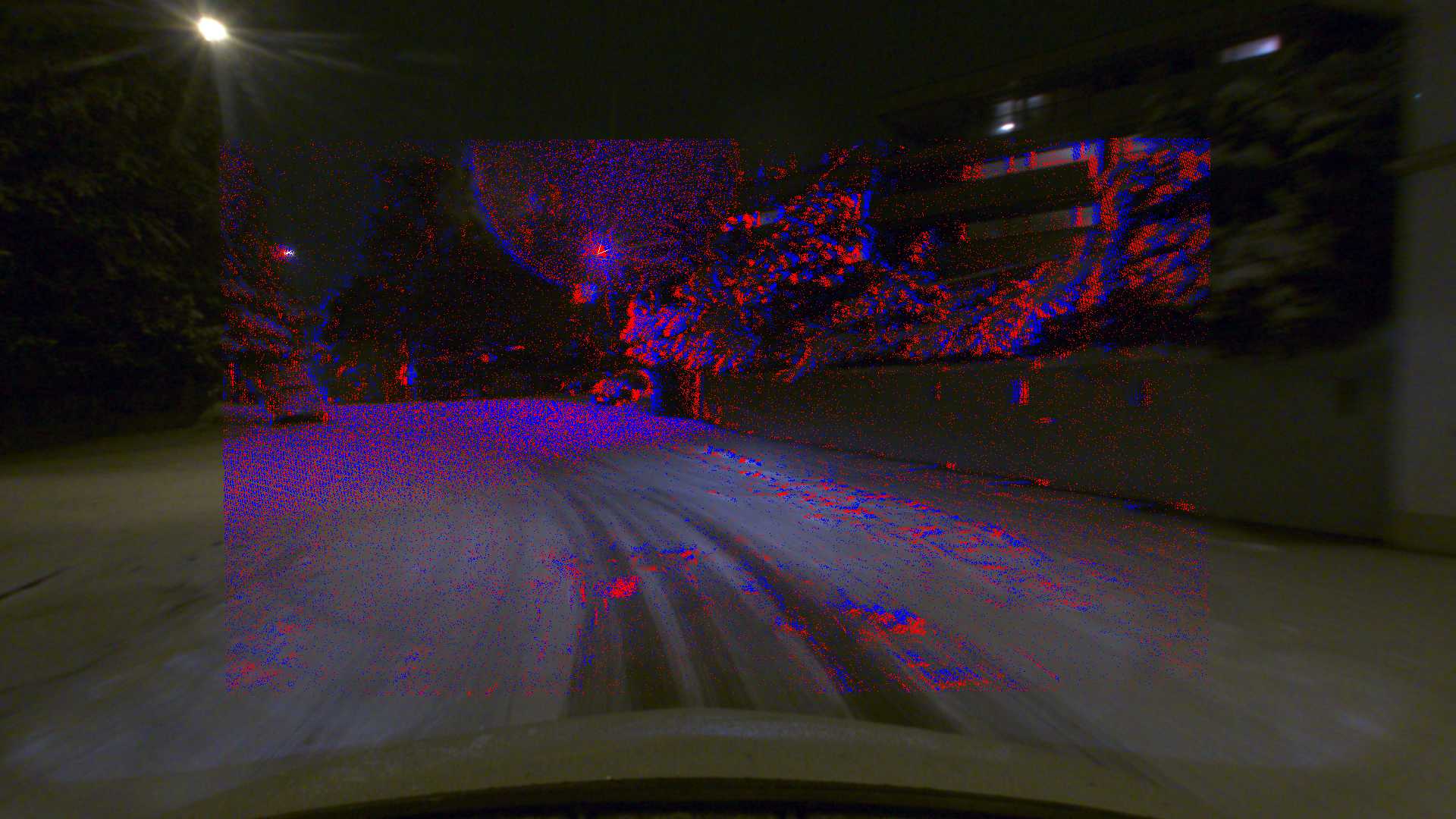} & 
\includegraphics[angle=90, trim=0 6835 0 0, clip,width=0.15\textwidth]{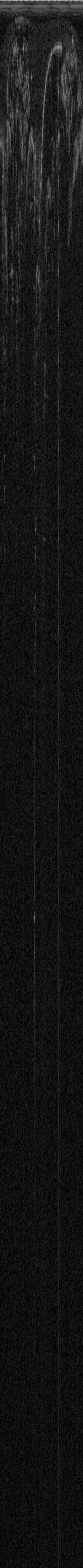} & 
\includegraphics[width=0.15\textwidth]{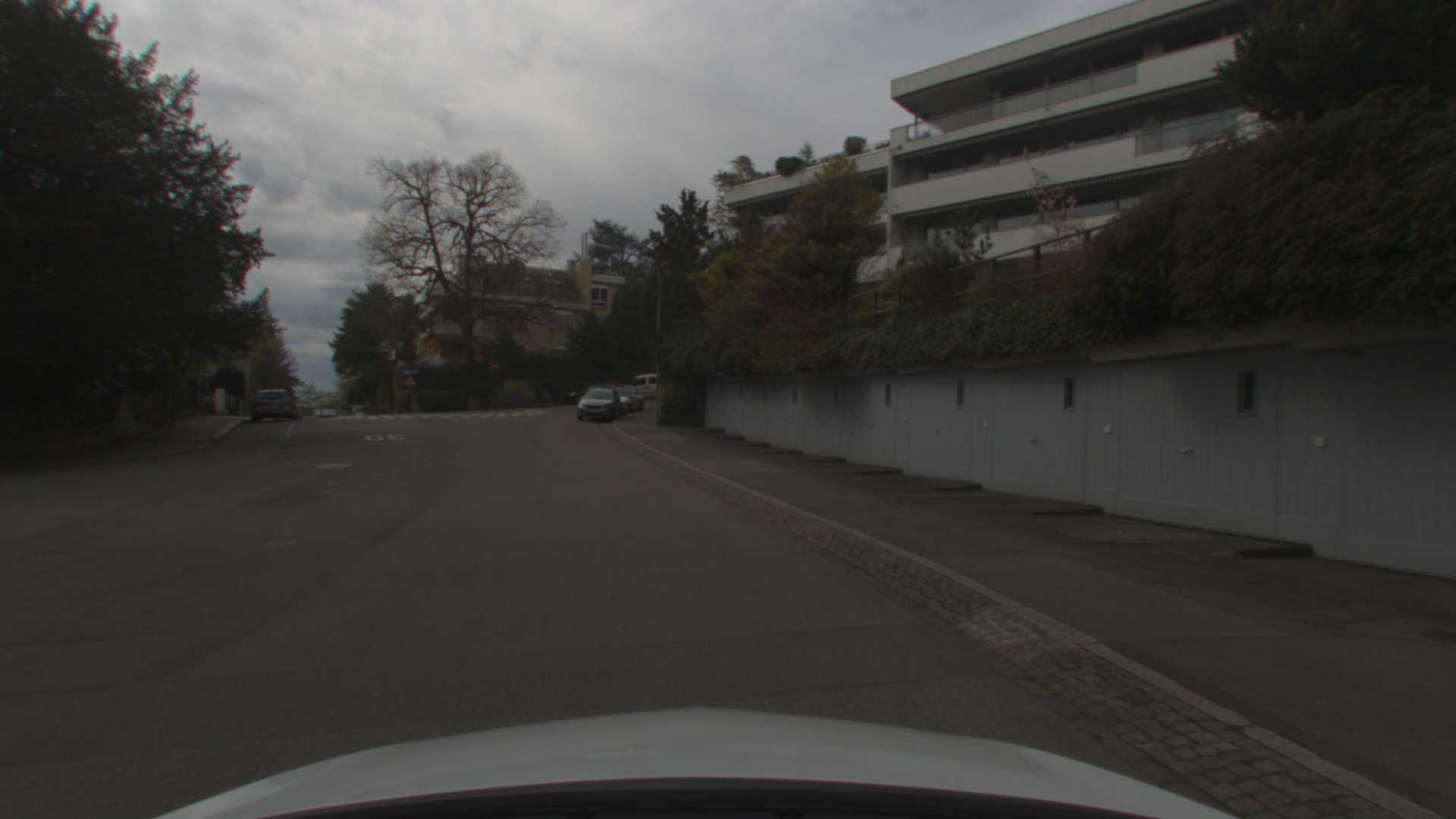} & 
\includegraphics[width=0.15\textwidth]{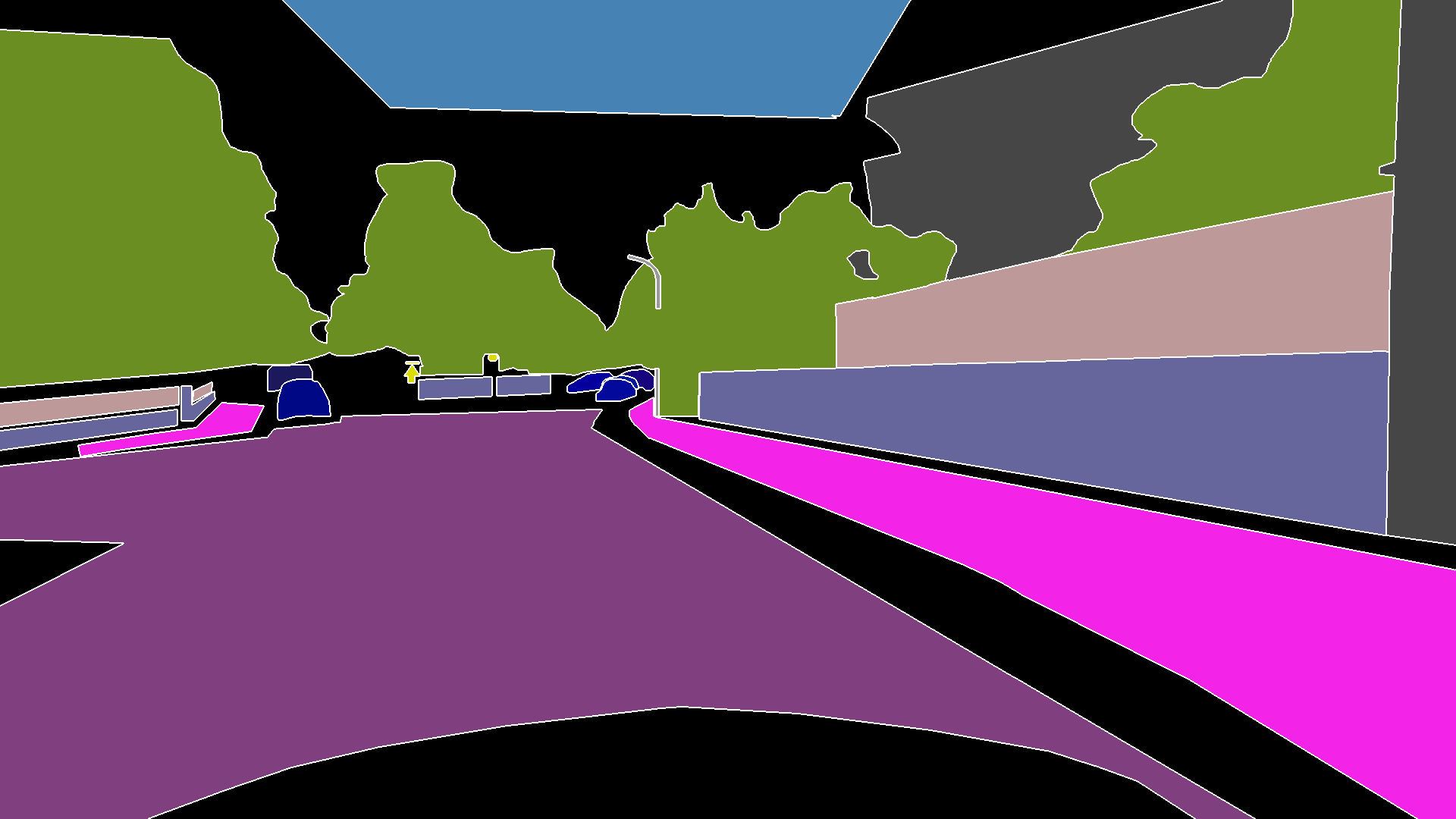}\\

\end{tabular}
\caption{\textbf{Example scenes from the MUSES dataset.} From left to right: RGB image, projected lidar points, projected events, azimuth-range radar scan, reference image, panoptic annotations. Best viewed zoomed in.}
\vspace{-8pt}
\label{fig:muses:samples}
\end{figure*}

A main characteristic of MUSES is the systematic coverage of two orthogonal axes of environmental degradation, namely illumination (day vs. night) and weather conditions (clear, fog, rain, snow). This results in a total of eight distinct condition combinations, allowing for controlled evaluation of model robustness across varying levels and types of degradation, as visible in the example scenes in Figure~\ref{fig:muses:samples}.

The dataset is deliberately skewed towards adverse conditions. Out of the 2,500 samples, 500 correspond to clear daytime reference scenes, while the remaining 2,000 samples capture adverse conditions distributed across both day and night. The detailed distribution across all condition combinations is summarized in Table~\ref{tab:muses_conditions}.

\begin{table}[ht]
\centering
\caption{\textbf{Distribution of MUSES samples across weather and illumination conditions.}}
\label{tab:muses_conditions}
\begin{tabular}{lcccc|c}
\toprule
 & Clear & Fog & Rain & Snow & Total \\
\midrule
Day   & 500 & 333 & 334 & 333 & 1500 \\
Night & 250 & 250 & 250 & 250 & 1000 \\
\midrule
Total & 750 & 583 & 584 & 583 & 2500 \\
\bottomrule
\end{tabular}
\end{table}

The dataset is split into 1,500 training, 250 validation, and 750 test samples, with splits geographically separated to ensure robust generalization. Test-set annotations are withheld and evaluation is performed via an online benchmark.

All scenes are annotated with panoptic labels following the 19-class taxonomy of Cityscapes. The RGB images are labeled through a two-stage process that leverages all available data, including recordings from both clear and adverse conditions, as well as pixel-accurate alignment of the different sensor modalities. This procedure improves label consistency and increases coverage in visually degraded regions, resulting in high-quality pixel-level panoptic labels.

Overall, MUSES constitutes a challenging and realistic benchmark for panoptic segmentation due to its multimodal nature, high-quality annotations, and extensive coverage of adverse conditions encountered in real-world driving. For further details on the sensor setup, calibration, and annotation protocol, we refer the reader to the MUSES paper~\cite{brodermann2024muses}.

\section{URVIS 2026: MUSES-AXPS Challenge}
The URVIS 2026 Challenge: MUSES-AXPS — Adverse-to-the-eXtreme Panoptic Segmentation is motivated by the need to evaluate and improve panoptic segmentation algorithms under degraded real-world conditions. It is built on the MUSES dataset~\cite{brodermann2024muses}, a multi-sensor benchmark designed for dense scene understanding in challenging weather and illumination scenarios. The challenge took place on Codabench \cite{codabench} \footnote{https://www.codabench.org/competitions/13395/}. A central objective of the challenge is to benchmark fusion strategies across a complete set of sensors (RGB camera, LiDAR, radar and event camera) and weather conditions, enabling a deeper understanding of sensor fusion under adverse environmental disturbances.

\subsection{Panoptic segmentation task}
The task of panoptic segmentation plays a crucial role in perception, as it brings together two fundamental vision tasks: semantic segmentation and instance segmentation \cite{kirillov2019panoptic}. This unified representation allows a system not only to distinguish object categories but also to differentiate between individual objects of the same type, offering highly detailed spatial and contextual information. In our case, the challenge is further amplified by adverse weather conditions, which degrade sensor data and make the segmentation task significantly more complex. Under such conditions, achieving reliable panoptic segmentation becomes even more critical to ensure robust perception and maintain safety in real-world environments. The MUSES dataset allows the participant to rely on four different sensors to achieve the best fusion for the task.

\subsection{Challenge phases}
The challenge was conducted in two separate phases: a public validation phase and a private final test phase. For the first phase of the challenge, we used the validation set to compute the evaluation metrics, thus enabling participants to fine-tune their models based on the provided ground-truth annotations. This phase remained open for nearly one month, allowing participants to make at most one hundred submissions; in total, thirty-two submissions were received during this period. For the final phase, which lasted only six days, with a maximum of five submission per participant, they were granted access to the full test-phase sensor data, a dataset distinct that used in the first phase, while the ground-truth annotations were deliberately withheld. A total of fifteen submissions from four teams were received during the final phase.

\subsection{Evaluation and Ranking Protocol}
We adopt the standard evaluation protocol for panoptic segmentation, based on Panoptic Quality (PQ), Segmentation Quality (SQ), and Recognition Quality (RQ). These metrics constitute the fundamental evaluation basis of the task, as they jointly assess both segmentation accuracy and semantic recognition performance.

For a set of matched prediction--ground-truth segments, PQ is defined as
\begin{equation}
\mathrm{PQ} =
\frac{\sum_{(p,g)\in \mathrm{TP}} \mathrm{IoU}(p,g)}
{|\mathrm{TP}| + \frac{1}{2}|\mathrm{FP}| + \frac{1}{2}|\mathrm{FN}|},
\end{equation}
where $\mathrm{TP}$, $\mathrm{FP}$, and $\mathrm{FN}$ denote the numbers of true positives, false positives, and false negatives, respectively, and $\mathrm{IoU}(p,g)$ is the intersection-over-union between a matched prediction segment $p$ and ground-truth segment $g$. Following the standard formulation, PQ can be decomposed as
\begin{equation}
\mathrm{PQ} = \mathrm{SQ} \times \mathrm{RQ},
\end{equation}
where SQ measures the average mask quality of matched segments,
\begin{equation}
\mathrm{SQ} =
\frac{\sum_{(p,g)\in \mathrm{TP}} \mathrm{IoU}(p,g)}
{|\mathrm{TP}|},
\end{equation}
and RQ evaluates the recognition quality,
\begin{equation}
\mathrm{RQ} =
\frac{|\mathrm{TP}|}
{|\mathrm{TP}| + \frac{1}{2}|\mathrm{FP}| + \frac{1}{2}|\mathrm{FN}|}.
\end{equation}
Intuitively, SQ reflects how accurately the predicted masks align with the ground-truth regions, while RQ reflects how well the method recognizes and detects valid panoptic instances. PQ therefore serves as the principal metric by jointly measuring both aspects.

To ensure a fair comparison among participants, we adopt a restricted evaluation protocol. First, the ground-truth annotations of the test set are not released. They are only made available for the validation split, where participants may use them during the validation phase for debugging, model development, and local analysis. Second, the test-set evaluation results are not publicly exposed during the challenge. All final results are computed by the organizers on the hidden test set. This design reduces the possibility of overfitting to public feedback and discourages challenge-specific tuning based on repeated access to test metrics.


Beyond the standard metrics, we further introduce weighted versions of the evaluation scores, namely weighted Recognition Quality (wRQ), weighted Segmentation Quality (wSQ), and weighted Panoptic Quality (wPQ), to better reflect the objective of the challenge. Specifically, let $\mathcal{W}=\{\text{clear},\text{fog},\text{rain},\text{snow}\}$ denote the set of weather conditions, and let $\lambda_w$ be the weight assigned to condition $w \in \mathcal{W}$, with
\begin{equation}
\lambda_w \ge 0.
\end{equation}
To place greater emphasis on performance under adverse weather, we assign a lower weight to clear weather under daytime and a higher weight to each adverse-weather condition. In particular, we set
\begin{equation}
\lambda_{\text{clear}}^{day} = 0.5, \qquad
\lambda_{\text{fog / rain / snow}}^{day / night} = 1,
\end{equation}
so that each adverse-weather condition receives twice the weight of clear weather.
We then define
\begin{equation}
\mathrm{wRQ} = \sum_{w\in\mathcal{W}} \lambda_w \, \mathrm{RQ}_w,
\end{equation}
\begin{equation}
\mathrm{wSQ} = \sum_{w\in\mathcal{W}} \lambda_w \, \mathrm{SQ}_w,
\end{equation}
and
\begin{equation}
\mathrm{wPQ} = \sum_{w\in\mathcal{W}} \lambda_w \, \mathrm{PQ}_w,
\end{equation}
where $\mathrm{RQ}_w$, $\mathrm{SQ}_w$, and $\mathrm{PQ}_w$ are the corresponding scores computed on the subset associated with weather condition $w$. Unlike a uniform average over all conditions, these weighted metrics assign higher importance to more challenging adverse-weather scenarios. Since this challenge primarily focuses on robust multimodal panoptic segmentation under extreme weather, the fog, rain, and snow subsets are given larger weights in the final evaluation. At the same time, the clear-weather subset is still retained with a non-zero proportion, ensuring that methods also maintain competitive performance under standard driving conditions. In this way, the proposed weighted protocol emphasizes robustness in adverse environments while preserving balanced performance across diverse real-world scenarios.

The final ranking of the challenge is determined by \textbf{wPQ} on the hidden test set, while wSQ and wRQ are reported as complementary indicators for detailed analysis.

\section{Methods}
In this study, we use the top four methods from the URVIS 2026 MUSES-AXPS Challenge, as well as some representative methods from recent research \cite{Cheng_2022_CVPR_mask2former,brodermann2025cafuser}.
\subsection{Two-Stage Mask2Former - MLJP Team}
\textbf{Description:}
We build upon a Mask2Former baseline for panoptic segmentation.
To improve robustness under adverse conditions, we adopt a two-stage training strategy: pretraining on clear-day data followed by fine-tuning on all weather and illumination conditions.
This strategy improves performance when evaluated on all conditions (+20.0 PQ\_all). We further explore a simple multimodal extension by incorporating LiDAR-derived features (depth, height, density), projected into the image plane and concatenated with RGB inputs.
The first convolution layer is adapted accordingly, while the rest of the architecture remains unchanged.
However, this naive fusion does not lead to performance improvement, suggesting that effective multimodal perception requires more structured cross-modal interaction.
\textbf{Implementation:}
We use Mask2Former with a ResNet-50 backbone, MSDeformAttn pixel decoder, and a multi-scale transformer decoder with 100 object queries.
The model is trained using AdamW with a base learning rate of $1\times10^{-4}$ and weight decay of $1\times10^{-4}$.
Training is performed with a batch size of 4 using mixed precision. The training consists of two stages: (1) pretraining on clear-day data for 800 epochs, and (2) fine-tuning on all conditions for 90 epochs.
All-condition training is initialized from an intermediate checkpoint (epoch 450) of the clear-day model. Standard data augmentation includes resizing and random horizontal flipping.
Gradient clipping and multi-step learning rate scheduling are enabled, although learning rate decay is not reached due to early stopping.
Total training time is approximately 17 hours on a single GPU (RTX 3090).

\subsection{RGB Mask2Former - ElieT Team} 
\textbf{Description:}
Our best-performing approach is the Mask2Former baseline trained end-to-end utilizing exclusively RGB images from the camera. To improve robustness over adverse conditions, we test leveraging the multi-sensor data projected in the image plane in two ways: (1) early fusion, by concatenating all modalities and adapting the first convolution layer, (2) intermediate fusion, using a shared-weight backbone across modalities and performing stage-wise feature fusion prior to the pixel decoder. The intermediate fusion strategies evaluated include naive feature averaging, feature projection before fusion, and a cross-attention module between RGB and auxiliary modalities. However, both fusion methods underperformed compared to our RGB-only baseline, indicating that simple fusion techniques are not sufficient to improve performance.

\textbf{Implementation:}
We use Mask2Former with a ResNet-50 backbone, MSDeformAttn pixel decoder, and a multi-scale transformer decoder with 100 object queries. The model is trained using AdamW with a base learning rate of $1\times10^{-4}$ and weight decay of $5\times10^{-2}$. The model is trained end-to-end on RGB images spanning all weather conditions for 160 epochs. Standard data augmentations are applied, including resizing, cropping, and random horizontal flipping. Furthermore, gradient clipping and a polynomial learning rate schedule are employed. The total training time is approximately 26h on a single Nvidia T4 GPU.

\subsection{LiDAR-guided Mask2Former - Michele24 Team}

\begin{figure}
    \centering
    \includegraphics[width=0.5\linewidth]{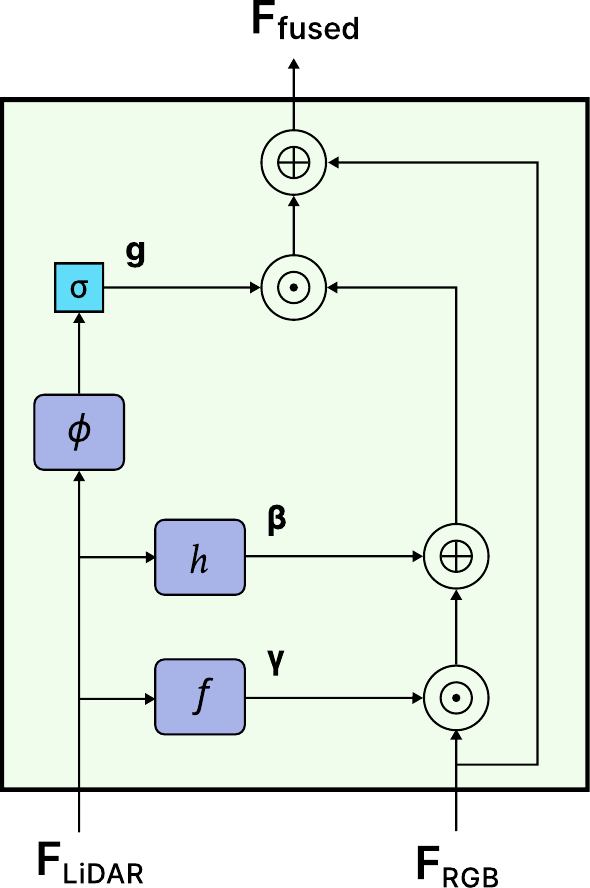}
    \caption{The feature modulation block for a given resolution. Taking image ($\bm{F}_{\text{RGB}}$) and LiDAR ($\bm{F}_{\text{LiDAR}}$) feature maps as input, the symbols $\bm{\gamma}$, $\bm{\beta}$, and $\bm{g}$ denote the scale, shift, and gating tensors, respectively. The blocks $f$, $h$, and $\phi$ represent $1 \times 1$ convolutions, while $\sigma$ indicates the sigmoid activation. The symbols $\oplus$ and $\odot$ indicate element-wise addition and the Hadamard product.}
    \label{fig:fusion-block-michele-cazzola}
\end{figure}

\textbf{Description}
Our approach incorporates a mid-fusion strategy between the Mask2Former architecture, featuring a Swin-T backbone, and multi-scale LiDAR features extracted via a lightweight convolutional encoder. The LiDAR point cloud is projected onto the RGB image plane, and a corresponding validity mask is concatenated to provide explicit information regarding the presence of the data.

We employ a gated residual modulation applied independently at each feature resolution, inspired by Feature-wise Linear Modulation (FiLM) \cite{perez2018film} and illustrated in \cref{fig:fusion-block-michele-cazzola}. Specifically, RGB features undergo an affine transformation where the scale ($\bm{\gamma}$) and shift ($\bm{\beta}$) parameters are derived from the LiDAR features using $1 \times 1$ convolutions. A gate ($\bm{g}$), followed by a sigmoid activation, is used to regulate the modulation intensity.
Our method shows modest improvements with respect to RGB-only and early-fusion baselines, with the addition of only 6\,M parameters.

\textbf{Implementation:} The model is trained end-to-end for 15 epochs without freezing any component. We utilize the AdamW optimizer and a cosine annealing scheduler, with a learning rate of $1 \times 10^{-5}$ (weight decay 0.01) for the Mask2Former components, and $5 \times 10^{-4}$ (weight decay 0.001) for the LiDAR encoder and fusion blocks.

Training is performed on a single NVIDIA Tesla T4 GPU with 15\,GB of available memory and a batch size of 2, requiring approximately 10 hours. To accommodate hardware constraints, input images are resized to a resolution of $1024 \times 512$. The resulting model exhibits a latency of 198\,ms and 138\,GFLOPs at inference time (measured with batch size 1), values that are only slightly higher than those of the RGB-only and early-fusion baselines.
The code is available at \href{https://github.com/MicheleCazzola/muses-axps}{https://github.com/MicheleCazzola/muses-axps}.

\subsection{Adapted CAFuser - wg Team}
\textbf{Description and implementation:}
We used a vGPU-32GB graphics card and selected the CAFuser model, which is used for Condition-Aware Multimodal Fusion for Robust Semantic Perception of Driving Scenes. The environment was Python 3.9.18, PyTorch 2.3.1 and CUDA 11.8. The main libraries used were Detectron2-v0.6. CAFuser is a condition-aware multimodal fusion architecture designed to enhance robust semantic perception in autonomous driving. It employs a Condition Token (CT), dynamically guiding the fusion of multiple sensor modalities to optimize performance across diverse scenarios. To train the CT, a verbo-visual contrastive loss aligns it with semantic environmental descriptors, enabling direct prediction from RGB features. The Condition-Aware Fusion module uses the CT to adaptively fuse sensor data based on environmental context. Further, CAFuser introduces modality-specific feature adapters, aligning inputs from different sensors into a shared latent space, and integrates these without loss of performance with a single shared backbone.

We used the pre-trained weights of CAFuser with Swin-T as the backbone for model training. After the training was completed, we retained the best weights and evaluated them on the MUSES dataset. During the evaluation process, we adjusted the configuration file to reduce the number of fusion layers, keeping only res5 and res4, or using FP16 inference. The final submitted result was the optimal outcome obtained.

\section{Experimental results}

The official results of the URVIS: MUSES-AXPS challenge are presented in table \ref{tab:weighted_PQ}. For the purpose of this challenge, we adapted the original panoptic segmentation metrics by introducing weather-weighted metrics to better reward performance under adverse weather conditions. To this end, we defined the weighted PQ: wPQ, the weighted SQ: wSQ and the weighted RQ: wRQ. These metrics are derived from the standard panoptic segmentation metrics, but each sample is assigned a weight according to its weather condition. Scores from clear day samples are multiplied by 0.5, while samples from any of the seven other conditions ( snow, fog, rain, day or night ) retained a weight of 1.

According to these criteria, the top performing participant is wg team achieving a weighted panoptic quality of 54.23. Interestingly, the top-performing method used the entire set of sensors ( RGB camera, event camera, LiDAR and radar). The second-best method relied on two sensors: the RGB camera and the LiDAR, while the last two methods used only RGB camera. These results suggest that additional sensing modalities can be beneficial under adverse conditions, but the gain cannot be attributed to modality count alone, as architecture design, pretraining, and fusion strategy also differ across methods.

\begin{table}[ht]
\caption{\textbf{Quantitative results for the URVIS MUSES-AXPS challenge}}
\label{tab:weighted_PQ}
\centering
\begin{tabular}{l|l|lll}
\toprule
Rank & team name              & wPQ   & wSQ   & wRQ   \\ \midrule
1 & wg       & 54.23 & 76.62 & 65.66 \\ 
2 & michele24     & 47.03 & 72.61 & 57.62 \\ 
3 &  eliet         & 45.84 & 73.23 & 56.40 \\ 
4 & mljp          & 36.15 & 68.28 & 45.58 \\ \bottomrule
\end{tabular}
\end{table}

In table \ref{tab:PQ_all}, we compare the results of baseline methods: Mask2Former \cite{Cheng_2022_CVPR_mask2former} and CAFuser-CA² \cite{brodermann2025cafuser}, the latter being the method provided by the authors of the MUSES dataset, with those obtained by the challenge participants. CAFuser-CA² remains the best-performing approach in terms of PQ across all weather conditions as well as the overall. The top two solutions outperform the RGB-only Mask2Former baseline, while the third-ranked submission remains slighly below it. Interestingly, the method from wg team achieves a particularly strong score in rainy conditions, reaching 57.9 PQ, which is almost identical to its performance in clear weather, 58.0 PQ. More generally, all participant methods perform noticeably better during daytime than nighttime and under clear weather compared to snow, rain and fog. Overall, the benchmark suggests that robust panoptic segmentation under adverse conditions does not necessarily require highly complex architectural redesigns. Strong pretrained segmentation models, combined with appropriate weather adaptation, already provide a surprisingly competitive starting point.

\begin{table*}[ht!]
\centering
\caption{\textbf{Comparison of the panoptic quality values per weather condition for baselines and challengers methods}}
\label{tab:PQ_all}
\begin{tabular}{l|l|lllllll}
\toprule
Rank & Team    & Clear & Fog  & Rain & Snow & Day  & Night & All  \\ \midrule
1 & wg     & 58.0  & 50.2 & 57.9 & 52.2 & 56.4 & 52.8  & 54.6 \\ 
2 & michele24   & 50.6  & 44.3 & 48.8 & 46.2 & 49.5 & 45.4  & 47.4 \\ 
3 &eliet       & 49.3  & 44.5 & 45.9 & 45.6 & 50.4 & 42.2  & 46.3 \\ 
4 & mljp        & 40.3  & 35.0 & 35.4 & 36.1 & 42.9 & 30.4  & 36.7 \\ \midrule
\multirow{2}{*}{Baseline} 
  & Mask2Former     & 48.8  & 46.5 & 45.4 & 45.1 & 49.3 & 39.4  & 46.9 \\ 
  & CAFuser-CA²     & 61.4  & 57.5 & 59.6 & 57.2 & 59.5 & 57.3  & 59.7 \\ \bottomrule
\end{tabular}
\end{table*}

\begin{figure*}
\centering

\begin{tabular}{ >{\centering\arraybackslash}m{2cm}  >{\centering\arraybackslash}m{4.5cm}  >{\centering\arraybackslash}m{4.5cm}  >{\centering\arraybackslash}m{4.5cm} }

& \textbf{Snow night} & \textbf{Rain night} & \textbf{Fog day} \\

Frame camera &
\includegraphics[width=\linewidth]{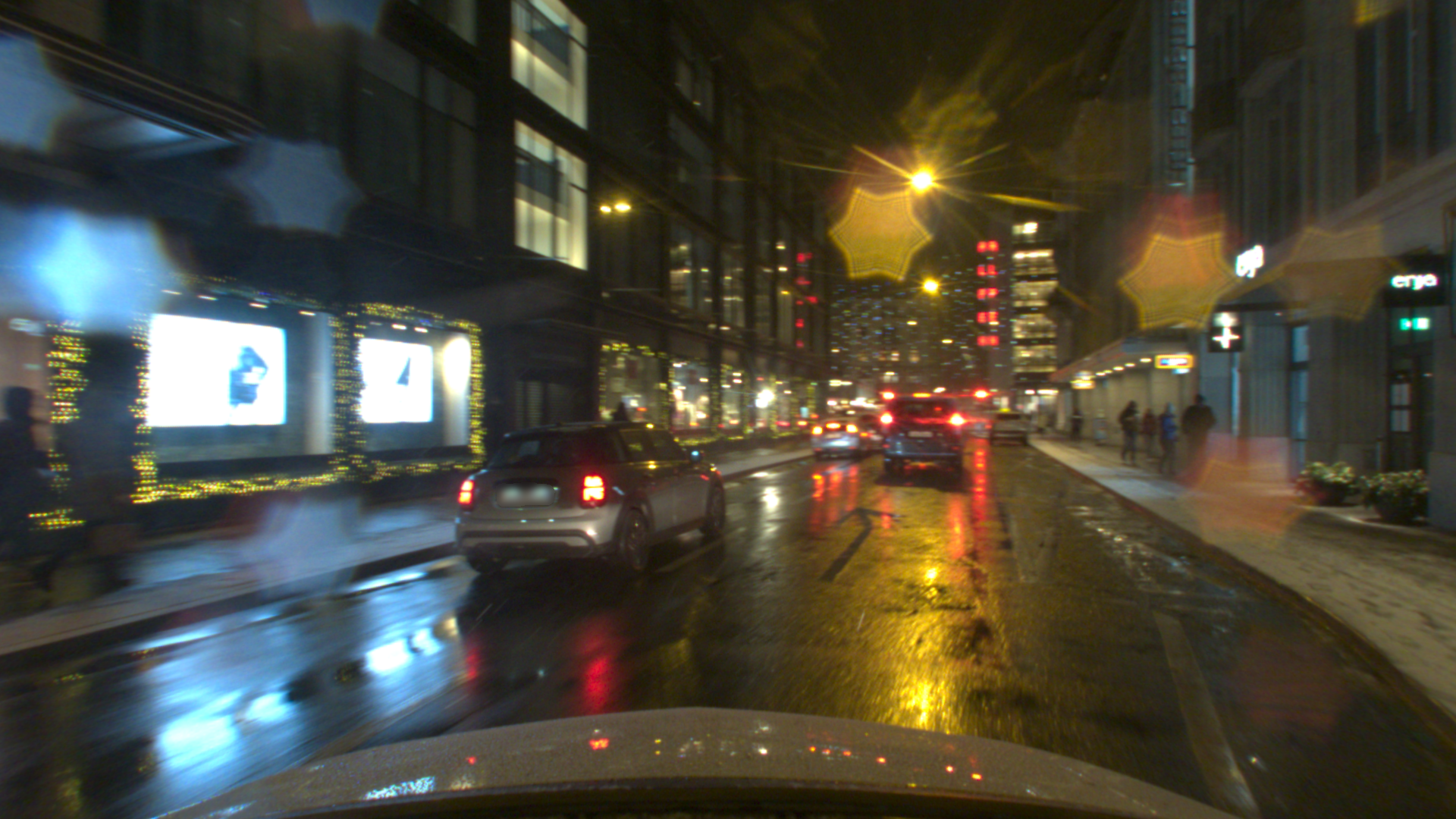} &
\includegraphics[width=\linewidth]{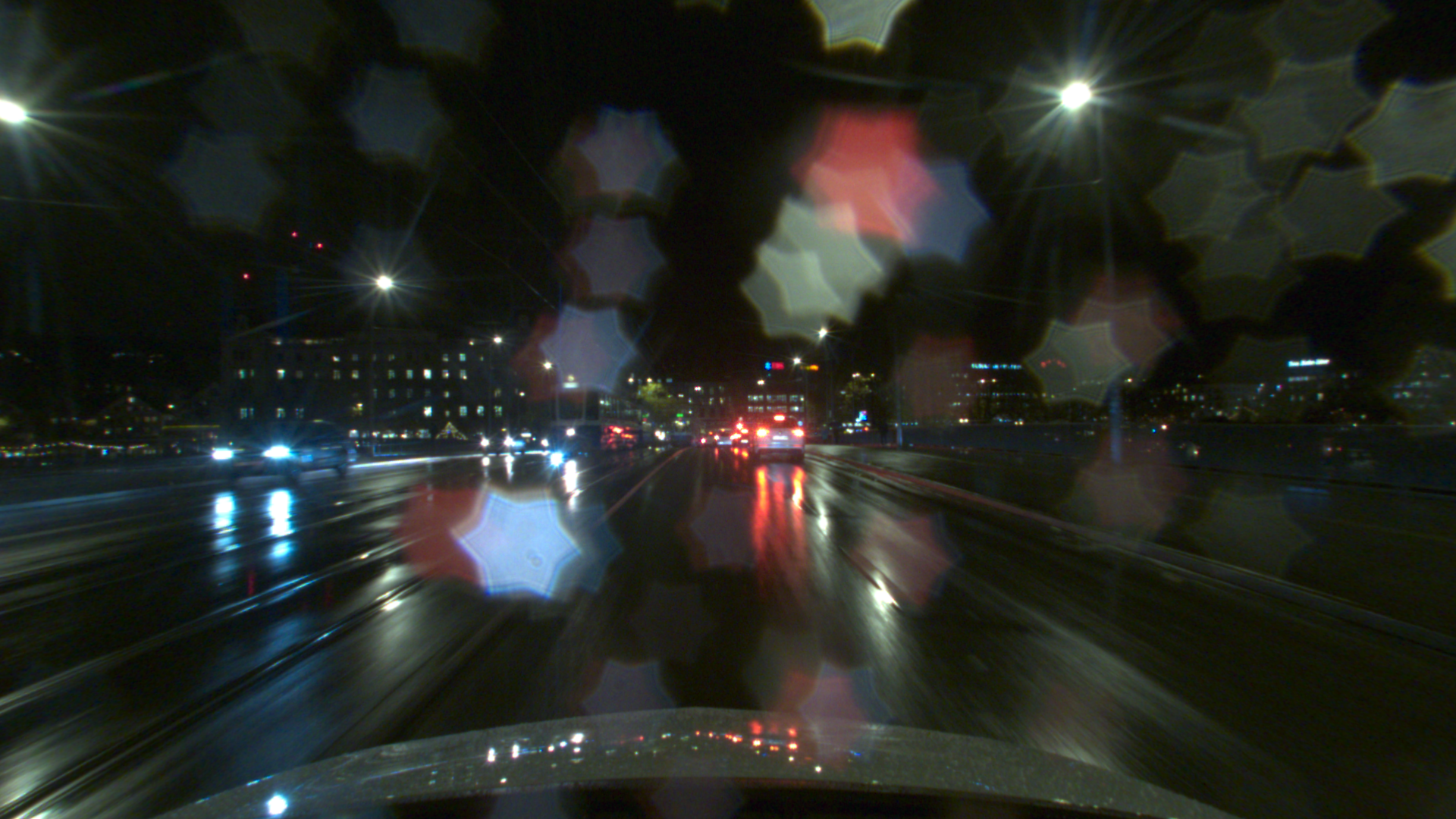} &
\includegraphics[width=\linewidth]{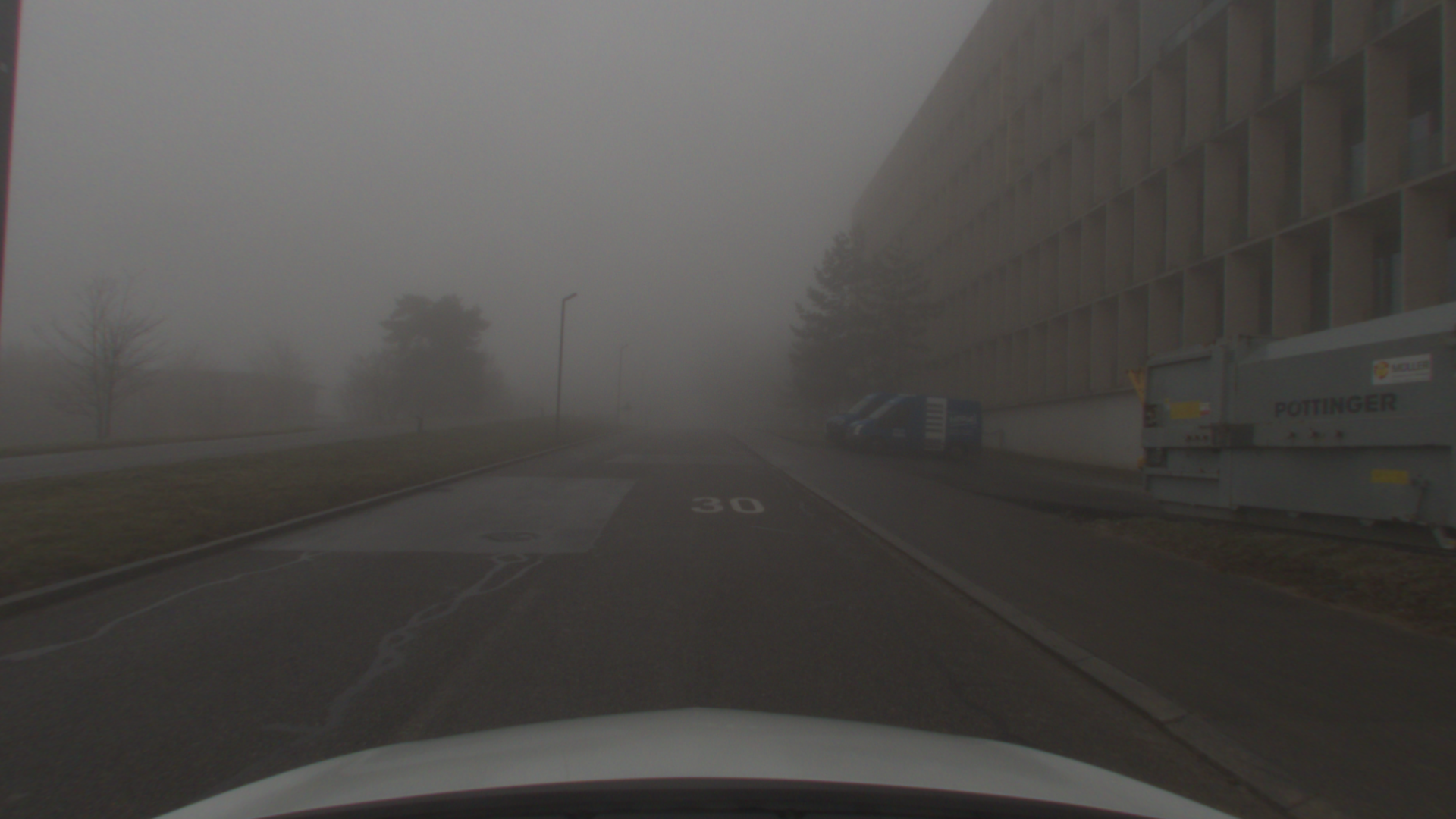} \\

Ground Truth &
\includegraphics[width=\linewidth]{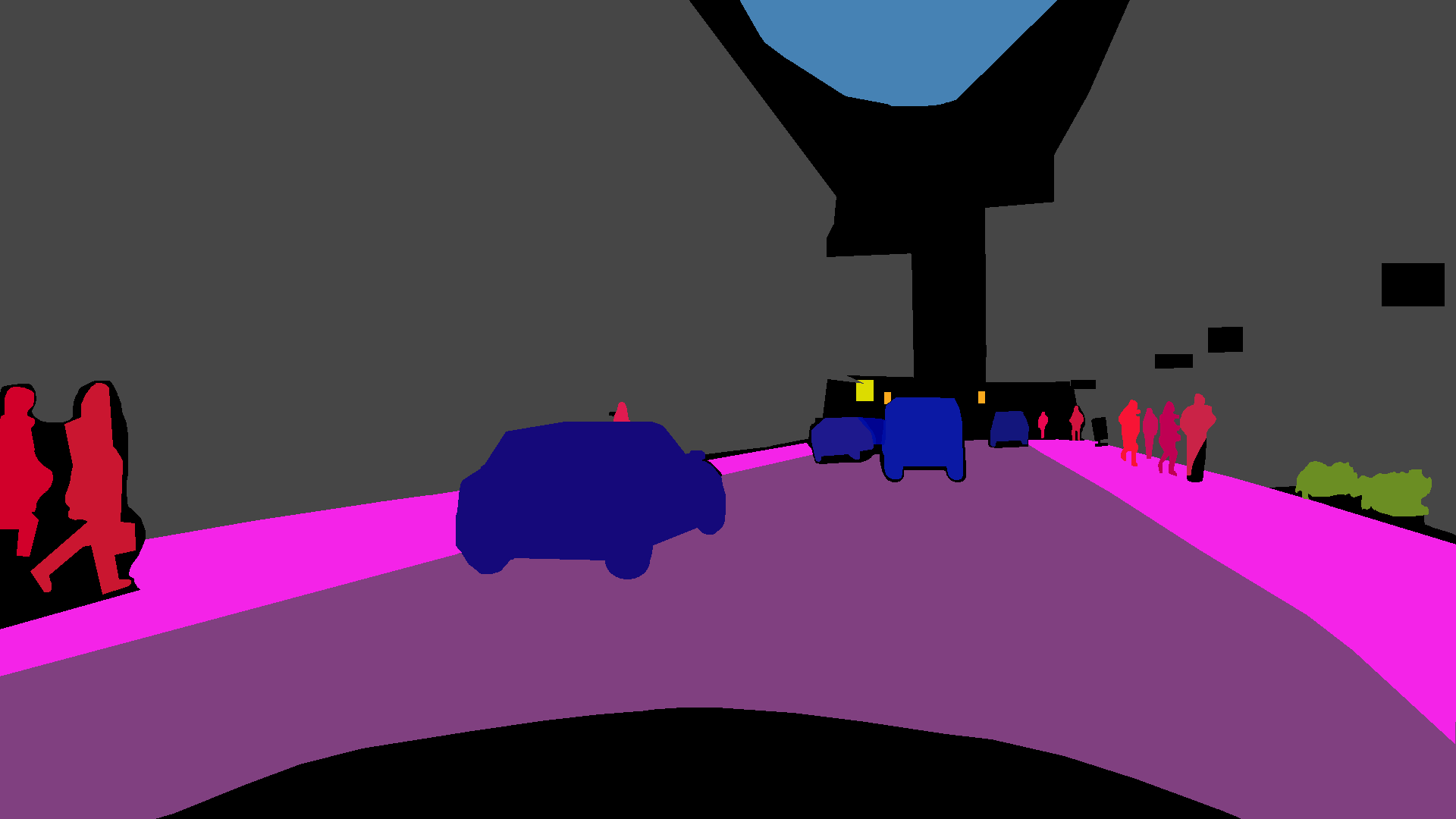} &
\includegraphics[width=\linewidth]{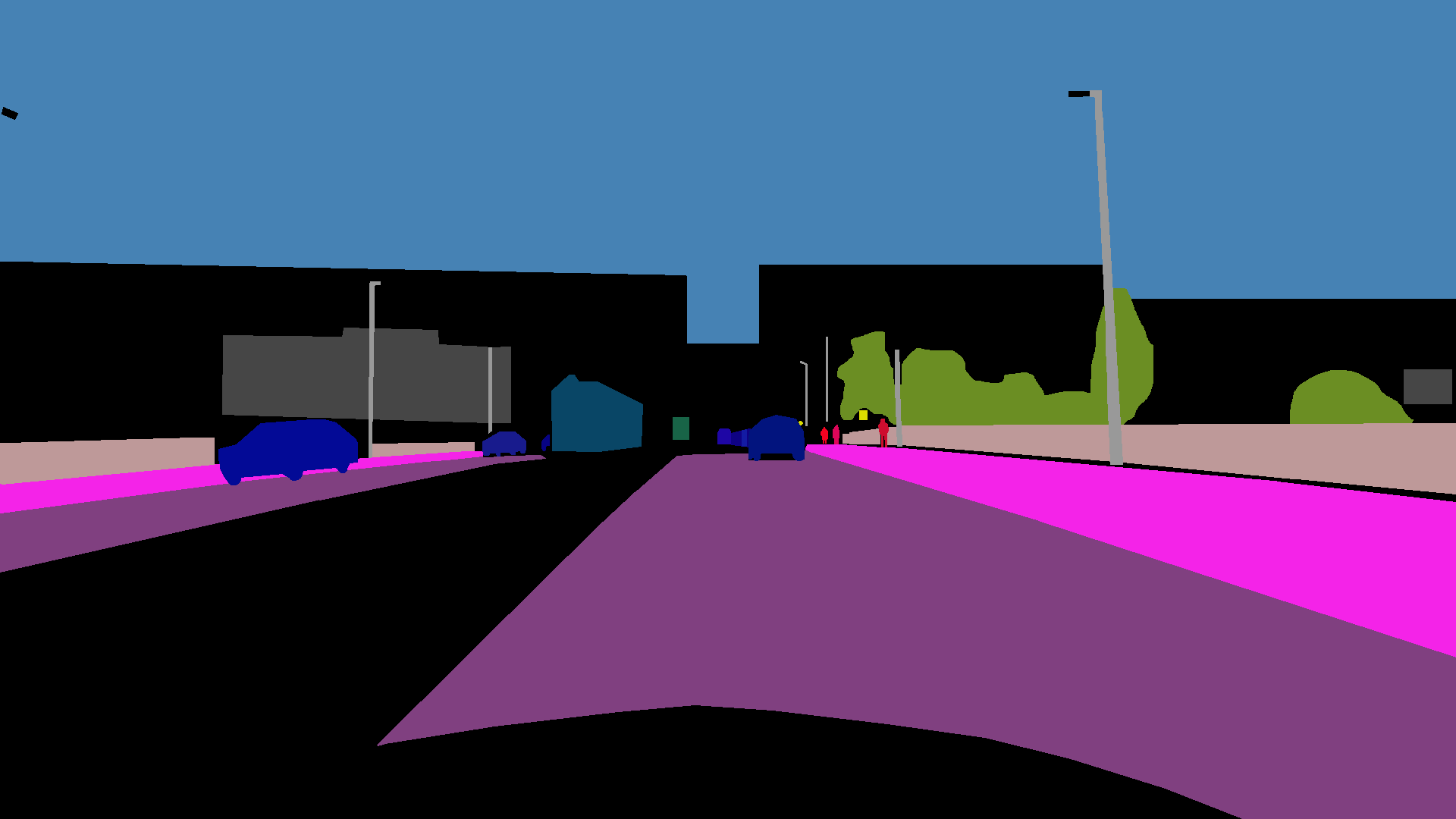} &
\includegraphics[width=\linewidth]{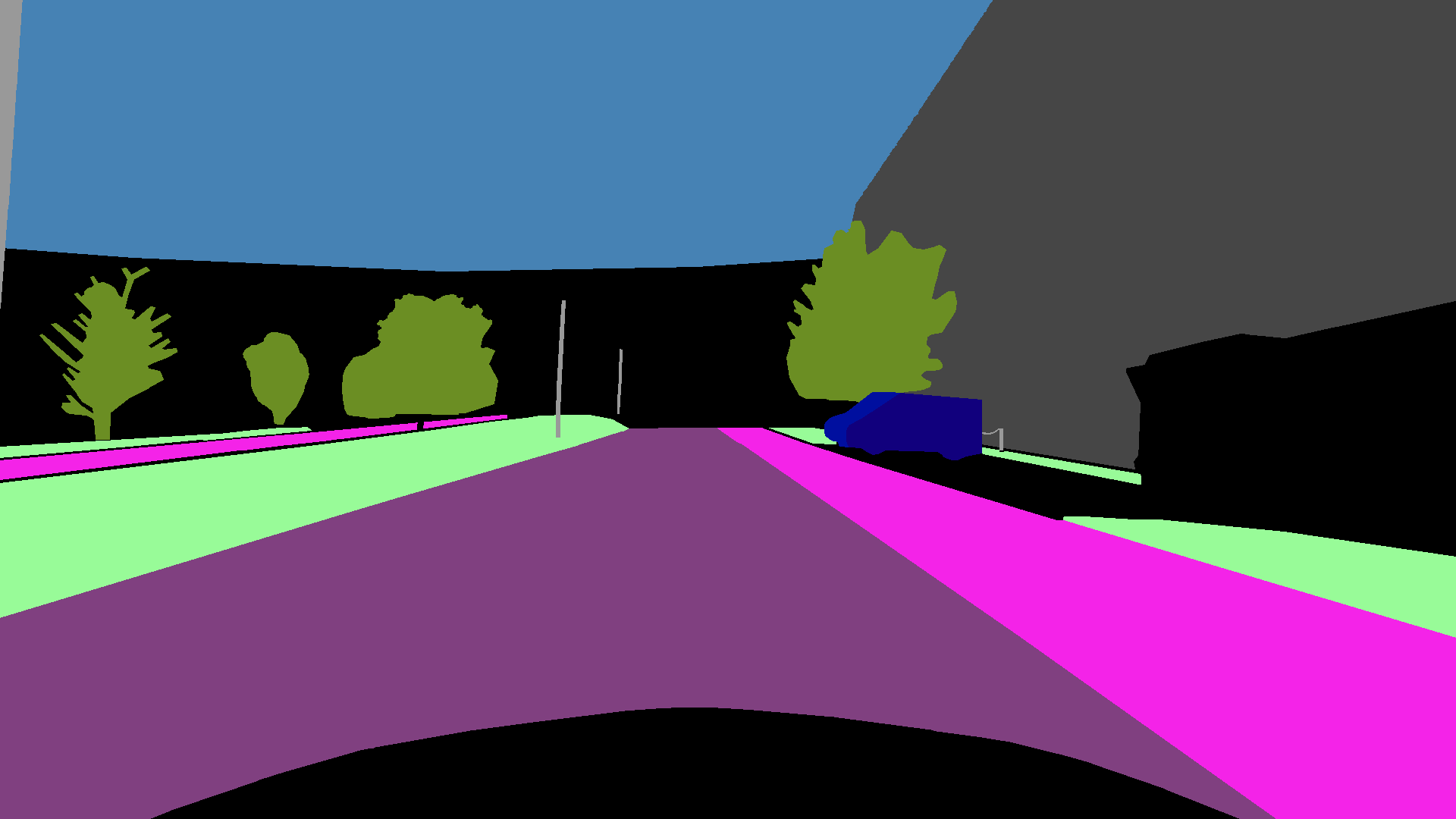} \\

wg &
\includegraphics[width=\linewidth]{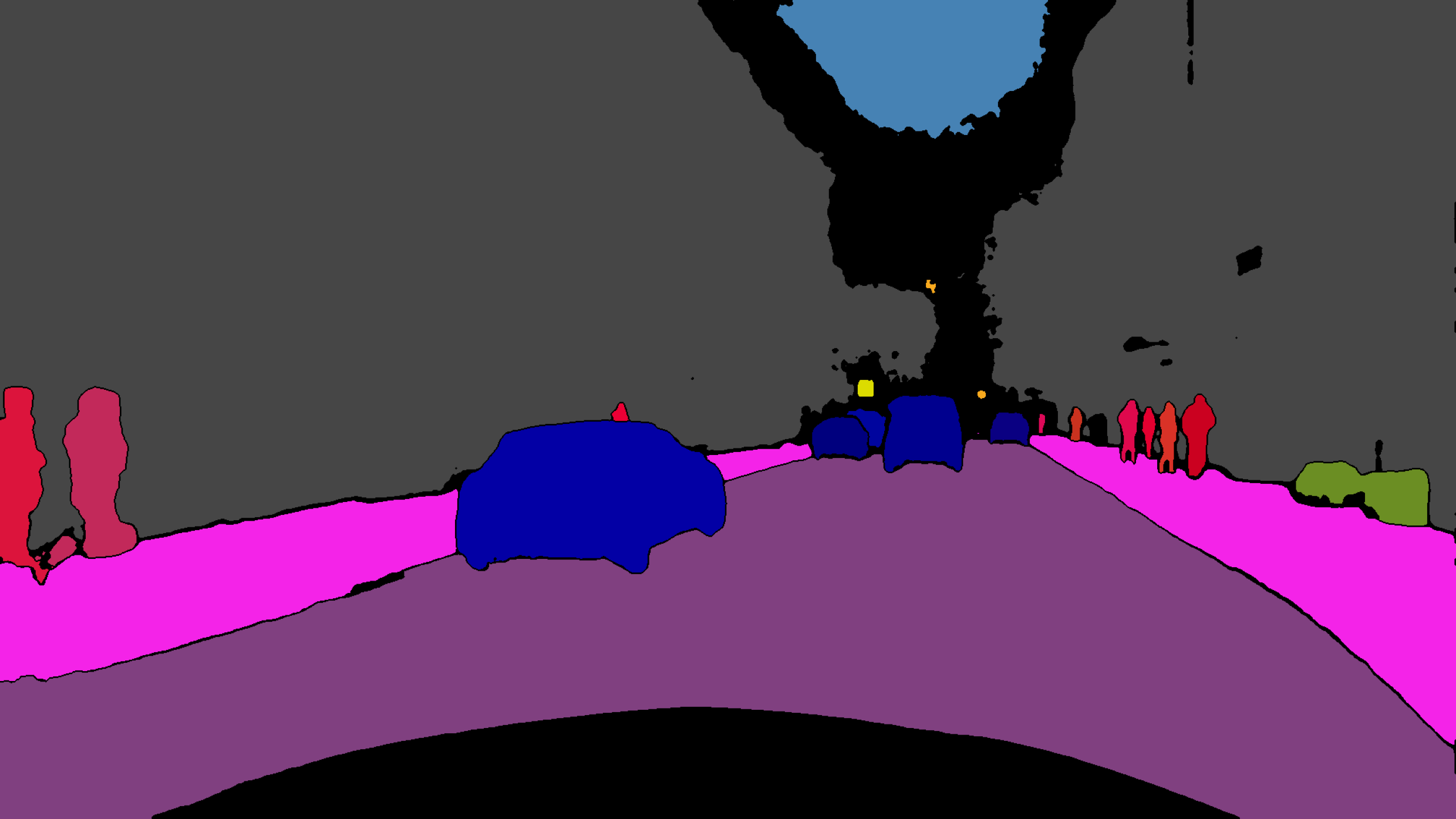} &
\includegraphics[width=\linewidth]{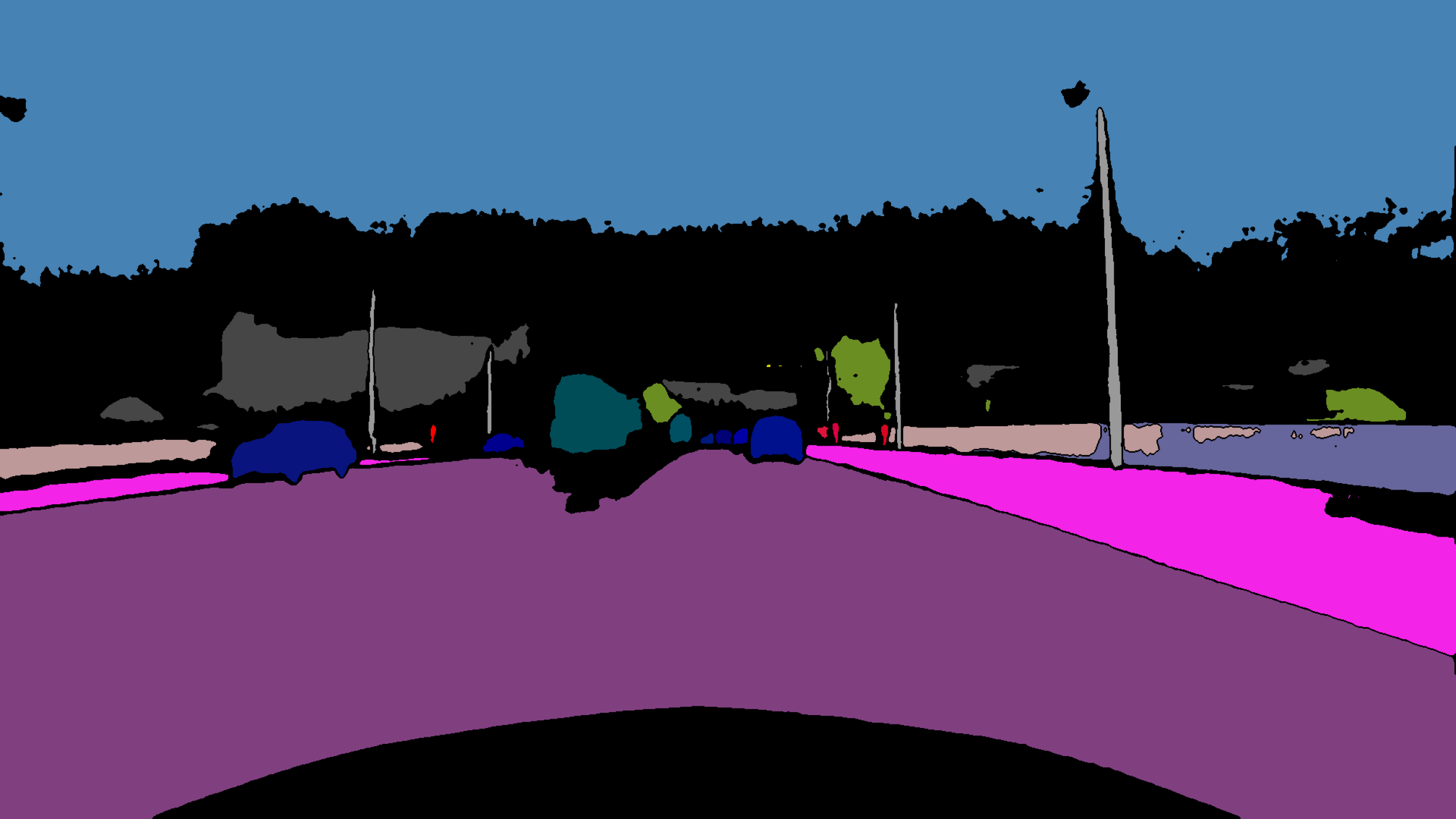} &
\includegraphics[width=\linewidth]{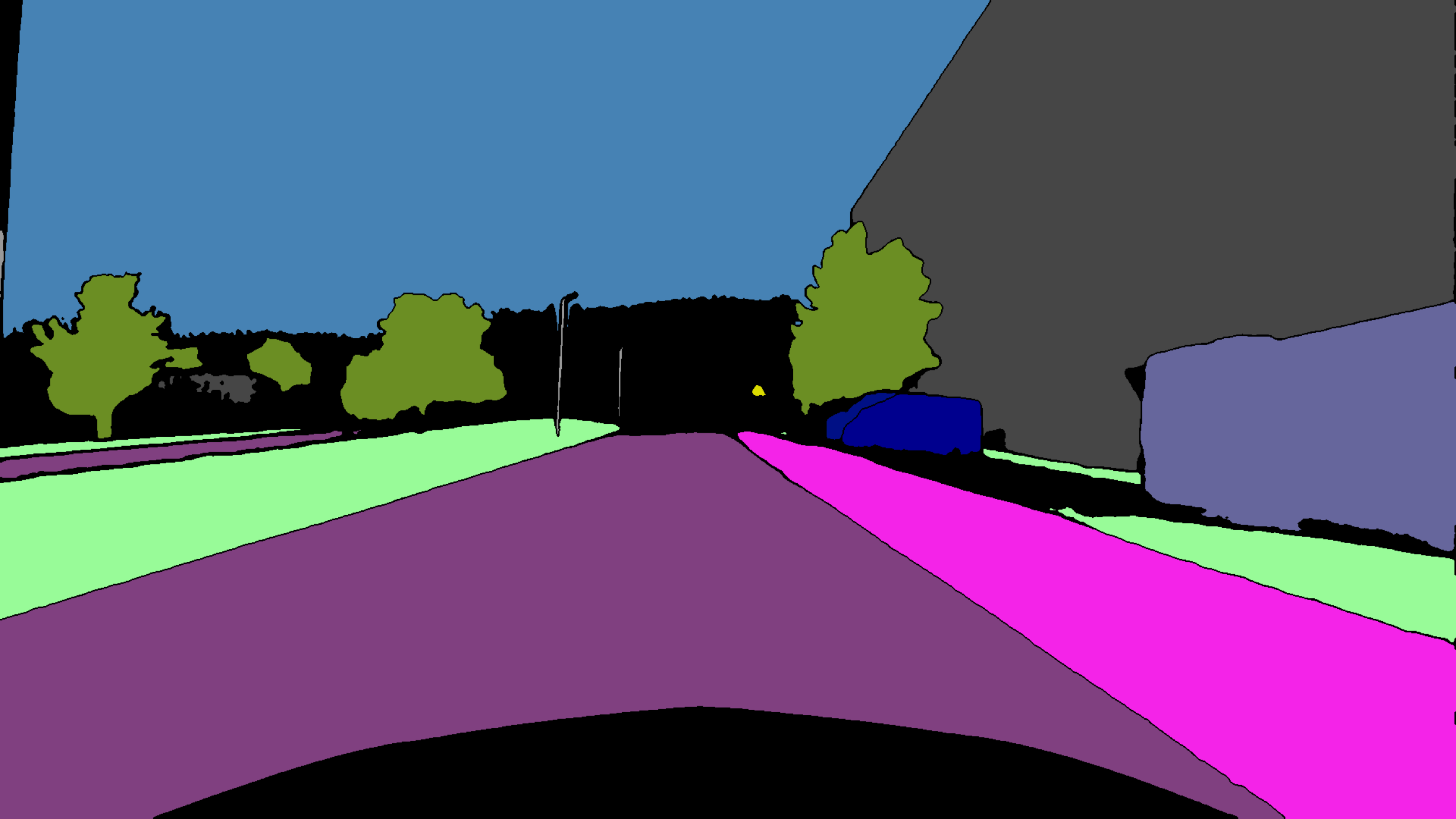} \\

michele24 &
\includegraphics[width=\linewidth]{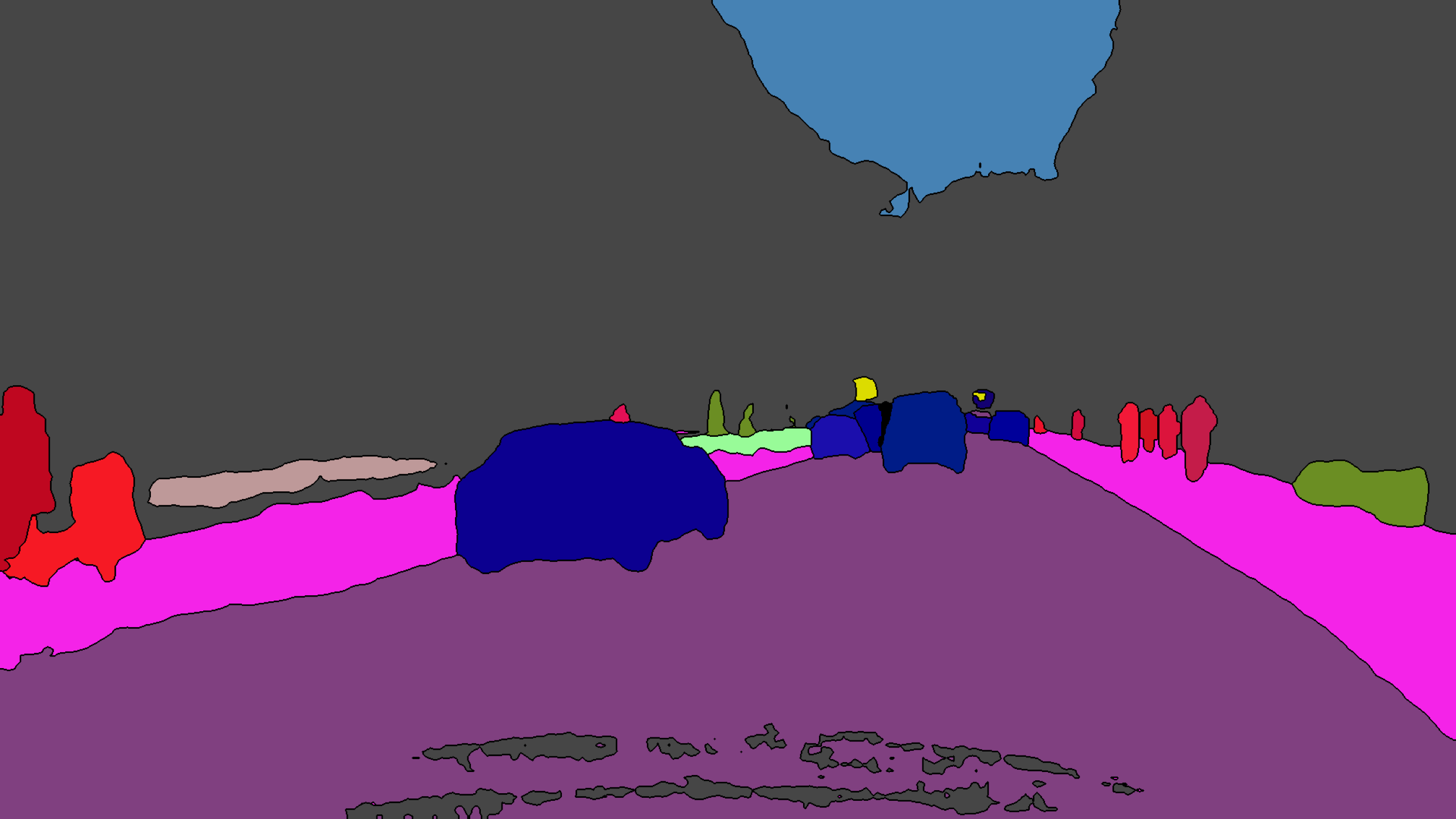} &
\includegraphics[width=\linewidth]{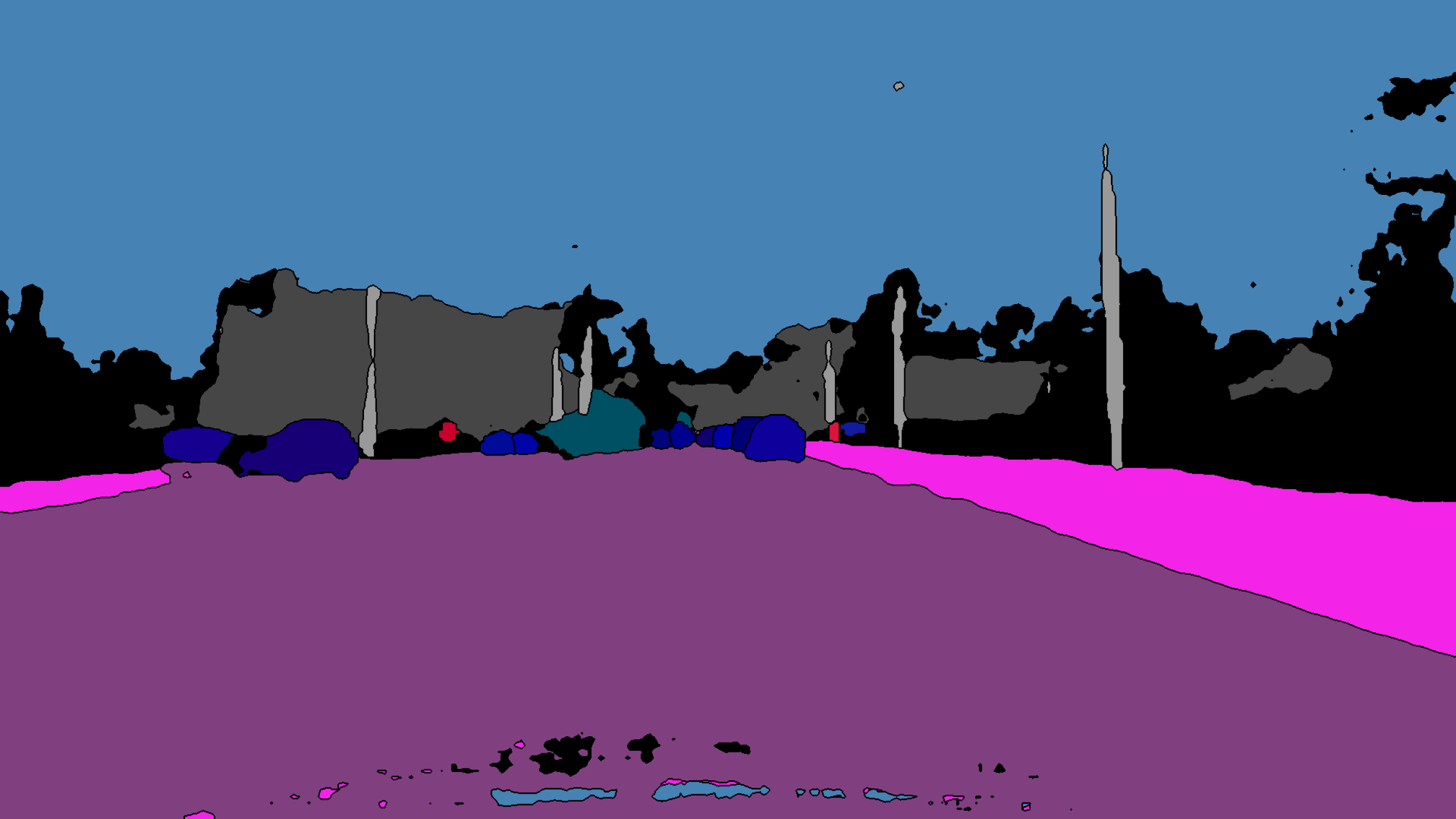} &
\includegraphics[width=\linewidth]{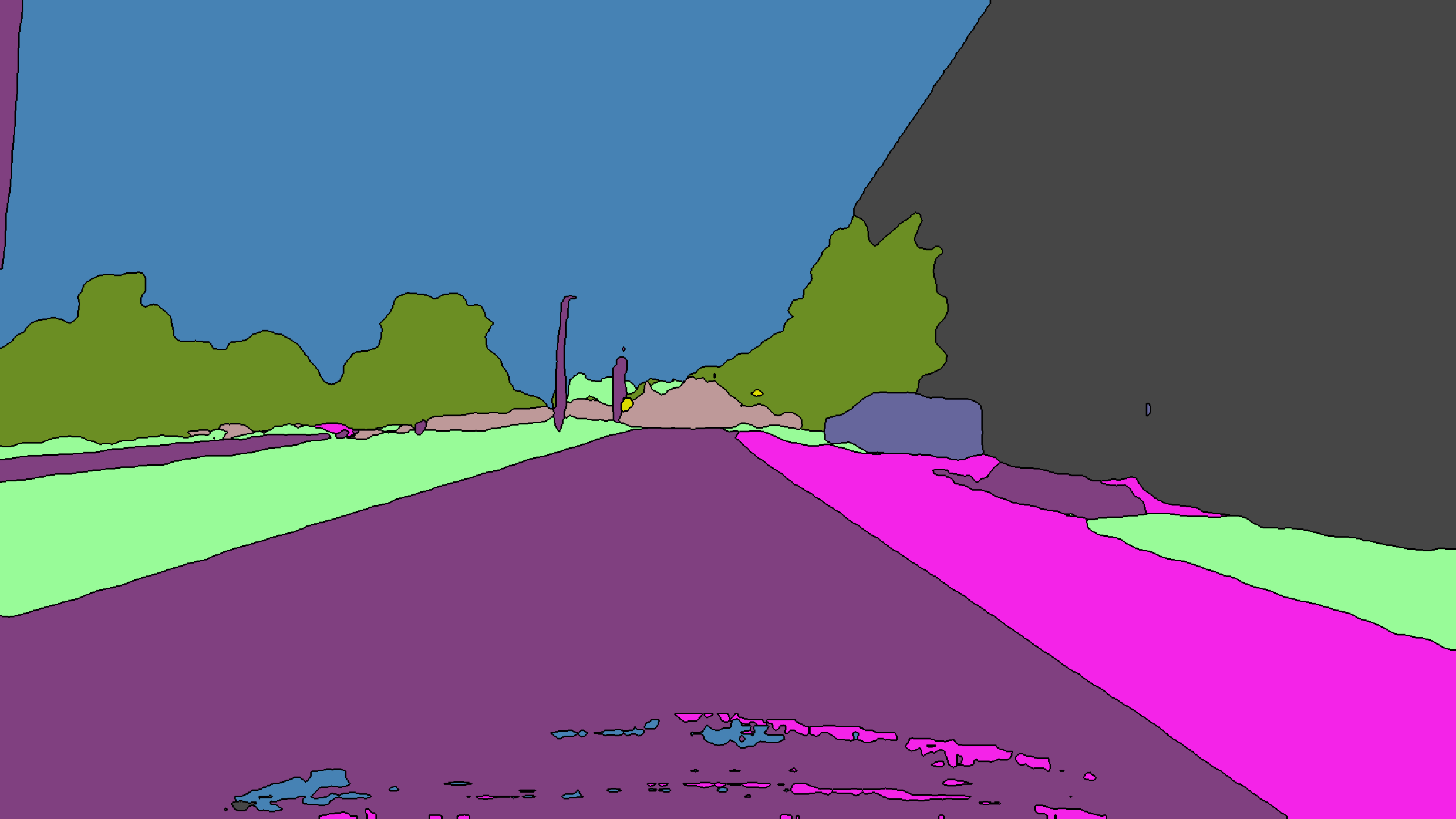} \\

elieT &
\includegraphics[width=\linewidth]{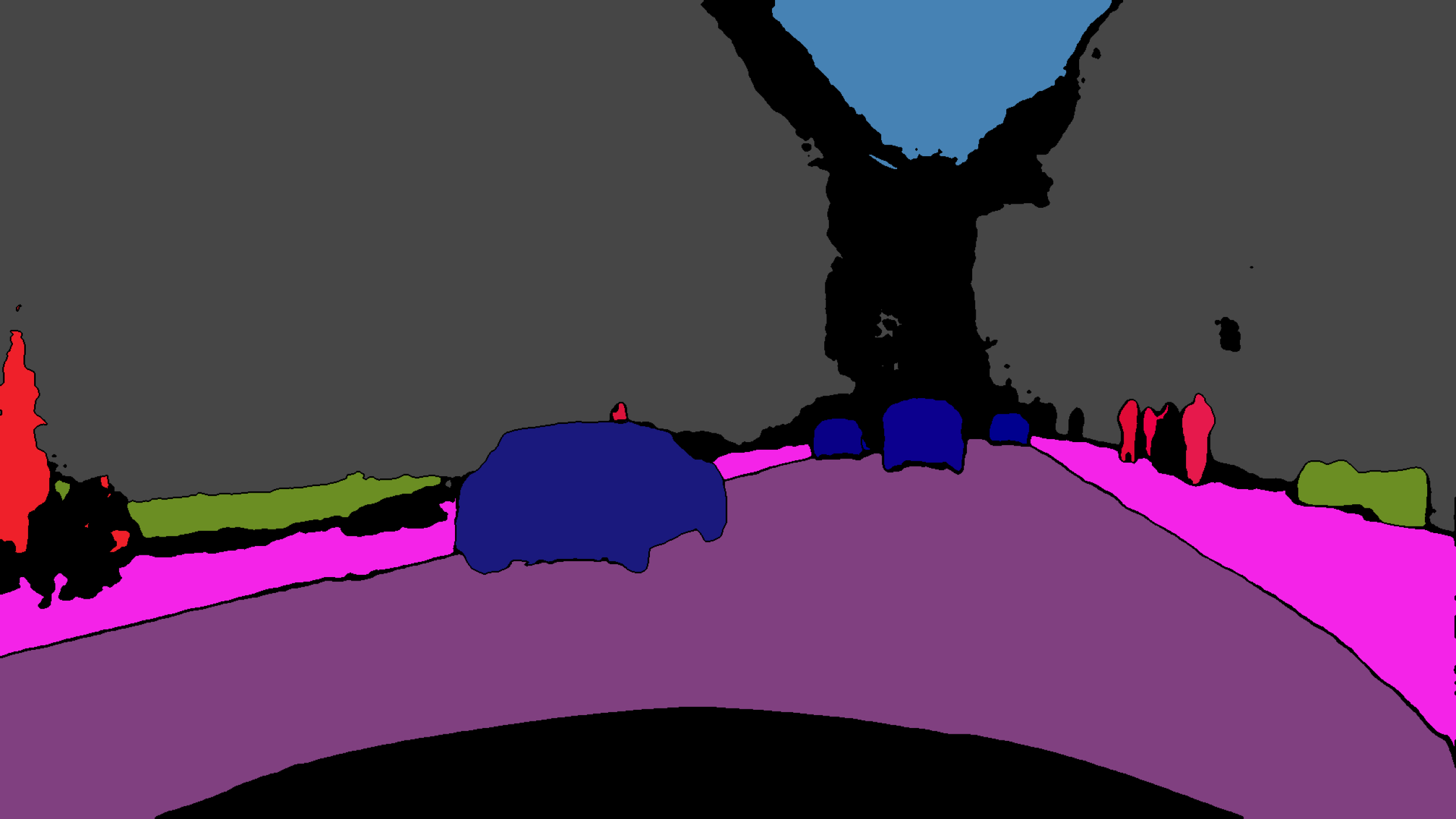} &
\includegraphics[width=\linewidth]{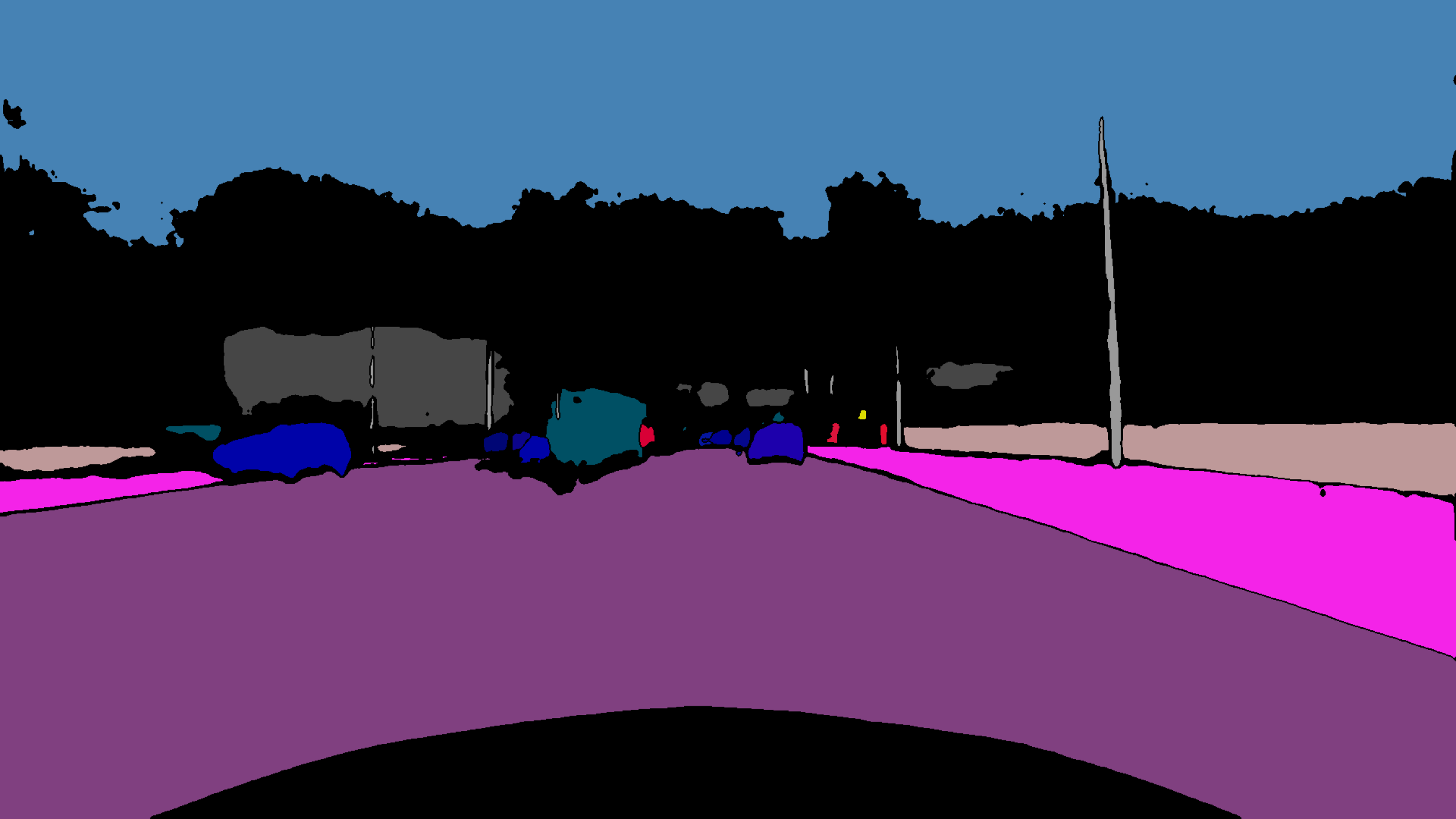} &
\includegraphics[width=\linewidth]{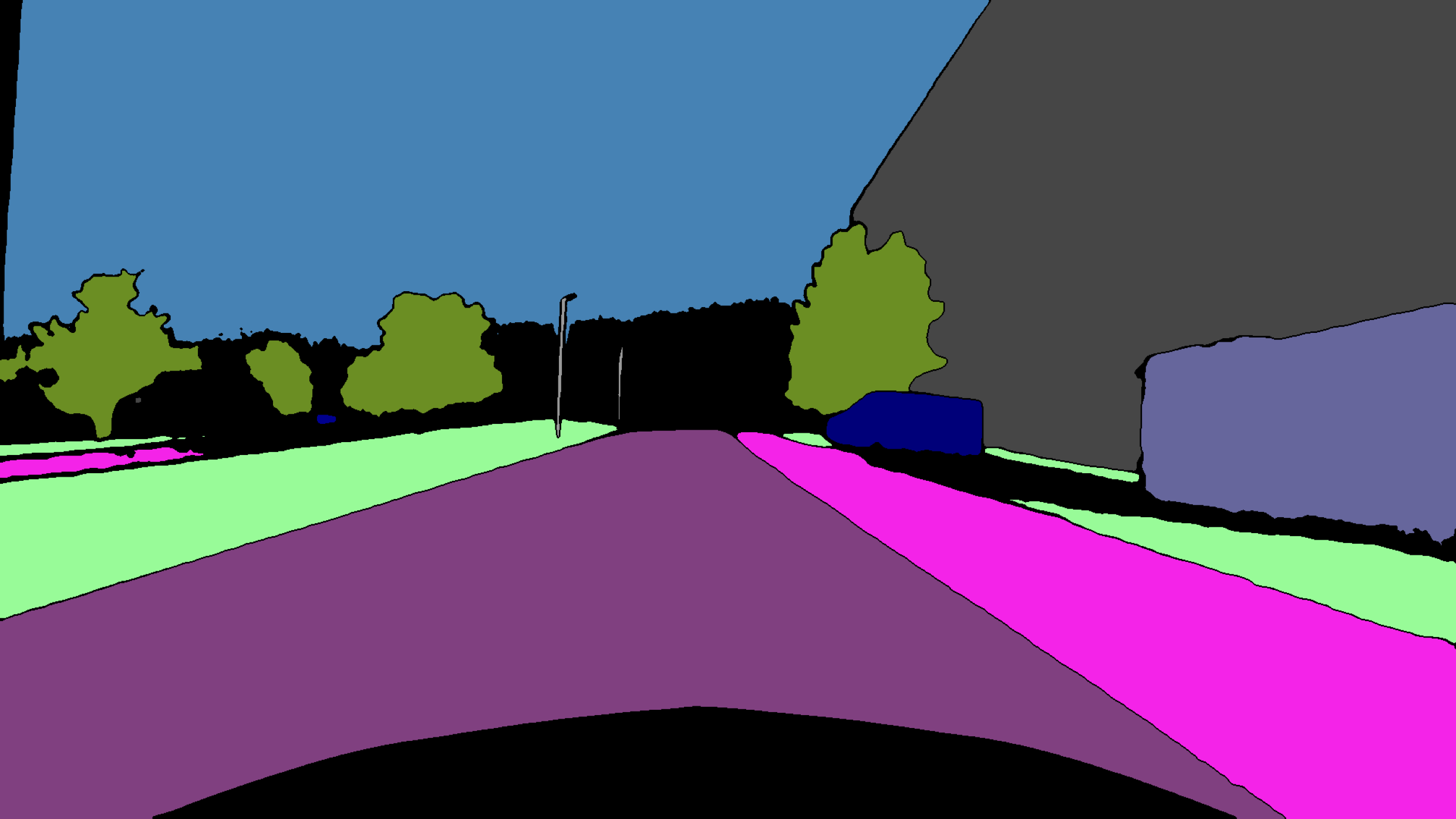} \\

mljp &
\includegraphics[width=\linewidth]{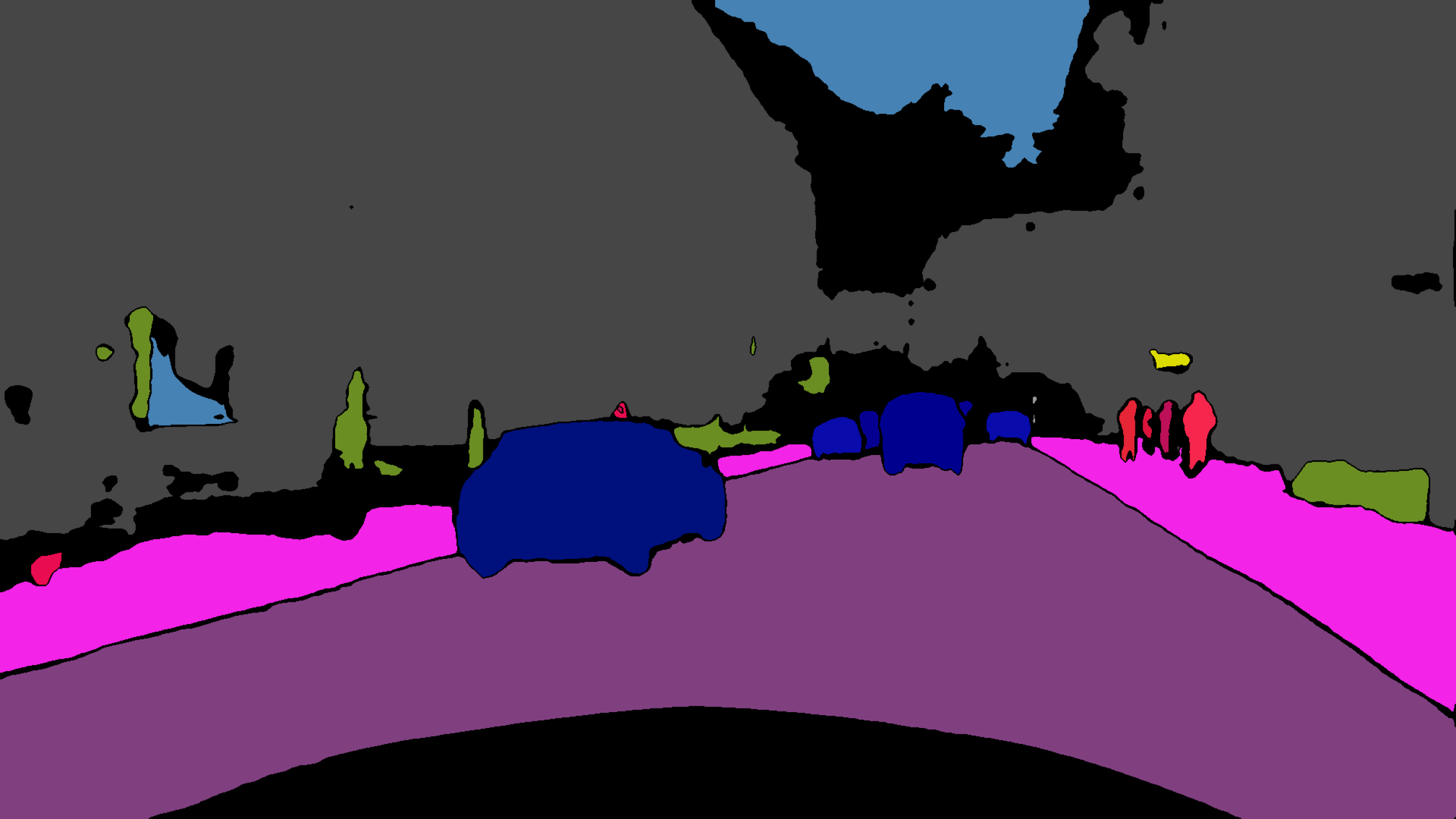} &
\includegraphics[width=\linewidth]{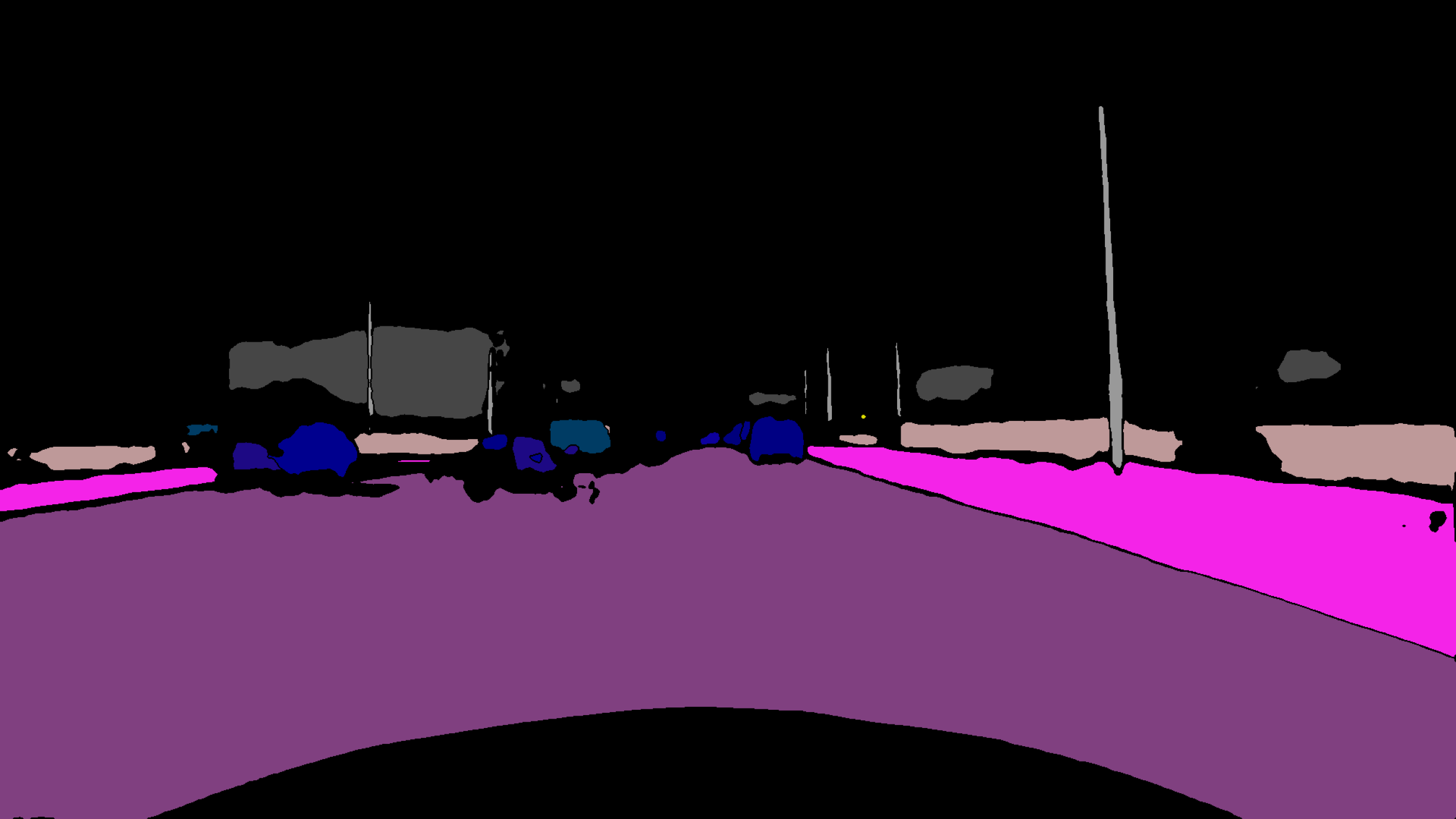} &
\includegraphics[width=\linewidth]{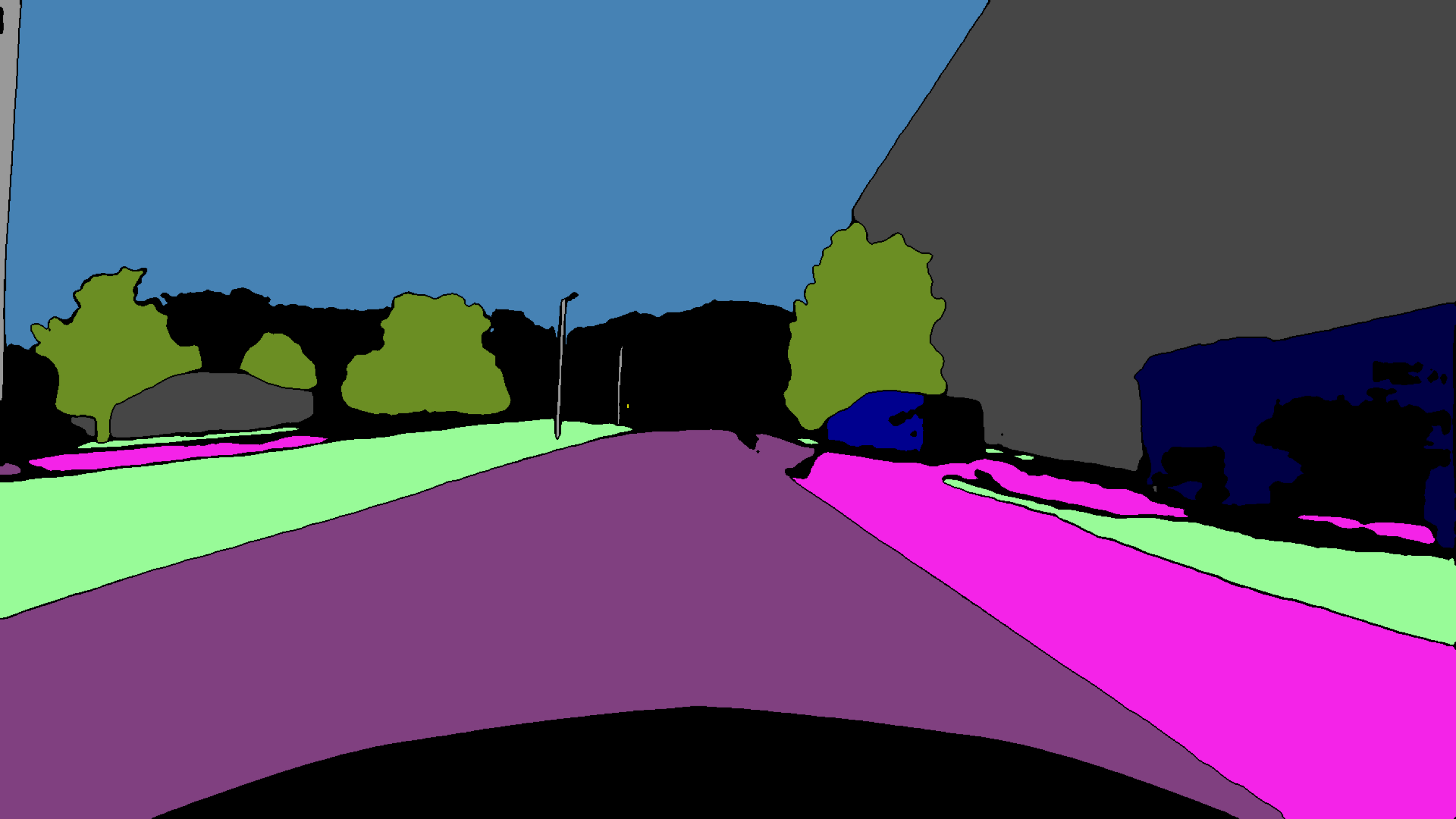} \\

\end{tabular}

\caption{\textbf{Qualitative results on MUSES test samples.}}
\label{fig:imgOutput}
\end{figure*}

In Figure \ref{fig:imgOutput}, we provide three qualitative comparisons between the panoptic segmentation ground truth and the predictions from the different teams, displayed using the Cityscape color palette. The first column corresponds to a real world difficult scene: an urban scene at night with snowfall and snow on the ground. Camera visibility is reduced due to melting snow. All methods successfully detect the cars. The pedestrians on the left constitute a particularly difficult detection case and are not well distinguished across the different approaches. The second column depicts a rainy night in a urban setting. Here, the water droplets greatly affect all participants' performance, with an even more pronounced effect in the background. For the last column, which shows a foggy daytime scene, with only a few object present in the frame. All participants provide strong predictions, producing accurate segmentation of the trees, road and building.

\section{Discussion \& Analysis}
The challenge results reveal three main trends. First, performance gains are largely driven by strong initialization and weather adaptation strategies rather than by drastic changes to the segmentation architecture itself. Second, Mask2Former remains a highly competitive baseline under relatively modest training budgets. Third, simply appending projected LiDAR cues to RGB inputs is insufficient to fully exploit the 3D sensing capability of LiDAR.

\textbf{Common effective practices.}
A recurring pattern across competitive methods is to start from a strong pretrained model or an easier training stage, and then adapt the model to the target adverse-condition setting. For instance, the \textit{MLJP team} adopts a two-stage schedule, first training on clear-day data and then fine-tuning on all weather and illumination conditions. Similarly, the winning \textit{wg} submission builds upon pretrained CAFuser weights. Taken together, these observations suggest that robust initialization and curriculum-style adaptation are practical and effective strategies for adverse-condition panoptic segmentation.

\textbf{Mask2Former remains a strong and efficient baseline.}
Despite its conceptual simplicity, Mask2Former remains one of the strongest baselines in this benchmark. 
The RGB-only Mask2Former baseline achieves 46.9 PQ$_{\text{all}}$, outperforming the third- and fourth-ranked challenge submissions and remaining close to the second-ranked method. 
From a practical perspective, it also offers an attractive accuracy--complexity trade-off: competitive performance can be obtained with a standard architecture, a single GPU, and comparatively moderate training time. Nevertheless, the gap to CAFuser-CA$^2$ and to the winning \textit{wg} submission indicates that there is still clear room for improvement once stronger multimodal and condition-aware reasoning is introduced. Since hardware, image resolution, and initialization differ across submissions, the reported training times should be interpreted only as rough indicators rather than as a strictly controlled efficiency comparison.

\textbf{More modalities do not automatically translate into better performance.}
The leaderboard does not support a simplistic conclusion that adding more modalities alone is sufficient to improve robustness. Instead, the results suggest that \emph{how} multimodal information is fused is more important than the mere number of sensors. Both MLJP and ElieT report that straightforward early or intermediate fusion strategies do not improve over RGB-only Mask2Former. By contrast, the \textit{michele24} method obtains a modest gain through gated residual modulation with an explicit validity mask, and the best-performing \textit{wg} submission employs a condition-aware fusion design together with a richer sensor suite. This indicates that adverse-weather robustness depends not only on access to complementary modalities, but also on structured fusion mechanisms that can adaptively determine when and how each modality should be trusted.

\textbf{Why naive LiDAR fusion remains limited.}
A key observation from several submissions is that simple LiDAR fusion strategies remain underwhelming. In most naive variants, LiDAR is first projected to the image plane and represented as auxiliary 2D channels, such as sparse depth, height, or intensity, which are then concatenated with RGB inputs or intermediate features. While convenient, this design largely collapses the native 3D structure of LiDAR into image-aligned tensors and therefore underutilizes its geometric information. Such representations do not explicitly preserve point sparsity patterns, vertical geometry, occlusion structure, or range-dependent reliability, all of which are important under adverse weather and low-visibility conditions. As a result, the network tends to treat LiDAR as a set of additional image channels rather than as a complementary 3D sensing modality. This likely explains why naive fusion often fails to outperform strong RGB-only baselines.

\textbf{Implications for future work.}
These findings suggest several promising directions for future research. First, stronger weather adaptation strategies, such as staged training or domain-specialized fine-tuning, appear to be consistently beneficial and should be further explored. Second, future multimodal panoptic segmentation models should preserve LiDAR geometry more explicitly, for example through point-, voxel-, or range-view encoders, rather than relying solely on projected 2D maps. Third, condition-aware fusion appears particularly important in adverse settings, where the reliability of each modality can vary significantly across weather and lighting conditions. More broadly, improving nighttime robustness remains an important open challenge, as nearly all methods still exhibit a clear performance drop from day to night.

\textbf{Summary.}
Overall, the results suggest that strong initialization and weather-aware adaptation are highly effective for robust panoptic segmentation. Meanwhile, Mask2Former remains a remarkably competitive baseline, indicating that large performance gains do not necessarily require complex redesigns of the segmentation head. In contrast, the mixed results of early- and mid-fusion variants suggest that simply concatenating projected LiDAR cues with RGB features is insufficient to fully exploit the complementary 3D information provided by LiDAR. To fully benefit from multimodal sensing, future methods will likely need more structured and geometry-aware fusion strategies than simple channel concatenation or image-plane projection alone.

\section{Conclusion}
This report presents the outcomes of the URVIS 2026 Challenge: MUSES-AXPS --- Adverse-to-the-eXtreme Panoptic Segmentation, conducted on the challenging MUSES dataset. We benchmark four participating methods against representative baseline approaches under diverse real-world weather and illumination conditions. The challenge results show that robust panoptic segmentation in adverse environments benefits substantially from strong pretraining and weather adaptation, while effective multimodal fusion remains an open research problem. Overall, this challenge provides a useful snapshot of the current state of adverse-condition panoptic segmentation and highlights promising directions for future progress in robust and condition-aware multimodal scene understanding.

\newpage
{
    \small
    \bibliographystyle{ieeenat_fullname}
    \bibliography{main}
}


\end{document}